\documentclass{article}

\usepackage{microtype}
\usepackage{graphicx}
\usepackage{booktabs} %

\usepackage{hyperref}

\usepackage{algorithmic,algorithm}

\usepackage{amsmath}
\usepackage{amssymb}
\usepackage{mathtools}
\usepackage{amsthm}
\usepackage{enumitem}
\usepackage[capitalize,noabbrev]{cleveref}

\theoremstyle{plain}

\theoremstyle{definition}

\theoremstyle{remark}

\usepackage{color}
\usepackage{subcaption}
\usepackage{diagbox}
\usepackage{caption}
\usepackage{wrapfig}

\usepackage[in]{fullpage}  
\usepackage{mathpazo} 

\usepackage[utf8]{inputenc} %
\usepackage[T1]{fontenc}    %
\usepackage{hyperref}       %
\usepackage{url}            %
\usepackage{booktabs}       %
\usepackage{amsfonts}       %
\usepackage{nicefrac}       %
\usepackage{microtype}      %
\usepackage{lipsum}         %
\usepackage{todonotes}
\hypersetup{
    colorlinks=true,
    citecolor=orange,
    linkcolor=blue
}

\AtBeginDocument{\setlength\abovedisplayskip{5pt}}
\AtBeginDocument{\setlength\belowdisplayskip{5pt}}

\usepackage{booktabs}       
\usepackage{amsfonts}       
\usepackage{nicefrac}       
\usepackage{microtype}      
\usepackage{xcolor}         
\usepackage{color}
\usepackage{algorithm}
\usepackage{algorithmic}
\usepackage{bm}
\usepackage{amsmath}
\usepackage{colortbl}
\usepackage{multirow}

\newcommand{\name}{{MASIA}}
\newcommand{\loss}{\mathcal{L}}
\newcommand{\expect}{\mathbb{E}}
\newcommand{\buffer}{\mathcal{D}}
\newcommand{\batch}{\mathcal{B}}

\newcommand{\zzz}[1]{\textcolor{red}{[zzz: #1]}}
\newcommand{\solved}[1]{\textcolor{blue}{[#1]}}
\newcommand{\explain}[1]{\textcolor{orange}{[#1]}}

\newcommand{\remove}[1]{}

\title{Efficient Communication via Self-supervised Information Aggregation for Online and Offline Multi-agent Reinforcement Learning}
\date{\vspace{-5ex}}

\usepackage{authblk}

\author{Cong Guan}
\author{Feng Chen}
\author{Lei Yuan}
\author{Zongzhang Zhang}
\author{Yang Yu\thanks{Corresponding author}}
\affil{National Key Laboratory for Novel Software Technology, Nanjing University\\
\small Email: \{guanc, chenf, yuanl\}@lamda.nju.edu.cn, \{zzzhang, yuy\}@nju.edu.cn}

\begin{document}
\maketitle
\begin{abstract}
Utilizing messages from teammates can improve coordination in cooperative Multi-agent Reinforcement Learning (MARL). Previous works typically combine raw messages of teammates with local information as inputs for policy. However, neglecting message aggregation poses significant inefficiency for policy learning. Motivated by recent advances in representation learning, we argue that efficient message aggregation is essential for good coordination in cooperative MARL. In this paper, we propose \textbf{M}ulti-\textbf{A}gent communication via \textbf{S}elf-supervised \textbf{I}nformation \textbf{A}ggregation (MASIA), where agents can aggregate the received messages into compact representations with high relevance to augment the local policy. Specifically, we design a permutation invariant message encoder to generate common information-aggregated representation from messages and optimize it via reconstructing and shooting future information in a self-supervised manner. Hence, each agent would utilize the most relevant parts of the aggregated representation for decision-making by a novel message extraction mechanism. Furthermore, considering the potential of offline learning for real-world applications, we build offline benchmarks for multi-agent communication, which is the first as we know. Empirical results demonstrate the superiority of our method in both online and offline settings. We also release the built offline benchmarks in this paper as a testbed for communication ability validation to facilitate further future research.
\end{abstract}

\maketitle

\section{Introduction}\label{sec:introduction}
 Multi-Agent Reinforcement Learning (MARL)~\cite{cooperativesuvery} has attracted widespread attention \cite{cooperativesuvery, du2021survey} recently, achieving remarkable success in  many complex domains~\cite{canese2021multi},
such as traffic signal control \cite{du2021multi}, droplet control \cite{liang2021parallel}, active voltage control \cite{DBLP:conf/nips/WangXGSG21}, and dynamic algorithm configuration~\cite{xue2022multiagent}. For better coordination on further applications, some issues like non-stationarity~\cite{papoudakis2019dealing}, scalability~\cite{christianos2021scaling} remain to be solved. To solve the non-stationarity caused by the concurrent learning of multiple policies and scalability as the agent number increases, most recent works on MARL adopt the \textit{Centralized Training and Decentralized Execution} (CTDE) \cite{ctde2016, lyu2021contrasting} paradigm, which includes both value-based methods \cite{vdn, qmix2018, qplex, ijcai2022p85} and policy gradient methods \cite{coma, maddpg, dop,ye2022towards}, or other techniques like transformer~\cite{wen2022multiagent}. Under the CTDE paradigm, however, the coordination ability of the learned policies can be fragile due to the partial observability in the multi-agent environment, which is a common challenge in many multi-agent tasks~\cite{mao2020information}. 
While recurrent neural networks could in principle relieve this issue by conditioning the policy on action-observation history \cite{hausknecht2015deep},
the uncertainty of other agents (e.g., states and actions) at execution time can result in catastrophic miscoordination and even sub-optimality \cite{ndq,i2c}.

Communication shows great potential in solving these problems~\cite{zaiem2019learning,zhu2022survey}, with which agents can share information such as observations, intentions, or experiences to stabilize the learning process, leading to a better understanding of the environment (or the other agents) and better coordination as a result. Previous communication methods either focus on generating meaningful information~\cite{ndq,kim2020communication,zhang2020succinct} for the message senders, or design techniques such as attention mechanism~\cite{tarmac,niu2021multi}, message gate~\cite{acml,i2c} to filter the most relevant information on raw received messages. These approaches treat the received information as a black box and tacitly assume that policy networks can automatically extract the most critical information from multiple raw messages during policy learning.
On this occasion, with the only signal given by reinforcement learning, the extraction process may be reasonably inefficient, especially in complex scenarios.

 Motivated by recent advances in state representation learning~\cite{spr, curl}, which reveals that auxiliary representation objectives could facilitate policy learning~\cite{ericsson2021self},
 we aim at efficiently aggregating information as compact representations for policy by designing a novel communication framework \textbf{M}ulti-\textbf{A}gent communication via \textbf{S}elf-supervised \textbf{I}nformation \textbf{A}ggregation (MASIA).
 Specifically, representations are optimized through self-supervised objectives, which encourages the representations to be both abstract of the true states and predictive of the future information. Since agents are guided towards higher cumulative rewards during policy learning, correlating representations with true states and future information could intensify the learning signals in policy learning. In this way, the efficiency of policy learning could be encouraged. Also, considering that permutation invariance of representations can also promote efficiency, we design a self-attention mechanism to maintain the invariance of obtained representations. We also design a network that weighs the aggregated representation for individual agents to derive unique and highly relevant representation to augment local policies for efficient coordination.

 On the other hand, the application of Reinforcement Learning (RL) in real-world scenarios faces significant challenges as the interaction with environments is typically costly or even impossible~\cite{levine2020offline}. Offline RL is recently proposed to help solve this concern, which only learns the agent policy from a fixed offline dataset without interaction with the environment. Despite the significance and popularity of this topic, current works primarily focus on single agent setting~\cite{levine2020offline,peng2019advantage,fujimoto2019benchmarking}, or multi-agent coordination without communication~\cite{icq,omar,tian2022learning,marlofflinebenchmark,zhang2023discovering}. Even communication plays a crucial role in multi-agent coordination, especially in partially observable scenarios~\cite {zhu2022survey}, there are neither any efficient approaches for this issue nor any testbeds to validate this setting.
 Based on this situation, we construct an offline testbed based on some popular scenarios where communication is indeed necessary to test the communication ability for different communication approaches. We hope these benchmarks can be utilized to benchmark the performance of various multi-agent communication algorithms and trigger more research about offline reinforcement learning concerning multi-agent communication.


To evaluate our method, we conduct extensive experiments on various cooperative multi-agent benchmarks, including Hallway~\cite{ndq}, Level-Based Foraging~\cite{papoudakis2021benchmarking}, Traffic Junction~\cite{tarmac}, and two maps from StarCraft Multi-Agent Challenge (SMAC)~\cite{ndq} to validate the communication effectiveness in online and offline settings. The online experimental results show that MASIA outperforms previous approaches, strong baselines, and ablations of our method, demonstrating the effectiveness of \name~for online learning. We also build the offline dataset based on the four environments above. Further experiments concerning offline learning show that \name~also performs well in the offline setting. 

Our main contributions are:
\begin{itemize}
    \item We propose a novel framework that uses a message aggregation network to extract from multiple messages generated by various teammates, with which we acquire a permutation invariant information aggregation representation. Agents can then use a novel \textit{focusing network} to extract the most relevant information for decision-making.
    \item Two representation objectives are introduced to make the information representation \textit{compact} and \textit{sufficient}, including the state reconstruction and multi-step future states prediction. 
    \item We construct an offline dataset for multi-agent communication, which considers multiple environments and various dataset settings. This dataset is set up to benchmark different communication algorithms under offline learning and encourage more research on offline multi-agent communication learning.
    \item Sufficient online results on various benchmarks and communication conditions demonstrate that our proposed approach significantly improves the communication performance, and visualization results further reveal why it works. Additional experimental results on the offline dataset justify the effectiveness of our approach in the offline setting, inspiring further research in this field.
\end{itemize}

\section{Problem Formulation}

This paper considers a fully cooperative MARL communication problem, which can be modeled as Decentralised Partially Observable Markov Decision Process under Communication (Dec-POMDP-Com)~\cite{pomdp} and formulated as a tuple $\langle \mathcal{N}, \mathcal{S}, \mathcal{A}, P, \Omega, O, R, \gamma , \mathcal{M} \rangle$, where $\mathcal{N} = \{1, \dots, n\}$ is the set of agents, $\mathcal{S}$ is the set of global states, $\mathcal{A}$ is the set of actions, $\Omega$ is the set of observations, $O$ is the observation function,  $R$ represents the reward function, $\gamma \in [0, 1)$ stands for the discounted factor, and $\mathcal{M}$ indicates the set of messages. At each time step, due to partial observability,  each agent $i \in \mathcal{N}$ can only acquire the observation $o_i\in \Omega$ drawn from the observation function $O(s, i)$ with $s\in \mathcal{S}$, each agent 
holds an individual policy $\pi(a_i\mid \tau_i,m_i)$, where $\tau_i$ represents the history $(o_i^1, a_i^1, \dots, o_i^{t-1}, a_i^{t-1}, o_i^t)$ of agent $i$ at current timestep $t$, and $m_i \in \mathcal{M}$ is the message received by the agent $i$. The joint action $\boldsymbol{a}=\langle a_1, \dots, a_n \rangle$ leads to next state $s'\sim P(s'\mid s, \boldsymbol{a})$ and the global reward $R(s, \boldsymbol{a})$. The formal objective is to find a joint policy $\boldsymbol{\pi}(\boldsymbol{\tau}, \boldsymbol{a})$ to maximize the global value function $Q_{\rm tot}^{\boldsymbol{\pi}}(\boldsymbol{\tau}, \boldsymbol{a}) =\mathbb{E}_{s,\boldsymbol{a} }\left[\sum_{t=0}^\infty\gamma^tR(s, \boldsymbol{a})\mid s_0=s, \boldsymbol{a_0}=\boldsymbol{a}, \boldsymbol{\pi}\right]$, with $\boldsymbol{\tau}=\langle \tau_1, \dots, \tau_n \rangle$. As each agent can behave as a message sender as well as a message receiver, this paper considers learning useful message representation in the received end, and agents only use local information $o_i$ as message to share within the team.

We optimize the policy by value-based MARL, where deep Q-learning \cite{mnih2015human} implements the action-value function $Q(s, \boldsymbol{a})$ with a deep neural network $Q(\boldsymbol\tau, \boldsymbol a; \boldsymbol\theta)$ parameterized by $\boldsymbol{\theta}$. This paper follows the CTDE paradigm. In the centralized training phase, deep Q-learning uses a replay memory $\mathcal{D}$ to store the transition tuple $\langle \boldsymbol{\tau}, \boldsymbol{a}, r, \boldsymbol{\tau}^\prime \rangle$. We use $Q(\boldsymbol\tau, \boldsymbol a; \boldsymbol\theta)$ to approximate $Q(s, \boldsymbol a; \boldsymbol\theta)$ to relieve the partial observability. Thus, the parameters $\boldsymbol\theta$ are learnt by minimizing the expected Temporal Difference (TD) error:
 \begin{equation*}
     \mathcal{L}(\boldsymbol{\theta})=\mathbb{E}_{\left(\boldsymbol{\tau}, \boldsymbol{a}, r, \boldsymbol{\tau}^{\prime}\right) \in \mathcal{D}}\left[\left(r+\gamma V\left(\boldsymbol{\tau}^{\prime} ; \boldsymbol{\theta}^{-}\right)-Q(\boldsymbol{\tau}, \boldsymbol{a} ; \boldsymbol{\theta})\right)^{2}\right],
 \end{equation*}
where $V\left(\tau^{\prime} ; \boldsymbol{\theta}^{-}\right)=\max _{\boldsymbol{a}^{\prime}} Q\left(\boldsymbol{\tau}^{\prime}, \boldsymbol{a}^{\prime} ; \boldsymbol{\theta}^{-}\right)$ is the expected future return of the TD target and $\boldsymbol \theta^-$ are parameters of the target network periodically updated with $\boldsymbol \theta$.

When considering the offline setting, we have an offline dataset which is denoted as $\mathcal{B}$. The dataset is collected by specific behavior policy, and it keeps fixed during the whole training process. For Dec-POMDP problems, the offline dataset $\mathcal{B}$ is typically decomposed of a number of trajectories, that is to say, $\mathcal{B}:=\left\{(s^t,\{o_i^t\}_\mathcal{N},\{a_i^t\}_\mathcal{N},r^t)_{t=1}^T\right\}$, where $T$ denotes the length of the trajectory, $r^t$ means the reward obtained at the $t$-th timestep.
\remove{\zzz{Some symbols, e.g., $N$, $r^t$, and $T$ are not defined}\solved{Solved: $\mathcal{N}$ is introducted in the first paragraph and $t$ can be explained by $r^t$}}
When further considering multi-agent communication, we additionally record the receivers' id of the messages sent by each agent. This information portrays the communication channels allowed by the current problem.

\section{Method}
\label{sec:method}

\begin{figure*}[t]
  \centering
  \includegraphics[width=0.8\textwidth]{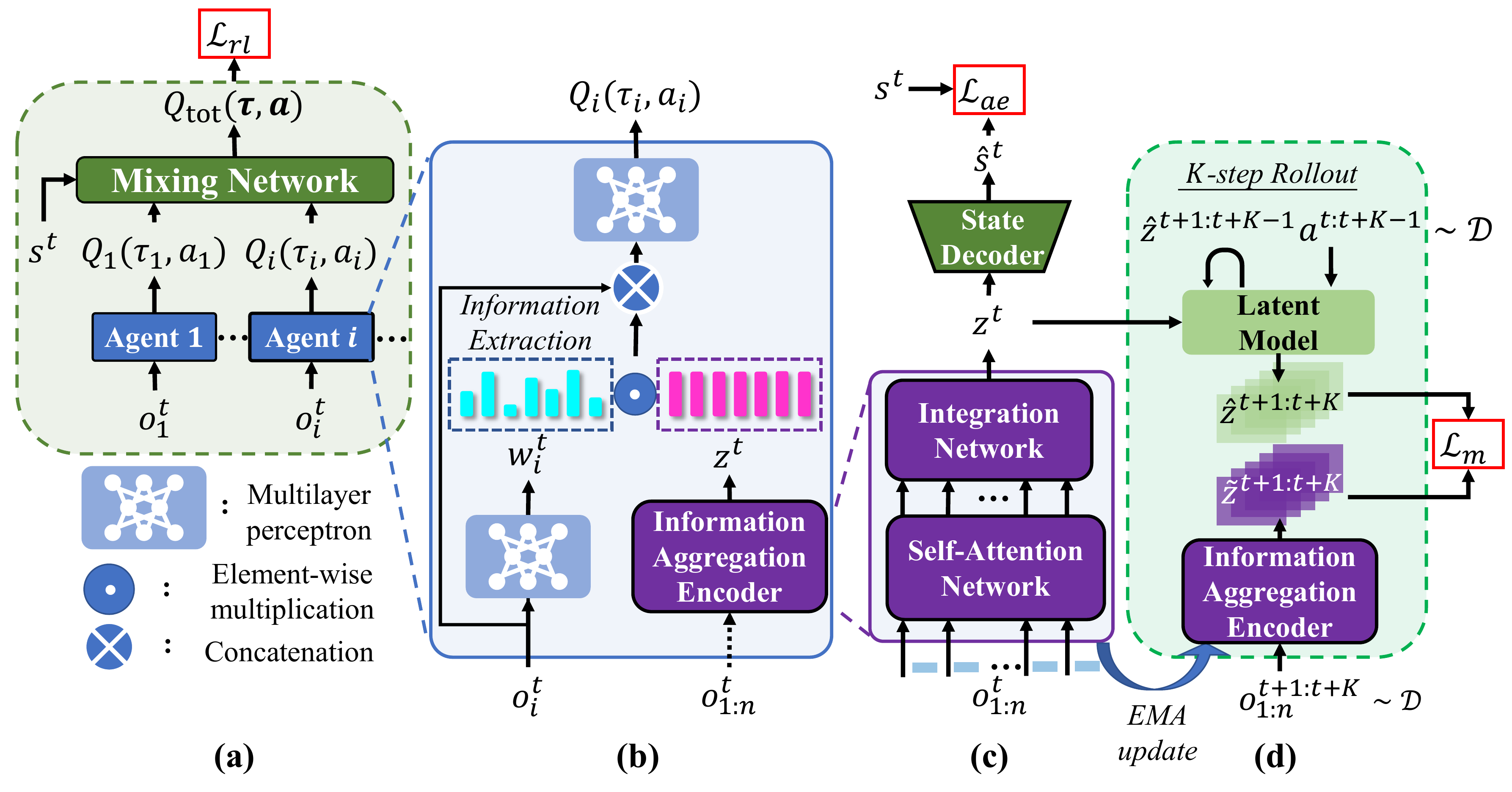}
  \caption{Structure of MASIA. (a) The overall architecture. (b) Information aggregation and extraction. (c) Information aggregation optimization. (d) Transition model learning. }
  \label{fig:framework}
\end{figure*}

In this paper, we propose efficient Multi-Agent communication via Self-supervised Information Aggregation (MASIA), a novel multi-agent communication mechanism for promoting cooperation performance. Redundant communications could increase the burden of information processing for each agent to make decisions and pose new challenges for information extraction since plenty of irrelevant information is contained in raw messages. To design an efficient communication mechanism, we believe two properties are of vital importance - \textit{sufficiency} and \textit{compactness}, where sufficiency means a rich amount of information, and compactness calls for higher information density.  

To meet the standard of sufficiency, a global encoder, which we call Information Aggregation Encoder (IAE), is shared among agents to aggregate the information broadcasted by agents into a common representation. With proper training, this representation could reflect the global observation so that each agent could obtain sufficient information from it to make decisions. As for compactness, we first design an auxiliary loss on the global representation to correlate it with the policy learning process, and make each agent only focus on the part of the representation related to its performance and coordination by the designed focusing network through excluding the unrelated parts. The entire framework of our method is shown in Figure~\ref{fig:framework}. Furthermore, a description of the training and execution processes can be found in Appendix~\ref{apdx:overall flow}. 


\subsection{Information Aggregation and Extraction} \label{sec:iae}
\textbf{Information Aggregation. } Believing that the true state should be reflected from combined messages, we design the aggregation encoder to be capable of subsuming all the messages sent from agents. Also, the communication system in multi-agent systems is flexible and permutation invariant in nature, which calls for a permutation invariant structure for the aggregation encoder. Based on these beliefs, we apply a self-attention mechanism to aggregate multiple messages from different teammates:
\begin{align}
\bm{Q},\bm{K},\bm{V} &= \mathtt{MLP}_{Q,K,V}([o^t_1,\dots,o^t_i,\dots,o^t_n]),\\
\bm{H} &= \mathtt{softmax}(\frac{\bm{Q}\bm{K}^T}{\sqrt{d_k}})\bm{V},
\end{align}
where the learnable matrices $Q$, $K$, and $V$ transform the perception from all agents into the corresponding query $\bm{Q}$ , key $\bm{K}$, and value $\bm{V}$, which are the concepts defined in the attention mechanism~\cite{vaswani2017attention}.
Specifically, each row vector of $\bm{H}$ can be seen as a querying result of one agent for all available information, and the hidden state $\bm{H}$ will be fed into the subsequent integration network to finally obtain the output aggregated representation $z^t$. A detailed discussion about the design of the integration network can be found in Appendix \ref{apdx:network_arch}. 
\begin{figure}[tbp!]
\centering
\includegraphics[width=0.6\linewidth]{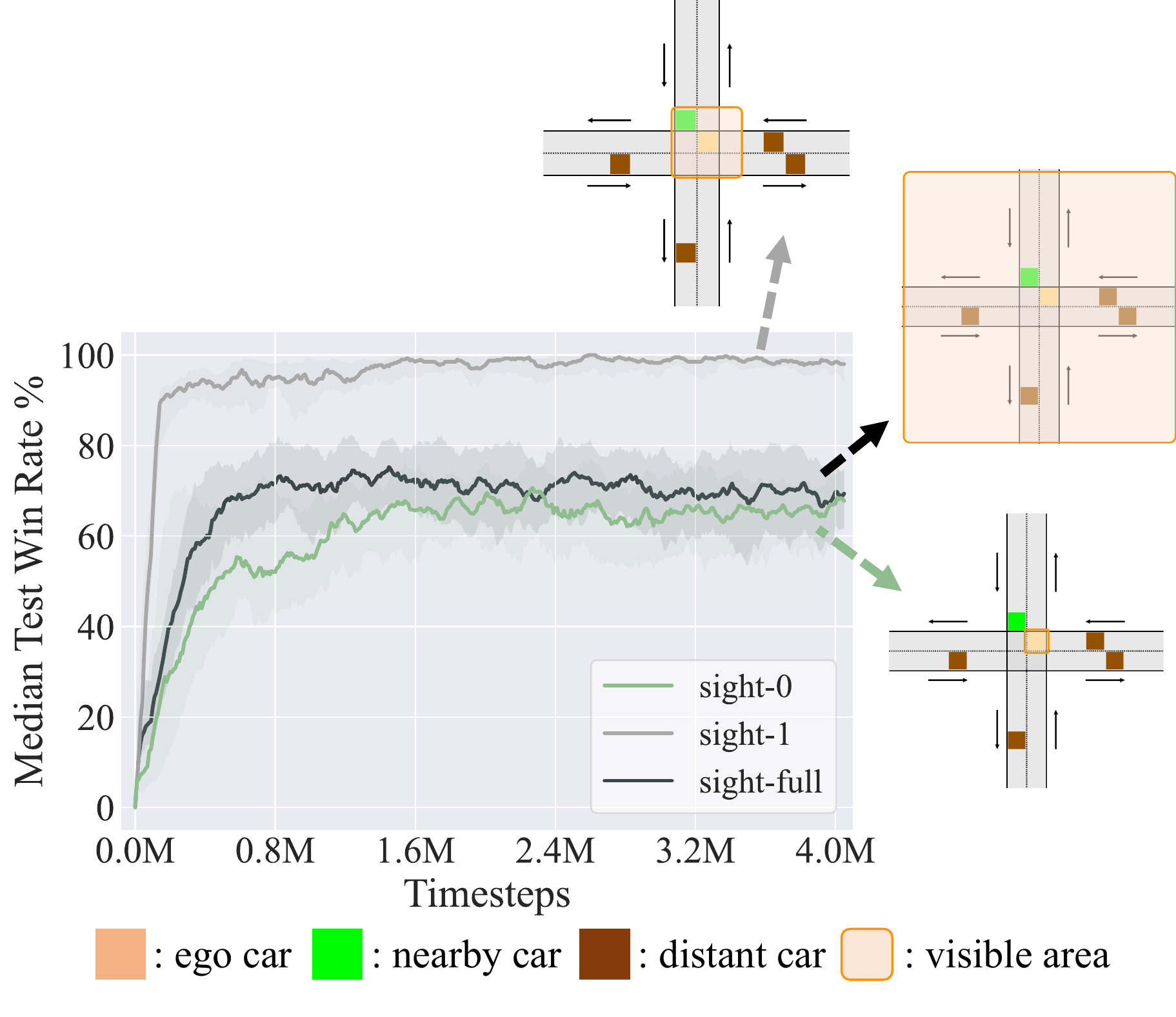}
\caption{A toy experiment for information redundancy on the task of Traffic Junction.}
\label{fig:traffic toy exp}
\end{figure}
In the centralized training phase, we use the aggregated representation $z^t$ as extra input in addition to the individual observation to feed the value function. Since $z^t$ contains the information required to determine the true state, taking $z^t$ as extra input could reduce the uncertainty about the environment states and produce better estimations on the Q-values for value functions under any value-based policy learning algorithm.

\textbf{Information Extraction. } Similar to the decision process of human beings, global messages are usually redundant for an individual agent to make good coordination in communication systems. 
For example, on the task of Traffic Junction~\cite{tarmac}, one natural idea is that the information of neighboring cars are more important for agents to perceive than those distant ones, and the unrelated information in global message may sometimes even confuse the agents and impede the learning when the map is large. A toy experiment shown in Figure~\ref{fig:traffic toy exp} supports this idea. We apply the QMIX algorithm in Traffic Junction tasks with different sight settings. The results show that the agents learn better policies when in a small-sight setting (sight-1) than both in a super-limited-sight setting (sight-0) and full-sight setting (sight-full), motivating a demand of efficient message extraction.

To make each agent capable of deciding its own perceptive area, we employ the \textbf{focusing network} to weigh the aggregated representations for each agent. 
The focusing network is designed as a Multi-Layer Perception (MLP) with the Sigmoid output activation function to ensure that each dimension of $w_i^t$ is bounded between $0$ and $1$. By taking element-wise multiplication with $z^t$, a unique representation could be distilled for individual agents. In this way, if the focusing network produces higher weights on specific dimensions, changes in aggregated representations on these dimensions would be more significant and thus, the agent would be more sensitive to aggregated representation on these parts. On the contrary, if some near-zero weights are outputted on some dimensions, information on those dimensions would be filtered out. In particular, although the information extraction process can, to some extent, reflect the specificity of each agent, we stress the local information by feeding it into the subsequent network together with the extracted representation.

\subsection{Information Representation Optimization}
\label{sec:inf rep opt}
As for the learning process of aggregated representation, we consider two typical objectives in global encoder training: reconstruction and multi-step prediction, which constrain the representations produced by the global encoder to be sufficient and compact, respectively. For the reconstruction objective, we employ an additional decoder, which aims to reconstruct the global state by the aggregated representation to allow self-supervision on the global encoder. Specifically, the decoder is optimized together with the aggregation encoder by reconstructing the global states $s^t$ from the multiple received messages $\bm{o}^t$: 
\begin{equation}
    \loss_{ae}(\theta, \eta) = \expect_{\bm{o}^t, s^t} \| g_{\eta}(z^t) - s^t\|^2_2,\quad~z^t = f_{\theta}(\bm{o}^t),~
\end{equation}
where $f_{\theta}, g_{\eta}$ denote the encoder network parameterized by $\theta$ and the decoder network parameterized by $\eta$, respectively. This loss term resembles a classical auto-encoder loss, while the decoder here is not to reconstruct the input, but to recover the global state from representations instead. By utilizing this loss, we guide the encoder to extract observational features that can help infer the global state and let $z^t=f_{\theta}(\bm{o}^t)$ be a sufficient representation. 

As for the multi-step prediction objective, we constrain the produced representation to be predictive of future information. Specifically, we design a transition model $h_{\psi}: \mathcal{Z}\times\mathcal{A}^n\rightarrow \mathcal{Z}$ parameterized by $\psi$ as auxiliary model, which predicts the aggregated representation $z^{t+1}$ on next step $t+1$ through the aggregated representation $z^t$ and joint action $\bm{a}^t$ on current step $t$. We regress the predicted aggregated representation after $k$-step rollout on the actual aggregated representation of future messages $\bm{o}^{t+k}$, updating both the aggregation encoder and the auxiliary model via the multi-step prediction loss:
\begin{equation}
    \begin{aligned}
    &\resizebox{0.49\textwidth}{!}{$\loss_{m}(\theta, \psi) = \expect_{\bm{o}^t, s^t, \bm{a}^t, \dots, \bm{a}^{t+K-1}, \bm{o}^{t+K}, s^{t+K}}
    \left[\sum_{k=1}^K \|\hat{z}^{t+k} - \Tilde{z}^{t+k}\|_2^2\right],$~}
    \\ &\hat{z}^{t+1} = h_{\psi}(\Tilde{z}^t, \bm{a}^t),~
    \\ &\hat{z}^{t+k} = h_{\psi}(\hat{z}^{t+k-1},~\bm{a}^{t+k-1}),\quad k=2,\dots,K,
    \\ &\Tilde{z}^{t+k} = f_{\theta}(\bm{o}^{t+k}),\quad k=0,\dots,K.
    \end{aligned}
    \end{equation}
To further stabilize the learning process, we apply the double network technique, which employs two networks with the same architecture but different update frequencies, for the aggeration encoder. The target network is updated via Exponential Moving Average (EMA) like in SPR~\cite{spr}.
By forcing the aggregated representation to be predictive of its future states, the aggregated representation could be more correlated with the information required for its decision-making, which meets the compactness standard. Combining these two objectives allows the aggregation encoder to extract more helpful information for agents to coordinate better.
To improve the capability of information extraction on individual agents, we also enhance the learning process of these components with an RL objective. Specifically, we consider minimizing the TD loss:
\begin{equation}
    \begin{aligned}
    \loss_{rl}(\theta,\phi) = \mathbb{E}_{\left(\boldsymbol{\tau}, \boldsymbol{a}, r, \boldsymbol{\tau}^{\prime}\right) \in \mathcal{D}}\Bigg[\Bigg(r+\gamma \max_{\boldsymbol{a}^{\prime}}Q_{\rm tot}\left(\boldsymbol{\tau}^{\prime},\boldsymbol{a}^{\prime} ; \theta^{-}, \phi^{-} \right)\\
    -Q_{\rm tot}(\boldsymbol{\tau}, \boldsymbol{a} ; \theta, \phi)\Bigg)^{2}\Bigg],~
    \end{aligned}
\end{equation}
where $Q_{\rm tot}$ is computed with individual $Q$-values. The computation of Q-values is actually dependent on the specific value-based learning algorithm. We apply it to prevalent methods, including VDN~\cite{vdn}, QMIX~\cite{qmix2018}, and QPLEX~\cite{qplex}. Moreover, the updating of the focusing network is coupled to the RL objective, making the weights produced by the focusing network could be task-sensitive, which could also facilitate policy learning. 

\subsection{Representation Pre-training for Offline Learning}\remove{\zzz{It seems that the offline method is straightforward. If a reviewer wants to reject the manuscript, his/her main reason might be that its method contribution over the conference version is not big.} \explain{Explain: it is a fact that the modifications concerning method are slight; we hope the offline part can support the core contribution of this extension.}}


Considering the significance of offline learning for real-world applications, we also cover the offline setting in this paper. When applying our approach to offline communication learning, we can only access a fixed offline dataset without interacting with the environment. Thus, the difficulty lies in how to fully utilize the limited offline data to learn multi-agent communication policies efficiently. On the other hand, as the aggregation encoder is updated together with the Q-networks, the representation space $\mathcal{Z}$ it derives is, to some extent, dynamic. Usually the encoder requires a certain amount of data for training to obtain a relatively stable representation space.\remove{\zzz{a stable point?}\explain{Explain: here we utilize "stable" to indicate that in the following training process the representation encoder will not fluctuate so much as to influence the training of the subsequent networks. I tend to use "be stable" because it is a more general expression while "a stable point" may be too absolute.}}

\begin{figure}
\centering
\includegraphics[width=\linewidth]{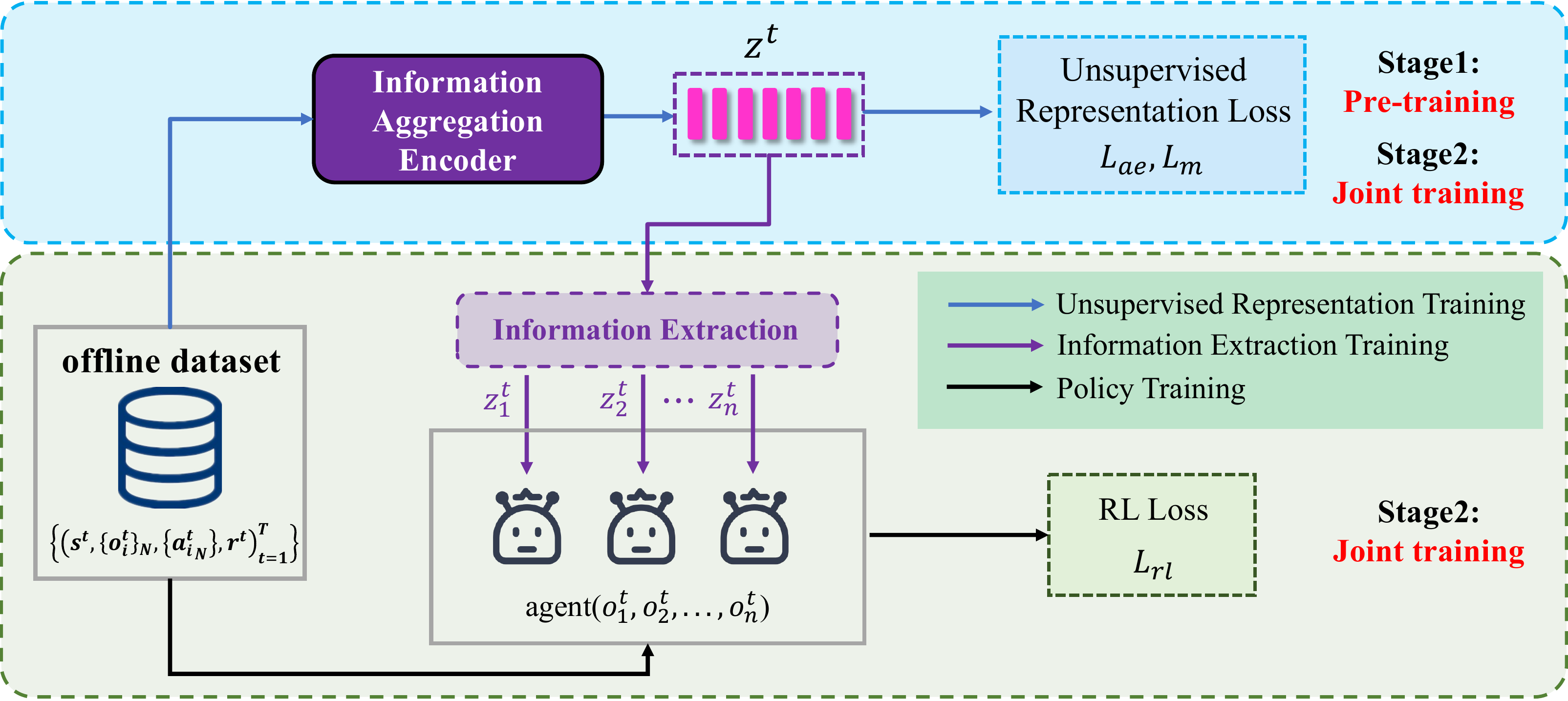}
\caption{The total training workflow of \name~for offline learning. Which includes two training stages, pre-training and joint training, respectively. During the stage of pre-training, we only optimize the unsupervised representation loss with the offline data, while for joint training we optimize the representation loss and RL loss together. Besides, we utilize agent$(o_1^t,...,o_n^t)$ to indicate the network inference process of the agent policy, which is to output action or Q-values.}
\remove{\zzz{agent(...) in the figure seems unclear.}\solved{Solved: add some additional explanations.}} 
\label{fig:representation pretraining}   
\end{figure}

Motivated by these analyses, we propose doing representation pre-training before optimizing the agent policy with the offline dataset. The core idea is to firstly obtain a relatively good aggregation encoder, of which the derived representation space is to some extent stable. Then we further update the Q-networks with the offline dataset and jointly fine-tune the aggregation encoder. The total workflow is depicted in Figure~\ref{fig:representation pretraining}. In fact, this practice brings some advantages: (1) the representation pre-training process typically optimizes the unsupervised learning objectives and updates only the aggregation encoder network, which does not require considering the Q-divergence problem; (2) pre-training the aggregation encoder network brings a more stable aggregation representation space, which will facilitate the following optimization of RL objective. With this practice, we aim to better utilize the offline dataset by considering the property of our approach. In some sense, it can also be considered a benefit of our approach's mechanism, and we adopt this practice in all offline experiments. Further ablation studies in Section~\ref{sec:offline_ablation} justifies the effectiveness of this practice.

\section{Experiment}
\label{sec:exp}
To evaluate the effectiveness of our approach both in the online and offline problem setting, we first conduct online communication learning on four benchmarks to validate the effectiveness of \name. Further, we build an offline dataset to support offline communication learning. Its purpose is to check the convergence performance of various multi-agent communication learning algorithms in the offline setting and justify whether \name can still perform well with only a fixed offline dataset.

\subsection{Online Multi-Agent Communication Learning}

We conduct experiments on various benchmarks with different communication request levels\footnote{The codes are available at \href{https://github.com/chenf-ai/MASIA}{https://github.com/chenf-ai/MASIA}}. Specifically, we aim to answer the following questions in this section:  1) How does our method perform when compared with multiple baselines in various scenarios (Section~\ref{evaluation of performance})?  2) What kind of knowledge has been learned by the information aggregation encoder (Section~\ref{insight of masia})? 3) How can the information extraction module extract the most relevant information for the individual from the learned embedding space (Section~\ref{Information Extraction of masia})?  4) Can MASIA be applied to different value decomposition baselines to improve their coordination ability and robustness in various communication conditions (Section~\ref{Generality of masia})?
\begin{figure*}[t!]
    \centering
    \includegraphics[width=0.8\linewidth]{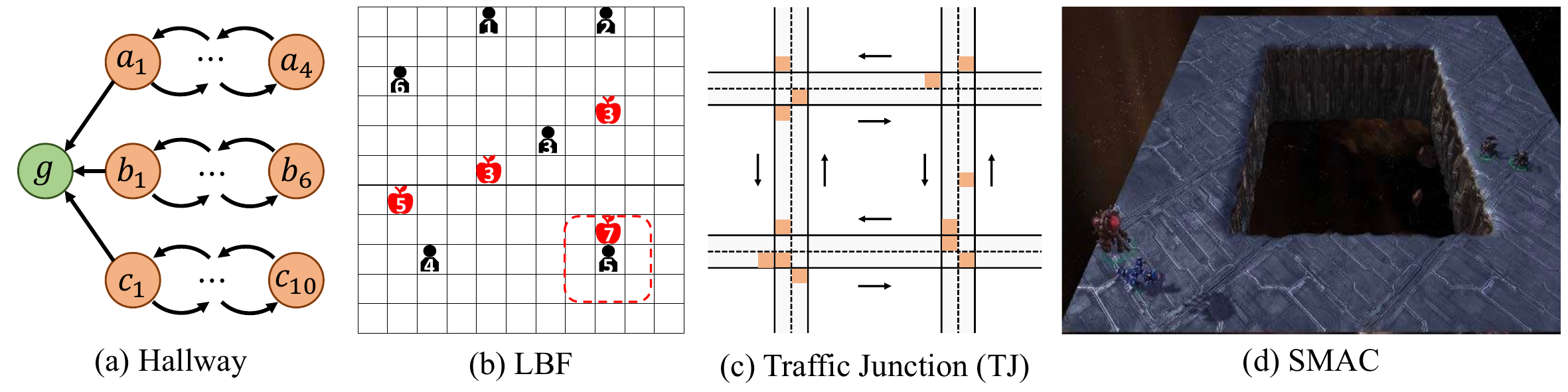}
	\caption{Multiple benchmarks used in our experiments.}
	\label{fig:env}
\end{figure*}
\begin{figure*}[t!]
	\centering
	\begin{subfigure}{0.95\linewidth}
	    \centering
	    \includegraphics[width=\linewidth]{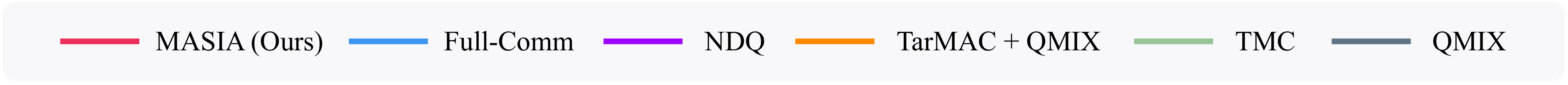}
	\end{subfigure}
	\begin{subfigure}{\linewidth}
		\centering
    	\begin{subfigure}{0.25\linewidth}
    		\centering
    		\includegraphics[width=\linewidth]{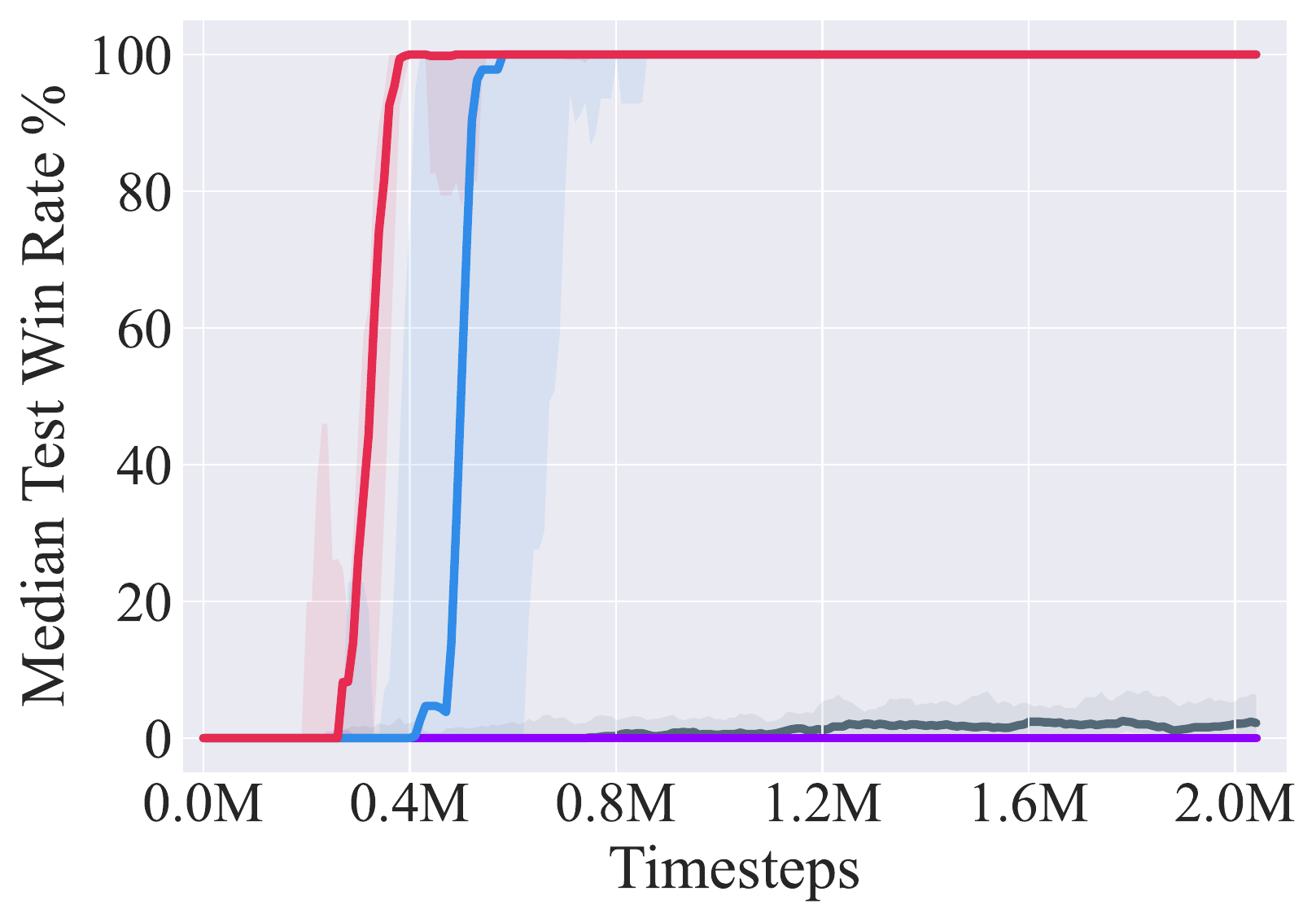}
    	    \caption{Hallway: 4x6x10}
    	\label{fig:hallway_easy}
    	\end{subfigure}
    	\hspace{-0.7em}
    	\begin{subfigure}{0.235\linewidth}
    		\centering
    		\includegraphics[width=\linewidth]{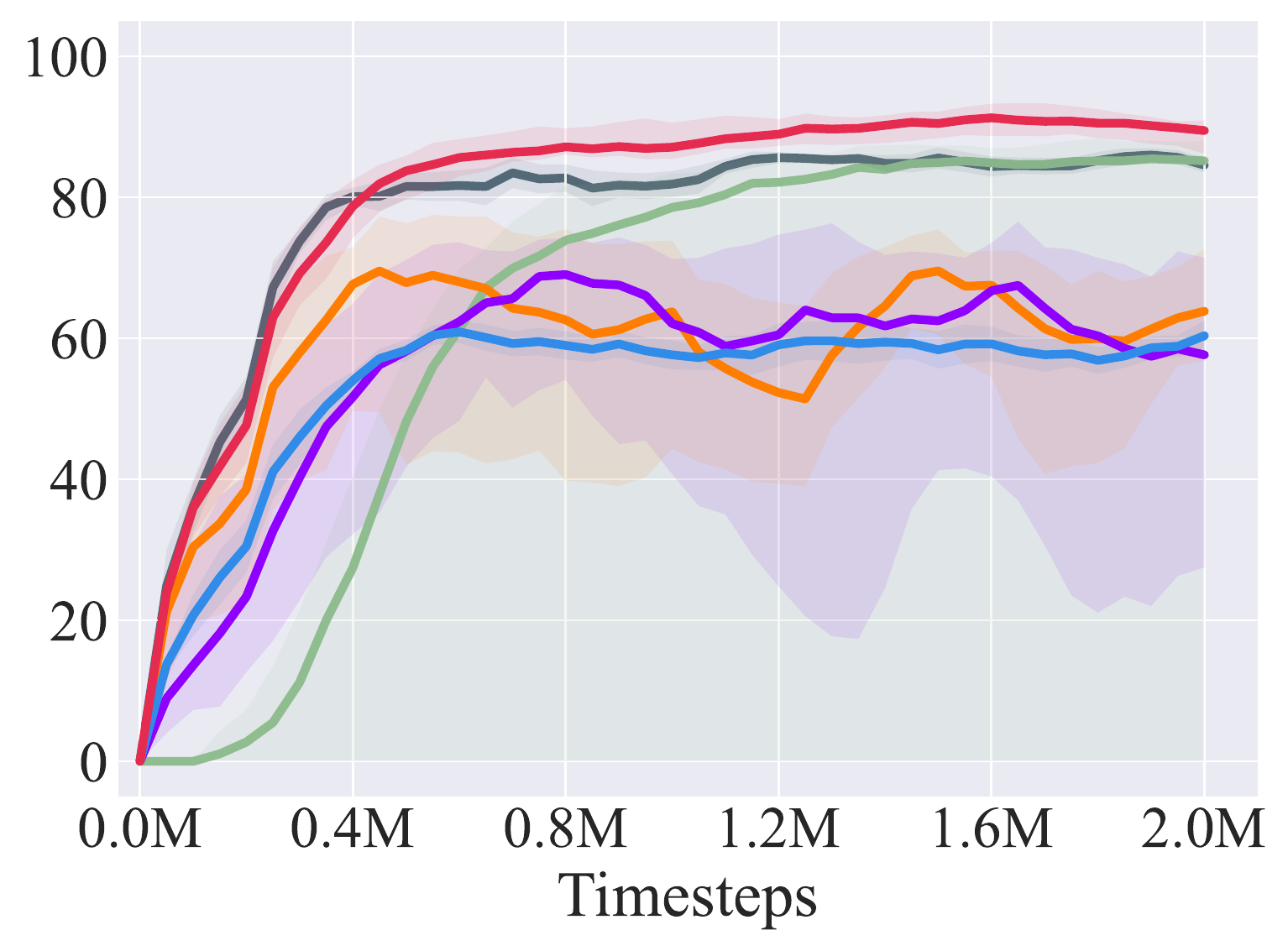}
    		\caption{LBF: 11x11-6p-4f-s1}
    	\label{fig:lbf_easy}
    	\end{subfigure}
        \hspace{-0.7em}
    	\begin{subfigure}{0.25\linewidth}
    		\centering
    		\includegraphics[width=\linewidth]{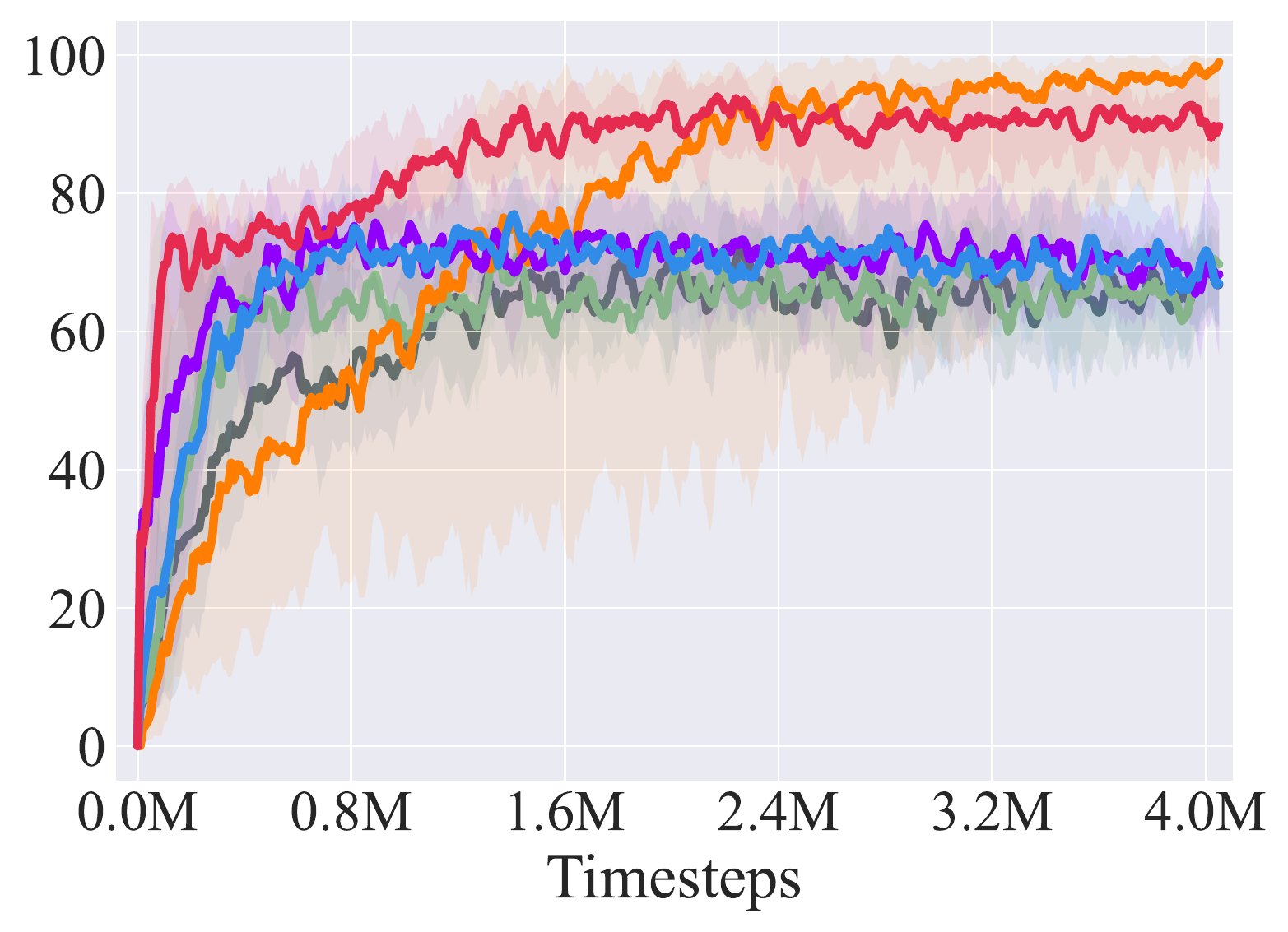}
    		\caption{\footnotesize{TJ: medium}}
    	\label{fig:traffic_easy}
    	\end{subfigure}
    	\hspace{-0.7em}
    	\begin{subfigure}{0.25\linewidth}
    		\centering
    		\includegraphics[width=\linewidth]{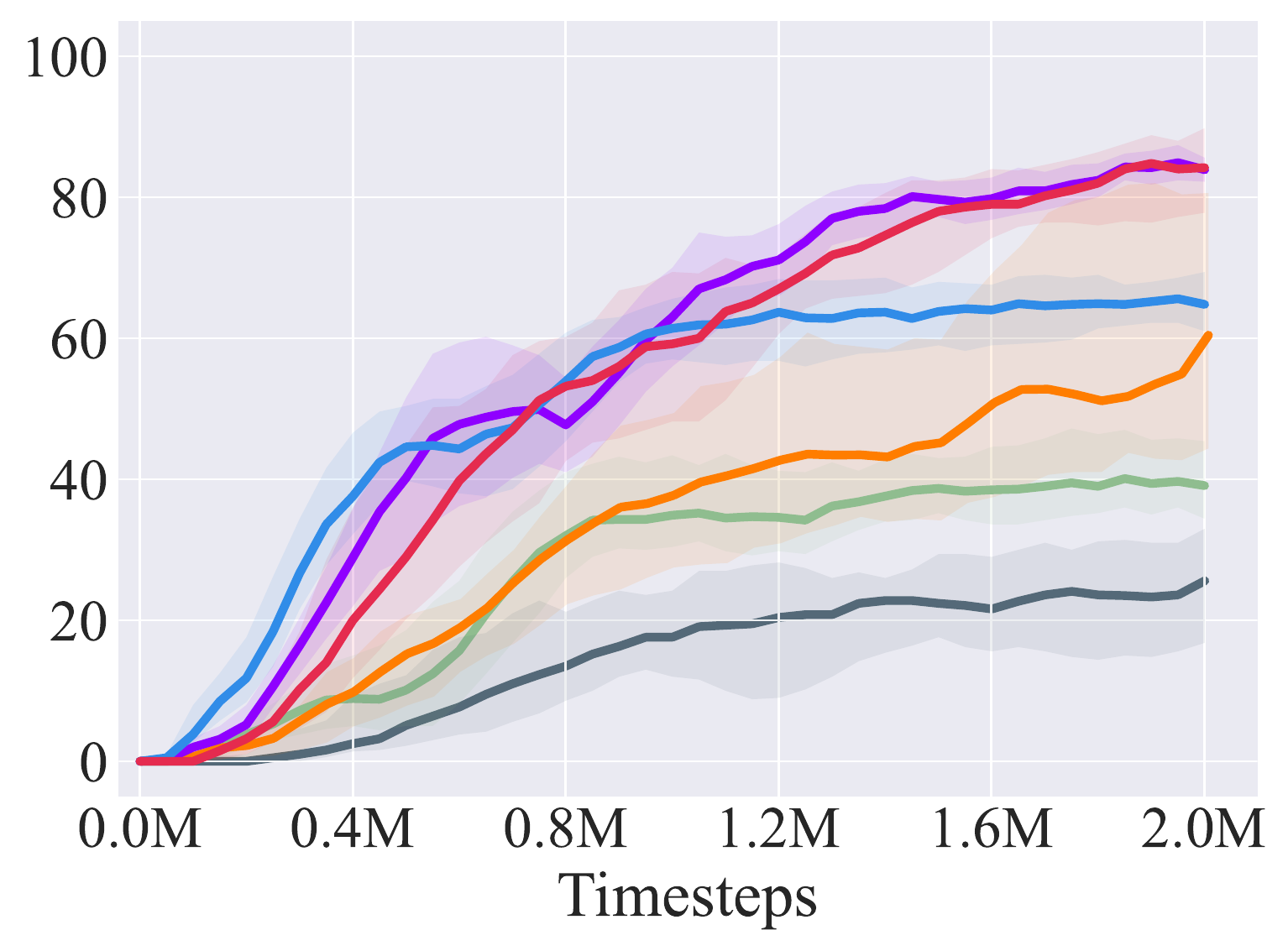}
    		\caption{SMAC: 1o10b\_vs\_1r}
    	\label{fig:sc2_easy}
    	\end{subfigure}
	\end{subfigure}
	\begin{subfigure}{\linewidth}
		\centering
    	\begin{subfigure}{0.25\linewidth}
    		\centering
    		\includegraphics[width=\linewidth]{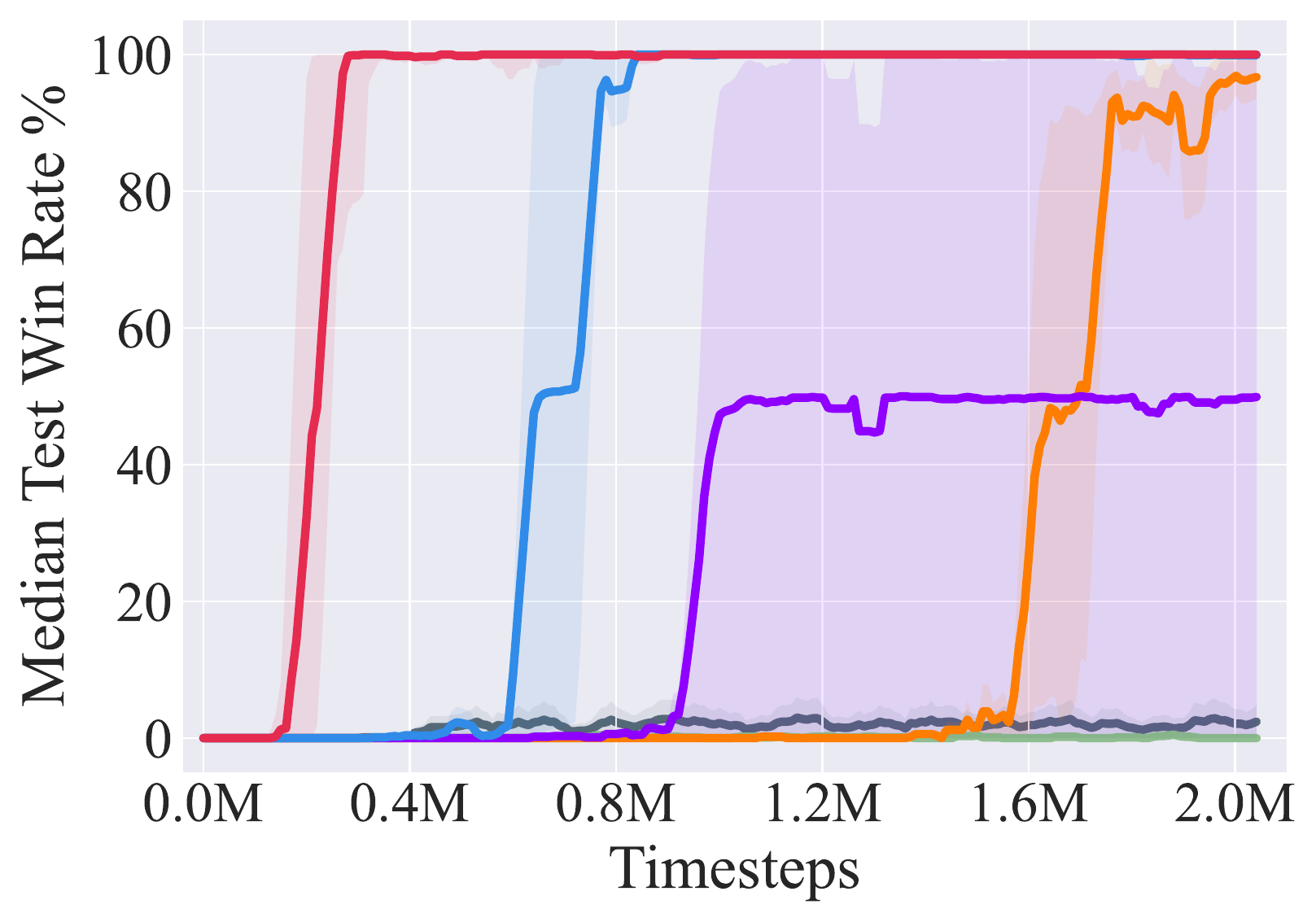}
    		\caption{Hallway: 3x5-4x6x10}
    	\label{fig:hallway_hard}
    	\end{subfigure}
    	\hspace{-0.7em}
    	\begin{subfigure}{0.24\linewidth}
    		\centering
    		\includegraphics[width=\linewidth]{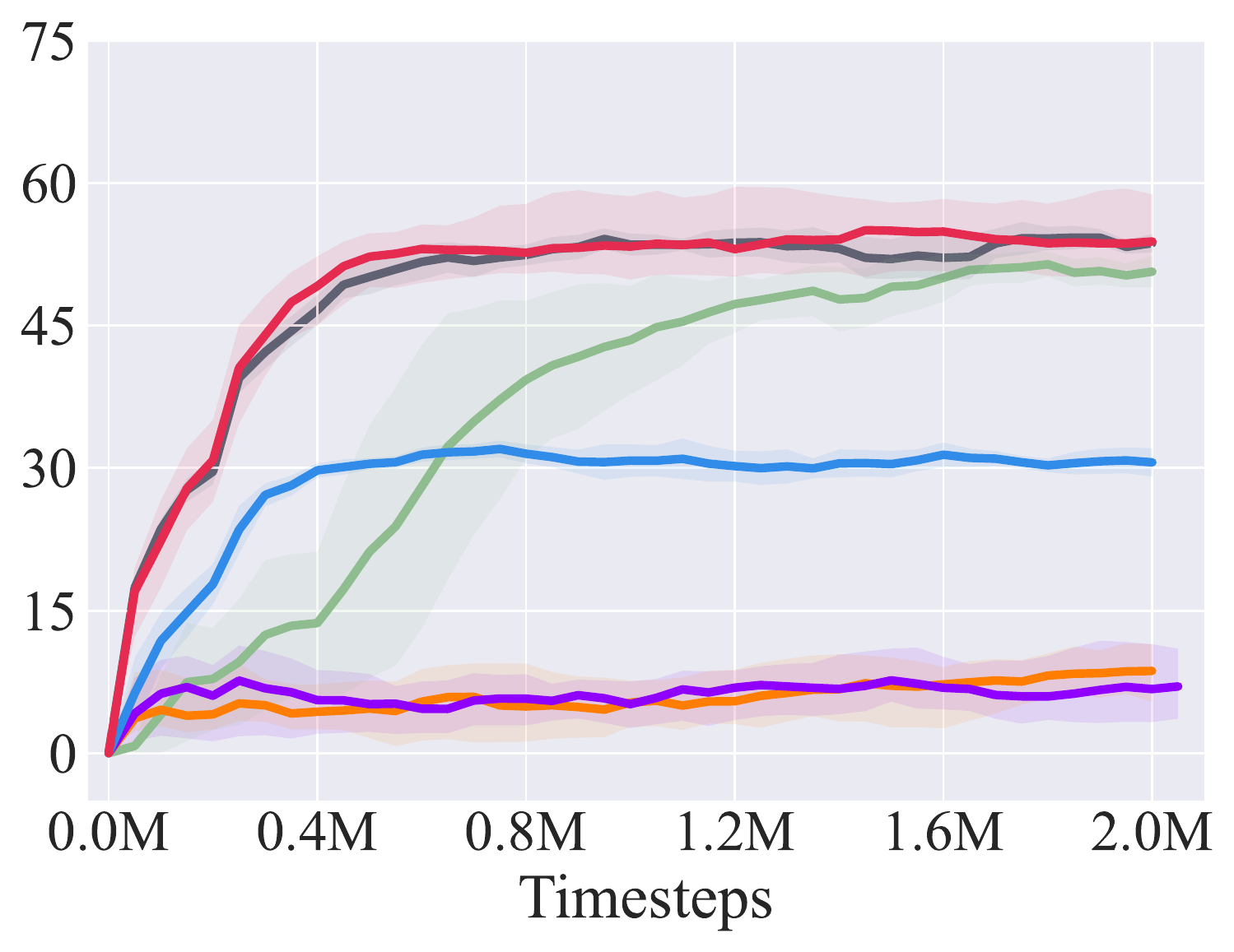}
    		\caption{LBF: 20x20-10p-6f-s1}
    	\label{fig:lbf_hard}
    	\end{subfigure}
    	\hspace{-0.7em}
    	\begin{subfigure}{0.25\linewidth}
    		\centering
    		\includegraphics[width=\linewidth]{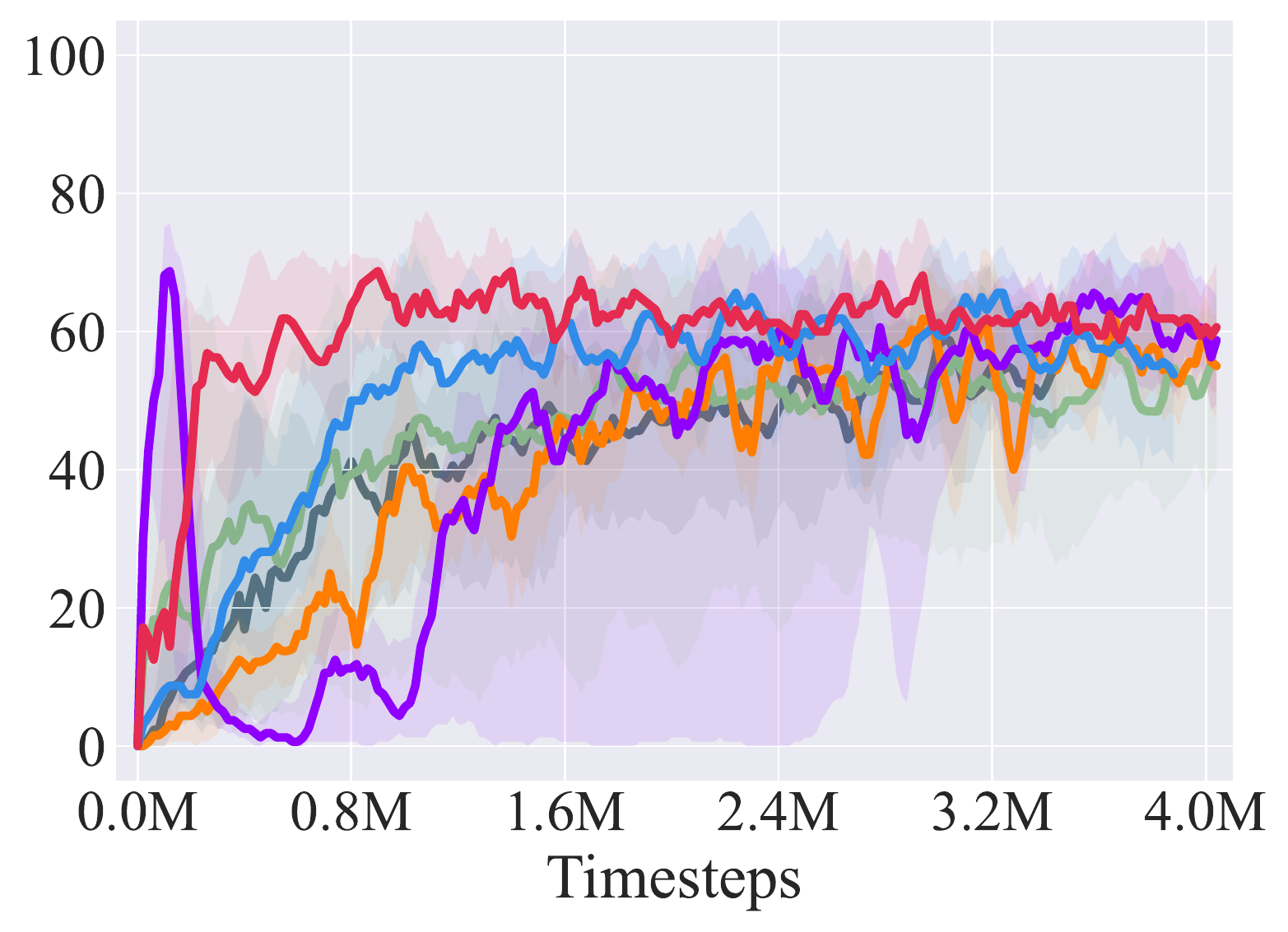}
    		\caption{TJ: hard}
    	\label{fig:traffic_hard}
    	\end{subfigure}
    	\hspace{-0.7em}
    	\begin{subfigure}{0.25\linewidth}
    		\centering
    		\includegraphics[width=\linewidth]{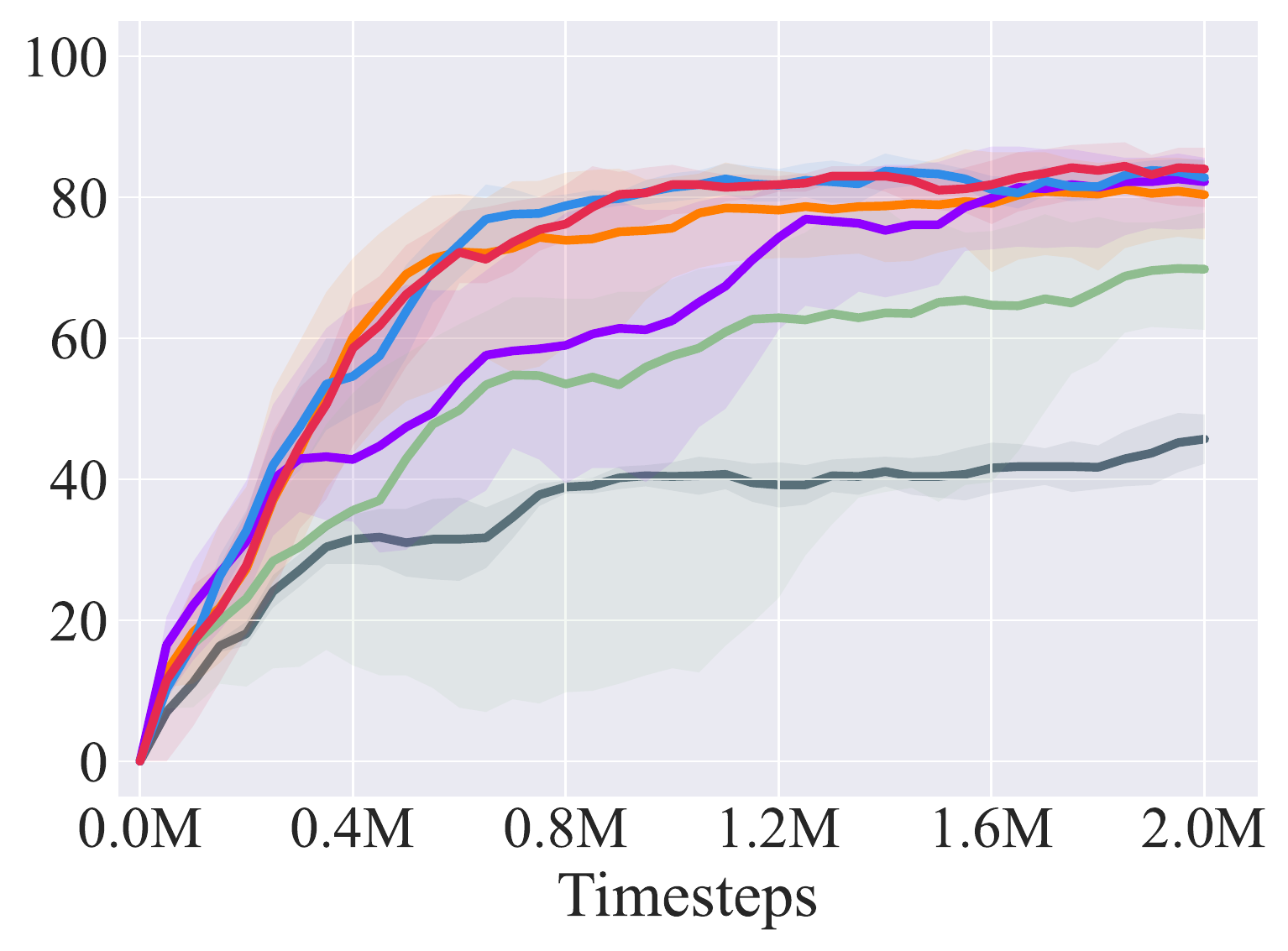}
    		\caption{SMAC: 1o2r\_vs\_4r}
    	\label{fig:sc2_hard}
    	\end{subfigure}
	\end{subfigure}
	\caption{Performance comparison with baselines on multiple benchmarks. }
	\label{fig:performance}
\end{figure*}
We compare MASIA against a variety of baselines, including communication-free methods and some state-of-the-art communication approaches. QMIX~\cite{qmix2018} is a strong communication-free baseline, and we use the implementation by PyMARL\footnote{Our experiments are all based on the PyMARL framework, which uses SC2.4.6.2.6923.} for comparison, which has shown excellent performance on diverse multi-agent benchmarks \cite{samvelyan2019starcraft}. TarMAC utilizes an attention mechanism to select messages according to their relative importance. The implementation we used is provided by \cite{ndq},
denoted as {TarMAC + QMIX}. NDQ \cite{ndq} aims at learning nearly decomposable Q functions via generating meaningful messages and communication minimization. TMC~\cite{zhang2020succinct} applies a temporal smoothing technique at the message sender end to drastically reduce the amount of information exchanged between agents. For the ablation study, we design a baseline only different in the communication protocol, which adopts a full communication paradigm, where each agent gets message from all other teammates at each timestep, denoted as FullComm.

We evaluate our proposed method on multiple benchmarks shown in Figure~\ref{fig:env}. Hallway \cite{ndq} is a cooperative environment under partial observability, where $m$ agents are randomly initialized at different positions and required to arrive at the goal $g$ simultaneously. We consider two scenarios with various agents and different groups, and different groups have to arrive at different times. Level Based Foraging (LBF)~\cite{papoudakis2021benchmarking} is another cooperative partially observable grid world game, where agents coordinate to collect food concurrently. Traffic Junction (TJ) \cite{tarmac} is a popular benchmark used to test communication ability, where many cars move along two-way roads with one or more road junctions following the predefined routes, and we test on the medium and hard maps. Two maps named {1o2r\_vs\_4r} and {1o10b\_vs\_1r} from SMAC~\cite{ndq} require the agents to cooperate and communicate to get the position of the enemies. For evaluation, all results are reported with median performance with $95\%$ confidence interval on $5$ random seeds. Details about benchmarks, network architecture and hyper-parameter choices of our method are all presented in Appendices \ref{benchmarkdt}, and \ref{apdx:network_arch}, respectively. 

\subsubsection{Communication Performance}
\label{evaluation of performance}
\begin{figure*}[t!]
  \centering
  \includegraphics[width=0.8\linewidth]{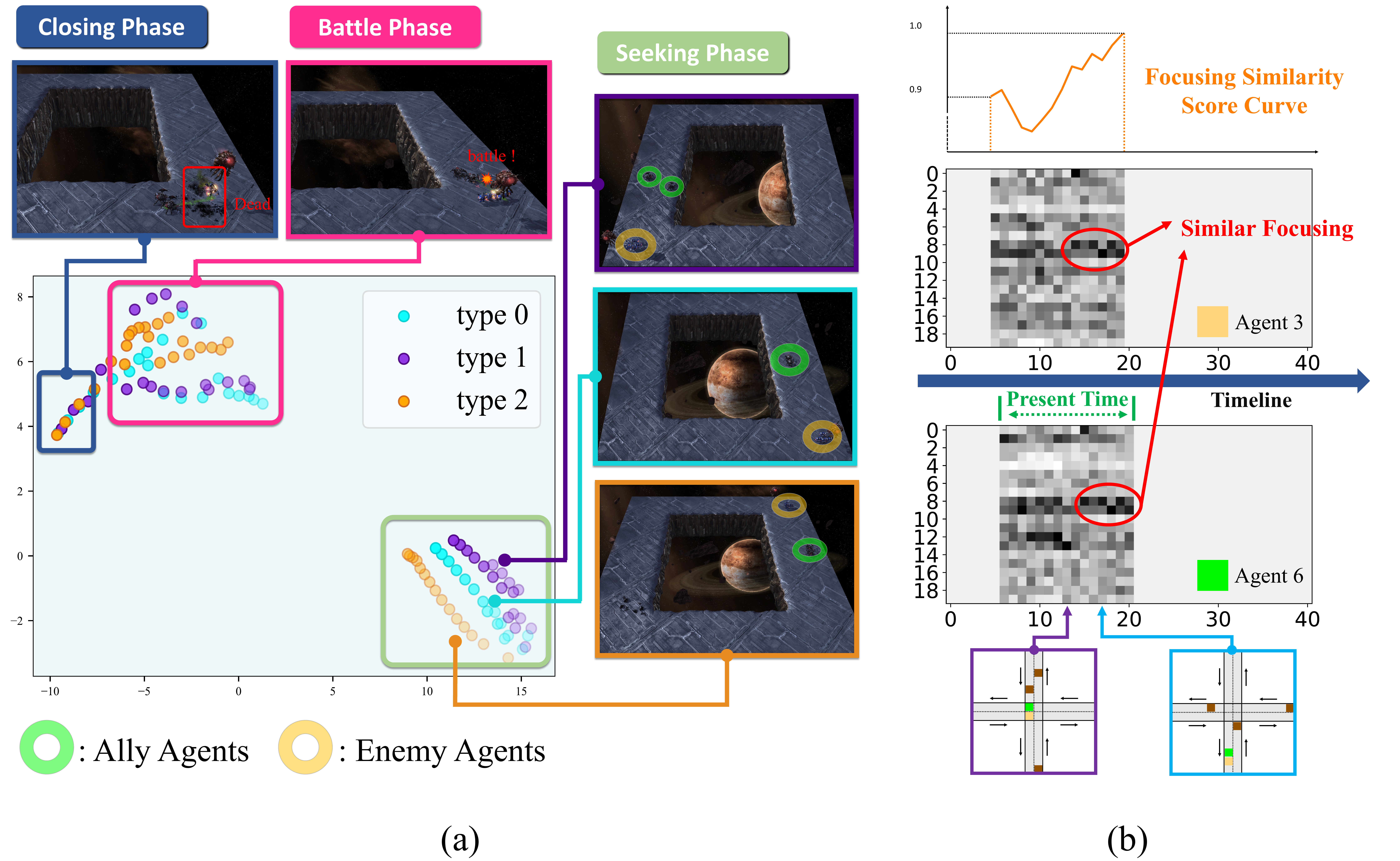}
  \caption{(a) Information aggregation visualization. Each plotted dot represents an aggregated representation. The three different colors respectively represent three different initialization situations for the enemy entities. For example, type $0$ shows the case when the enemies initialize at the lower right corner. To distinguish aggregated representations on different timesteps, we mark larger timesteps by darker shades of the dots. (b) Visualization of variations of selected agents' focus weight $w_i^t$ in a single episode. We use the horizontal axes for timesteps in one single episode and vertical axes for dimensions of the aggregated representation. The weight is reflected through luminance, and the darker the cell, the larger the weight.\remove{\zzz{(a), (b) in the figure look too big.}\solved{Solved: I've scaled this picture down a bit}}
  }
  \label{fig:decomposition}
\end{figure*}
We first compare MASIA against multiple baselines to investigate the communication efficiency on various benchmarks. As illustrated in Figure~\ref{fig:performance}, MASIA achieves the best performance with low variance on all benchmarks, indicating MASIA's strong applicability in scenarios with various difficulties. In Hallway (Figure~\ref{fig:hallway_easy} \& Figure~\ref{fig:hallway_hard}), where frequent communications are required for good performance (method without communication such as QMIX fails), other communication methods such as TarMAC, NDQ, and TMC achieve low performance or even fail in this environment. This indicates that inappropriate message generation or message selection would injure the learning process. We believe the reason why FullComm succeeds is that there is hardly any redundancy in agents' observations in Hallway. Our MASIA also succeeds in this environment, showing superiority over others.
The dominating performance of MASIA is even more significant in extended Hallway (Figure~\ref{fig:hallway_hard}), where agents are separated into different groups. In this environment, MASIA can help agents extract information about their teammates and learn a coordination pattern more efficiently. In LBF (Figure~\ref{fig:lbf_easy} \& Figure~\ref{fig:lbf_hard}), existing communication-based MARL methods like NDQ, Fullcomm, and TarMAC struggle due to the sparsity of rewards, especially when the foods are more sparsely distributed (Figure~\ref{fig:lbf_hard}). In contrast to the performance of QMIX in Hallway, QMIX performs well in LBF, which is attributed to the fact that agents can observe the grids near them, and the mixing network of QMIX can help improve the coordination ability of the fixed group of agents in the training phase. Our method achieves comparable performance with QMIX and TarMAC, showing its strong coordination ability even in sparse reward scenarios. In Traffic Junction (Figure~\ref{fig:traffic_easy} \& Figure~\ref{fig:traffic_hard}), TarMAC and NDQ have high variance due to the instability of their messages, while MASIA gains high sample efficiency and can generate steady messages since it aims to reconstruct the state. On the SMAC benchmarks (Figure~\ref{fig:sc2_easy} \& Figure~\ref{fig:sc2_hard}), we test on two complex scenarios requiring communication to succeed, where one overseer is in active service to get the information of the enemies. Messages are demanded since the agents have limited sight, so other teammates need the overseer's messages to identify the enemies' positions. 
Our method MASIA can maintain the high efficiency of learning and always have competitive performance when converged, which is superior to other baselines.
\subsubsection{Insights into Information Aggregation Encoder}
\label{insight of masia}

To determine what kind of knowledge the encoder has learned through training, we conduct a visualization analysis on the map {1o2r\_vs\_4r} from SMAC to demonstrate the information contained in the aggregation representation $z^t$. We project the aggregation representation vectors into two-dimensional plane by t-SNE~\cite{van2008visualizing} in Figure~\ref{fig:decomposition}a. We take trajectories from $3$ scenarios of different types of initialization under a task where agents have to seek the enemy at the start, discover the enemy, and finally battle with it for better performance. It can be observed that (1) the aggregated representations could be well distinguished by phases. Projected representations in the seeking phase are far from those in the battle phase and closing phase. This implies that our learned representations could well reflect the true states. (2) the aggregated representations are first divergent in the seeking phase when enemies have been initialized, but become increasingly interlaced later until the closing phase, when enemies have been wiped out after a fierce battle. Since enemies are highly related to decision making, such a result verifies that the aggregated representations exploit also reward information. To sum up, the visualization results show that MASIA can extract valuable global information with these representations.

\subsubsection{Study about Individual Information Extraction}
\label{Information Extraction of masia}
To demonstrate the effectiveness of our information extraction module, we analyze the weights computed by the focusing network. Specifically, we select the TJ (medium) task for evaluation and compare the weights produced by two different agents. In this environment, agents could dynamically enter or leave the plane, making the agents staying in the environment flexible through time. It can be observed that agent $3$ and agent $6$ put focus on similar areas of the aggregated representation. Especially after timestep $15$, when agents $3$ and $6$ are in similar situations and distant from the intersection, their focuses are nearly the same. This verifies that the global state information has been successfully extracted to individual agents. Also, on the top of the figure, we draw a figure to measure the relationship between the cosine similarity of weight vectors of different agents against timesteps. It reveals that the similarity of their focus rises after these two agents begin to proceed in the same line (indicated by the render images posted on the lower parts of Figure~\ref{fig:decomposition}b). This also conforms to the intuition that similar messages should be extracted for similar observations. 
\subsubsection{Generality of Our Method}
\label{Generality of masia}
\begin{figure}[htbp]
	\centering
	\begin{subfigure}[t]{0.48\linewidth}
    	\centering
		\includegraphics[width=\linewidth]{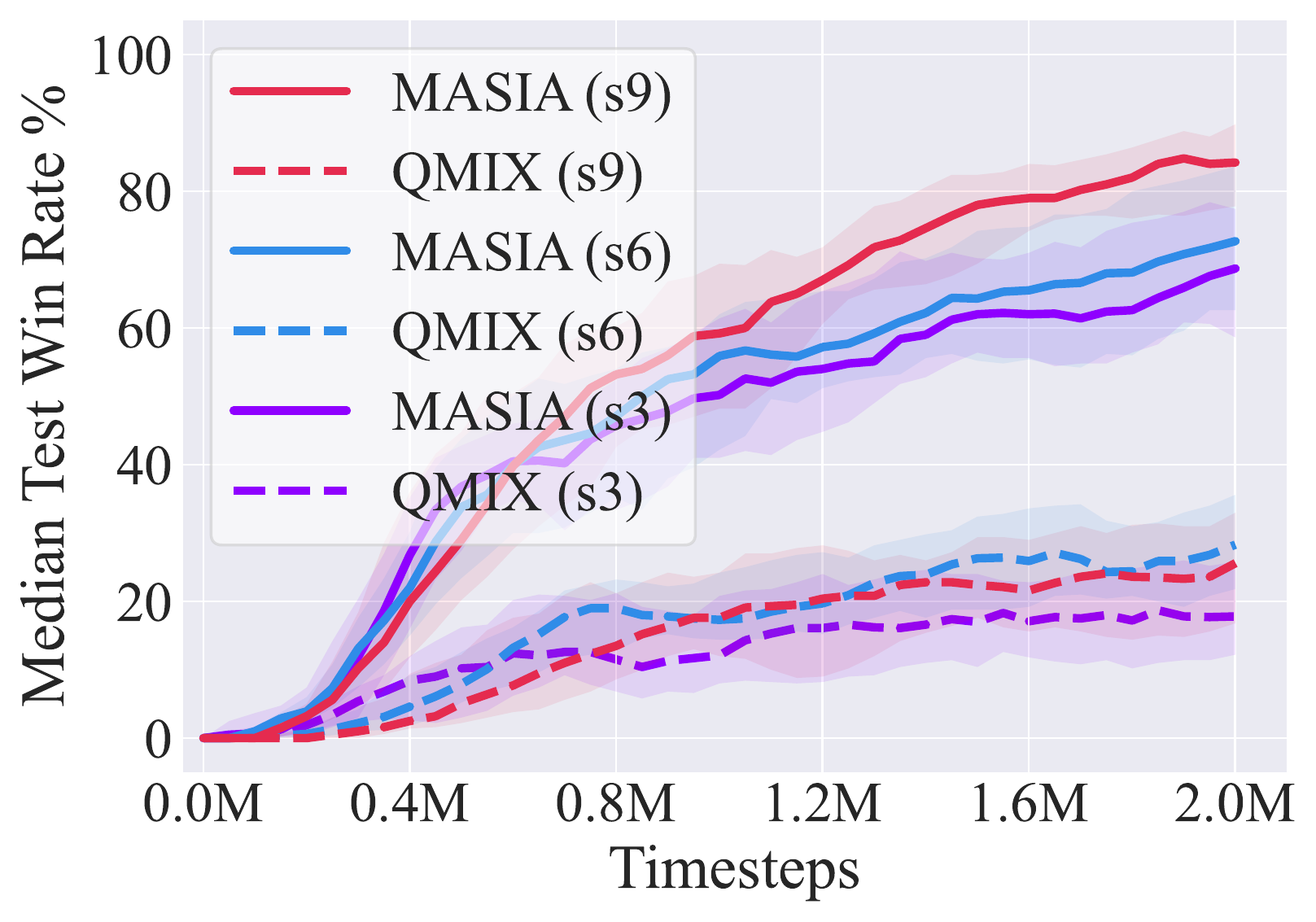}
    \end{subfigure}
    \centering
    \begin{subfigure}[t]{0.48\linewidth}
		\centering
		\includegraphics[width=\linewidth]{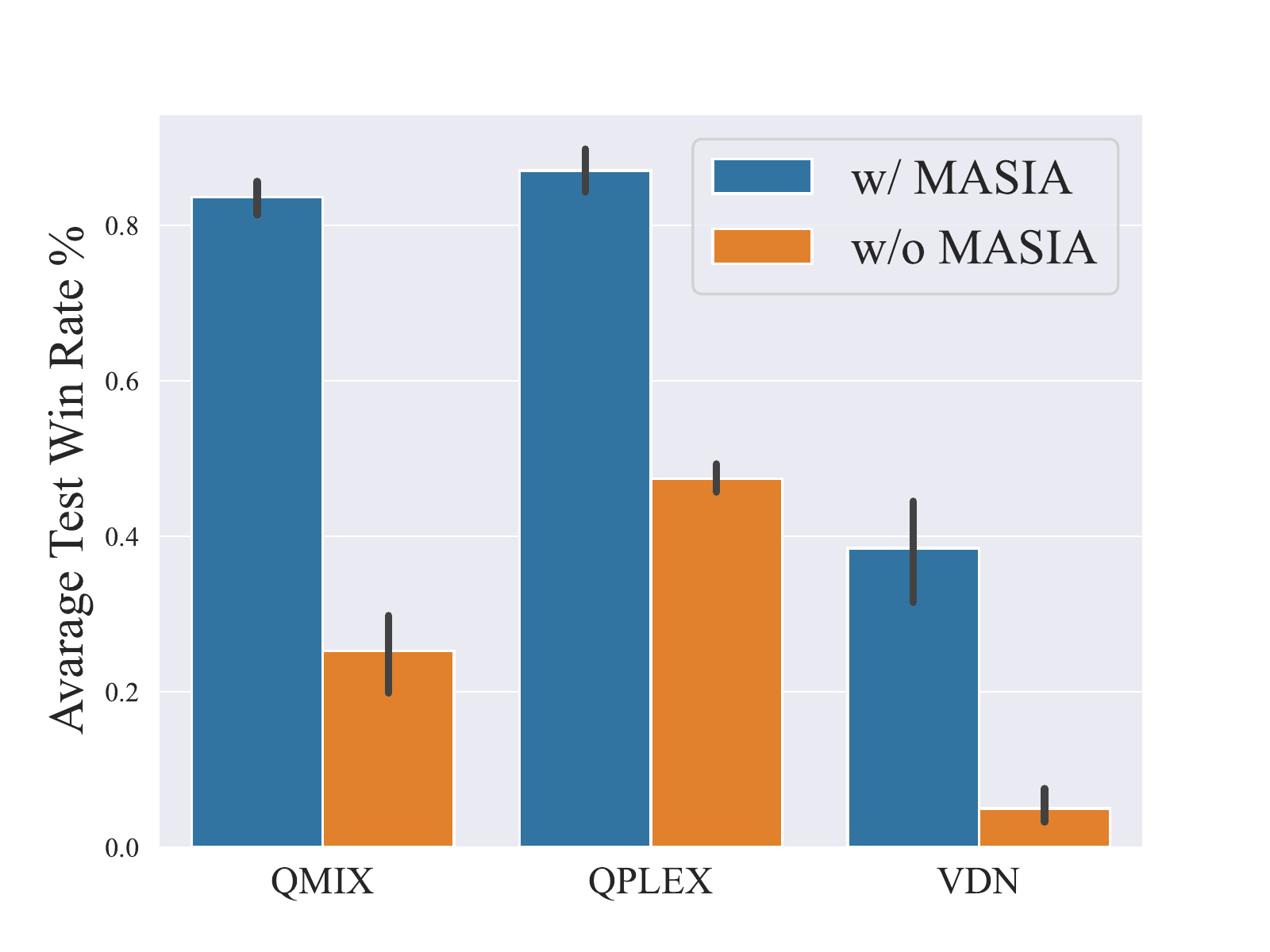}
	\end{subfigure}
    \caption{(a) Performance comparison with varying sights, where s$n$ means the sight range is $n$. (b) The increase of winning rates brought about by MASIA on map 1o10b\_vs\_1r.}
    \label{fig:agnostic}
\end{figure}

\begin{table*}[t]
\centering
\caption{The format of our offline dataset for multi-agent communication.\remove{\zzz{The format seems not an important part that should be shown in detail in the main body. I guess the readers might be not interested in it.} \explain{Explain: We hope that this table will give the reader an intuitive impression of the offline dataset structure in MA.}}}
\label{tab:data_format}
\scalebox{1.0}{
\begin{tabular}{c|c|c|c|c|c|c|c}
\hline
step $0$ & $\cdots$ & \multicolumn{2}{c|}{step $t$} & \multicolumn{2}{c|}{step $t+1$} & $\cdots$ & step $T$ \\
\hline
$s^0$ & $\cdots$ & \multicolumn{2}{c|}{$s^t$} & \multicolumn{2}{c|}{$s^{t+1}$} & $\cdots$ & $s^{T}$\\
\hline
$r^0$ & $\cdots$ & \multicolumn{2}{c|}{$r^t$} & \multicolumn{2}{c|}{$r^{t+1}$} & $\cdots$ & $r^{T}$ \\
\hline
\multirow{4}{*}{agent$_{i=1}^{0}$} & \multirow{4}{*}{$\cdots$} & \multirow{4}{*}{agent$_{i=1}^{t}$} & $o_1^{t}$ & \multirow{4}{*}{agent$_{i=1}^{t+1}$} & $o_1^{t+1}$ & \multirow{4}{*}{$\cdots$} & \multirow{4}{*}{agent$_{i=1}^{T}$} \\
    & & & $a_1^{t}$ & & $a_1^{t+1}$ & \\
    & & & $done_1^{t}$ & & $done_1^{t+1}$ & \\
    & & & ${receiver\ id}_1^{t}$ & & ${receiver\ id}_1^{t+1}$ & \\
\hline
$\cdots$ & $\cdots$ & \multicolumn{2}{c|}{$\cdots$} & \multicolumn{2}{c|}{$\cdots$} & $\cdots$ & $\cdots$ \\
\hline
\multirow{4}{*}{agent$_{i=n}^{0}$} & \multirow{4}{*}{$\cdots$} &\multirow{4}{*}{agent$_{i=n}^{t}$} & $o_n^{t}$ & \multirow{4}{*}{agent$_{i=n}^{t+1}$} & $o_n^{t+1}$ & \multirow{4}{*}{$\cdots$} & \multirow{4}{*}{agent$_{i=n}^{T}$} \\
    & & & $a_n^{t}$&&$a_n^{t+1}$& \\
    & & & $done_n^{t}$&&$done_n^{t+1}$& \\
    & & & ${receiver\ id}_n^{t}$ & & ${receiver\ id}_n^{t+1}$ & \\
\hline
\end{tabular}
}
\end{table*}
We aim to verify that the proposed approach is agnostic to various sight ranges and applied value-based MARL methods. We first conduct experiments on the map {1o10b\_vs\_1r} to show that MASIA could also generalize well on agents with limited observations. The results in Section \ref{evaluation of performance} show the performance of MASIA when the agents have a sight range of $9$. When we narrow the agents' sight ranges, as shown in Figure~\ref{fig:agnostic}a, by receiving and aggregating messages from teammates, the performance of MASIA does not suffer from a significant drop. Our information aggregation and extraction modules prevent the agent from forfeiting knowledge about the state when the sight range is further limited.

To show the generality of the MASIA framework, we also carry out experiments to integrate MASIA with current baselines, including VDN, QMIX, and QPLEX. As illustrated in Figure~\ref{fig:agnostic}b, when integrated with MASIA, the performance of these baselines can be vastly improved on the map {1o10b\_vs\_1r} from SMAC. In this scenario, one overseer is in service to monitor the enemies. Without communication, the other agents have to search the map for the enemies exhaustively. While with reliable communication, they could communicate with each other and the overseer for better coordination. The results demonstrate that MASIA can efficiently aggregate the messages and improve the agents' coordination ability for these value-based MARL methods. 

\subsubsection{Ablation Studies for Online Learning}
\begin{figure}[t]
    \centering
    \includegraphics[width=0.85\linewidth]{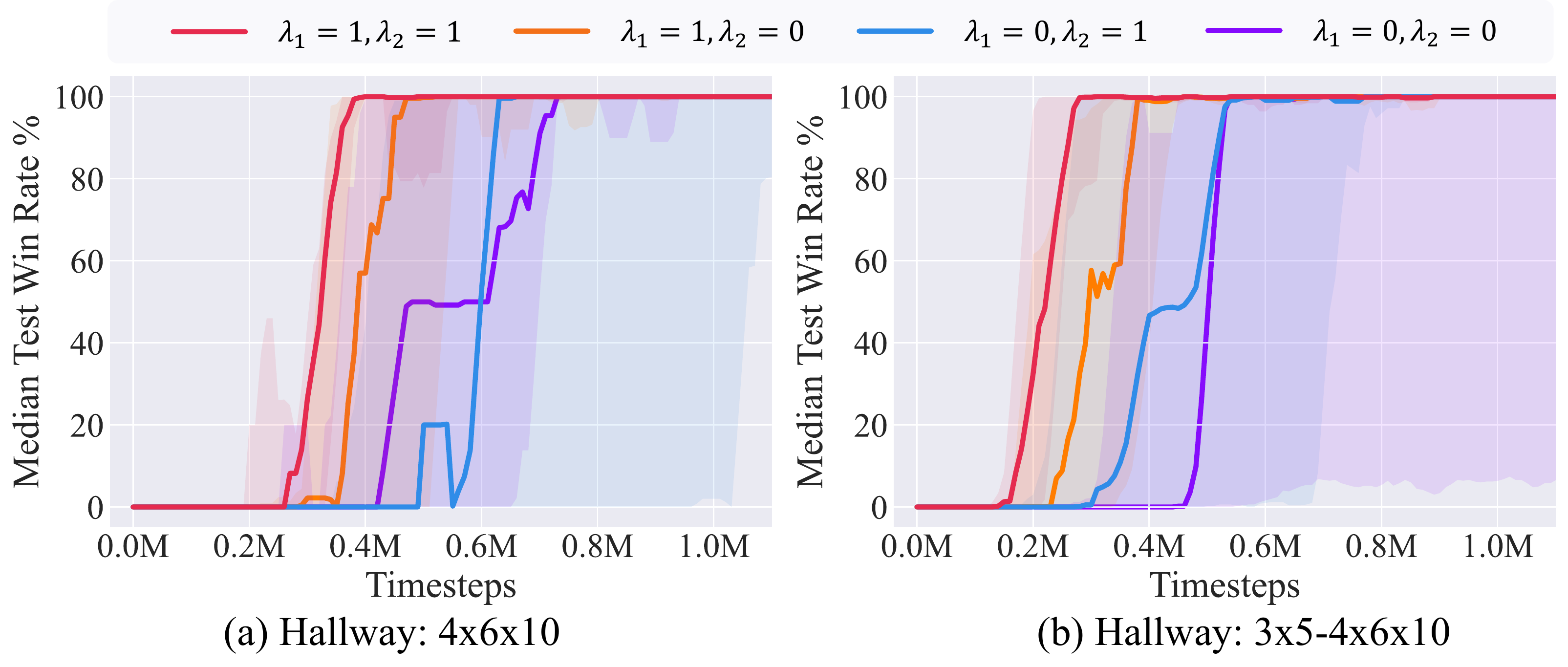}
	\caption{Ablation experiments for online learning in Hallway.}
	\label{fig:ablation hallwayonline}
\end{figure}

In our work, we propose two representation objectives to make the aggregated information representation compact and sufficient. To further justify the effectiveness of these two objectives, we conduct ablation studies for online experiments. Specifically, we design three ablations: (1)$\lambda_1=1,\lambda_2=0$; (2)$\lambda_1=0,\lambda_2=1$; (3)$\lambda_1=0,\lambda_2=0$, which respectively corresponds to (1) only use encoder-decoder learning loss; (2) only use latent model learning loss; (3) neither loss is used. While our method (\name) corresponds to $\lambda_0=1,\lambda_1=1$. The experimental results for the tasks of Hallway are illustrated in Figure~\ref{fig:ablation hallwayonline}.

From the experimental results, we can see that MASIA with two representation objectives outperforms the ablations. The proposed two represetantion objectives help \name~learn to solve the task faster in both settings of 4x6x10 and 3x5-4x6x10. Especially, some random seeds of the third ablation even fail to solve the task 3x5-4x6x10 within 2M samples, which shows the indispensable roles of these two representation objectives. They offer good guides and accelerate the task learning.

\subsection{Offline Multi-Agent Communication Learning}

Currently, in the field of multi-agent communicative reinforcement learning\remove{\zzz{It seems that I didn't see the statement ``multi-agent communication reinforcement learning'', please check it}\solved{Solved: we change into ``multi-agent communicative reinforcement learning``; this expression is adopted in \textit{Mis-spoke or mis-lead: Achieving Robustness in Multi-Agent Communicative Reinforcement Learning}. All expressions in this paper have been unified}}, there is still a lack of appropriate evaluation criteria for offline learning. Thus, to explore the possibility of learning multi-agent communication policies with offline dataset and to test the effectiveness of various multi-agent communication algorithms in the offline setting, we construct a set of offline datasets and conduct experiments on them. 
In specific, to make our constructed dataset as close to the real-world data as possible, we learn from D4RL~\cite{fu2020d4rl} which offers a benchmark for single-agent offline learning, we collect data in a similar way to build our dataset. Different from D4RL, our dataset considers the property of Dec-POMDP, which means that it also contains observations for each agent. Besides, as we focus on multi-agent communication setting, the information of message receivers at each timestep is also included in the dataset. We provide a description of the whole dataset structure in Table~\ref{tab:data_format}. From this table, we can see that each trajectory in the dataset consists of the individual observations and individual actions for each agent. Besides, the variable ${receiver\ id}_i^t$ indicates who can receive the messages sent by agent $i$ at timestep $t$. In specific, the data is typically stored as vectors. For example, ${receiver\ id}_i^t$ may be stored as [0, 1] which indicates that agent 0 and agent 1 can receive the messages sent by agent $i$ at timestep $t$.
\remove{\zzz{To some extent, these fake values are confusing and not helpful to make the manuscript clearer.} \solved{Solved: We remove the fake values and simply take an example in the text.}}
\label{sec:offline_exp}
\begin{table}[t]
\centering
\caption{The range of evaluation return values for the three different levels of behavior policies.} 
\label{tab:replay_policy}
\scalebox{0.95}{
\begin{tabular}{|c|ccc|}
\hline
\diagbox{Environment}{Level}& Good & Medium & Poor \\ 
\hline
Hallway: 4x6x10 &0.75\,-\,1.0 & 0.5\,-\,0.75 & 0.0\,-\,0.5 \\
\hline
Hallway: 3x5-4x6x10 & 1.5\,-\,2.0 & 1.0\,-\,1.5 & 0.0\,-\,1.0 \\
\hline
LBF: 11x11-6p-4f-s1 & 0.75\,-\,1.0 & 0.5\,-\,0.75 & 0.0\,-\,0.5 \\
\hline
LBF: 20x20-10p-6f-s1 & 0.75\,-\,1.0 & 0.5\,-\,0.75 & 0.0\,-\,0.5 \\
\hline
SMAC: 1o2r\_vs\_4r & 15\,-\,20 & 10\,-\,15 & 0\,-\,10 \\
\hline
SMAC: 1o10b\_vs\_1r & 15\,-\,20 & 10\,-\,15 & 0\,-\,10 \\
\hline
TJ: easy & 0\,-\,(-7) & (-7)\,-\,(-17) & (-17)\,-\,(-600) \\
\hline
TJ: medium & 0\,-\,(-7) & (-7)\,-\,(-17) & (-17)\,-\,(-600) \\
\hline
\end{tabular}
}
\end{table}

Considering that the real-world data is typically of greatly varying quality, we design three different dataset settings: 1) \textbf{expert}, 2) \textbf{noisy}, and 3) \textbf{replay}, respectively. These data-collection schemes aim to systematically cover different real-world settings. More details are listed as below:
\begin{itemize}
    \item \textbf{Expert Dataset}. We train an online policy until convergence and greedily sample data with the final expert policy. Such practice is also adopted in Fu et al.~\cite{fu2020d4rl}, Gulcehre et al.~\cite{gulcehre2021regularized} and Kumar et al.~\cite{kumar2020conservative}.
    \item \textbf{Noisy Dataset}. The noisy dataset is generated with an expert policy that selects actions via $\varepsilon$-greedy method with $\varepsilon=0.2$. Creating a dataset from a fixed noisy policy is similar to the dataset collection process in Fujimoto et al.~\cite{fujimoto2019benchmarking}, Kumar et al.~\cite{kumar2020conservative} and Gulcehre et al.~\cite{gulcehre2021regularized}. This scheme aims to simulate the real-world scenarios where experts may occasionally make mistakes.\remove{\zzz{The following three paragraphs look strange. Each paragraph looks too short.} \solved{Solved: we rewrite the section of Replay Dataset.}}
    \item \textbf{Replay Dataset}. This dataset contains the samples that represent different periods of the online policy learning, to which similar scheme can be found in Agarwal et al.~\cite{agarwal2020optimistic} and Fujimoto et al.~\cite{fujimoto2019benchmarking}. To consider offline data of different qualities, we include three forms of \textbf{Replay} dataset. Concretely, we save some intermediate policies during online training, and artificially divide these saved policies into three levels (Poor, Medium and Good) based on the evaluation return values as shown in Table~\ref{tab:replay_policy}. Then we collect three forms of datasets with different mixtures of behavior policies: 1) \textbf{Replay (Poor)} consists entirely of data collected by Poor-Level behavior policies (Poor-Level data); 2) \textbf{Replay (Medium)} consists of 80\% Medium-Level data and 20\% Poor-Level data; (3) \textbf{Replay (Good)} consists of 60\% Good-Level data, 20\% Medium-Level data and 20\% Poor-Level data.
\end{itemize}
 
\begin{table*}[t]
\centering
\caption{Experimental results of different methods under different benchmarks.}
\label{tab:offline}
\scalebox{0.7}{
\begin{tabular}{|c|c|c|c|c|c|c|c|}
\hline
\multicolumn{2}{|c|}{\diagbox{\textbf{Environment/Dataset}}{\textbf{Method}}}&\cellcolor{lightgray}\textcolor[RGB]{250,0,0}{\textbf{ICQ+MASIA}}&\textbf{ICQ+Fullcomm}&\textbf{ICQ+NDQ}&\textbf{ICQ+TarMAC}&\textbf{ICQ+TMC}&\textbf{ICQ}\\ 
\hline  
\multirow{5}{*}{\textbf{Hallway: 4x6x10}}&Expert&\cellcolor{lightgray}\textbf{100.00}\,$\pm$\,\textbf{0.00}\,\,\, &\cellcolor{lightgray}\textbf{100.00}\,$\pm$\,\textbf{0.00}\,\,\,&85.10\,$\pm$\,5.55&97.80\,$\pm$\,4.28&0.04\,$\pm$\,0.24&0.25\,$\pm$\,0.64\\
&Noisy&\cellcolor{lightgray}\textbf{99.90}\,$\pm$\,\textbf{0.07}&99.60\,$\pm$\,0.95&99.50\,$\pm$\,0.74&99.80\,$\pm$\,0.51&14.10\,$\pm$\,11.62&1.55\,$\pm$\,2.82\\
&Replay (Poor)&\cellcolor{lightgray}\textbf{94.50}\,$\pm$\,\textbf{3.59}&87.90\,$\pm$\,8.95&62.20\,$\pm$\,9.21&83.60\,$\pm$\,7.92&0.52\,$\pm$\,0.94&0.11\,$\pm$\,0.35\\
&Replay (Medium)&\cellcolor{lightgray}\textbf{97.40}\,$\pm$\,\textbf{3.60}&94.40\,$\pm$\,5.07&57.40\,$\pm$\,8.24&92.80\,$\pm$\,5.51&1.39\,$\pm$\,1.69&0.12\,$\pm$\,0.40\\
&Replay (Good)&\cellcolor{lightgray}\textbf{99.70}\,$\pm$\,\textbf{0.72}&91.10\,$\pm$\,8.46&11.30\,$\pm$\,13.95&97.70\,$\pm$\,2.59&0.33\,$\pm$\,1.01&0.02\,$\pm$\,0.17\\

\hline 
\multirow{5}{*}{\textbf{Hallway: 3x5-4x6x10}}&Expert&\cellcolor{lightgray}\textbf{99.70}\,$\pm$\,\textbf{0.57}&98.90\,$\pm$\,0.48&86.50\,$\pm$\,5.17&98.50\,$\pm$\,2.26&1.18\,$\pm$\,1.48&0.36\,$\pm$\,0.73\\
&Noisy&\cellcolor{lightgray}\textbf{99.90}\,$\pm$\,\textbf{0.10}&99.90\,$\pm$\,0.20&99.10\,$\pm$\,1.17&99.80\,$\pm$\,1.84&7.67\,$\pm$\,4.43&5.97\,$\pm$\,3.71\\
&Replay (Poor)&\cellcolor{lightgray}\textbf{73.40}\,$\pm$\,\textbf{10.63}&40.80\,$\pm$\,20.46&10.20\,$\pm$\,16.61&53.40\,$\pm$\,26.28&0.00\,$\pm$\,0.00&0.00\,$\pm$\,0.00\\
&Replay (Medium)&\cellcolor{lightgray}\textbf{81.90}\,$\pm$\,\textbf{11.54}&38.60\,$\pm$\,15.92&2.78\,$\pm$\,4.02&35.60\,$\pm$\,24.91&0.00\,$\pm$\,0.00&0.00\,$\pm$\,0.00\\
&Replay (Good)&\cellcolor{lightgray}\textbf{95.20}\,$\pm$\,\textbf{4.02}&77.50\,$\pm$\,13.98&76.50\,$\pm$\,12.65&88.30\,$\pm$\,8.35&0.01\,$\pm$\,0.10&0.06\,$\pm$\,0.46\\

\hline 
\multirow{5}{*}{\textbf{LBF: 11x11-6p-4f-s1}}&Expert&\cellcolor{lightgray}\textbf{88.70}\,$\pm$\,\textbf{13.09}&76.50\,$\pm$\,15.61&82.60\,$\pm$\,4.33&88.30\,$\pm$\,11.37&80.50\,$\pm$\,14.13&82.90\,$\pm$\,12.53\\
&Noisy&\cellcolor{lightgray}\textbf{89.60}\,$\pm$\,\textbf{9.42}&78.40\,$\pm$\,14.59&80.60\,$\pm$\,4.93&88.80\,$\pm$\,10.09&82.10\,$\pm$\,10.94&80.10\,$\pm$\,11.59\\
&Replay (Poor)&77.20\,$\pm$\,15.14&66.40\,$\pm$\,15.53&70.10\,$\pm$\,5.38&\cellcolor{lightgray}\textbf{78.10}\,$\pm$\,\textbf{13.81}&70.90\,$\pm$\,15.14&71.20\,$\pm$\,12.55\\
&Replay (Medium)&\cellcolor{lightgray}\textbf{82.70}\,$\pm$\,\textbf{10.92}&70.70\,$\pm$\,15.93&74.80\,$\pm$\,5.74&80.50\,$\pm$\,11.75&73.60\,$\pm$\,13.45&75.70\,$\pm$\,13.13\\
&Replay (Good)&\cellcolor{lightgray}\textbf{89.10}\,$\pm$\,\textbf{11.41}&76.90\,$\pm$\,13.43&79.30\,$\pm$\,4.60&87.40\,$\pm$\,8.96&78.20\,$\pm$\,12.09&78.30\,$\pm$\,11.92\\

\hline 
\multirow{5}{*}{\textbf{LBF: 20x20-10p-6f-s1}}&Expert&\cellcolor{lightgray}\textbf{54.70}\,$\pm$\,\textbf{10.39}&42.90\,$\pm$\,10.53&45.80\,$\pm$\,4.27&52.50\,$\pm$\,10.83&46.40\,$\pm$\,10.26&44.90\,$\pm$\,10.53\\
&Noisy&54.10\,$\pm$\,10.66&43.30\,$\pm$\,11.89&47.30\,$\pm$\,5.30&\cellcolor{lightgray}\textbf{54.60}\,$\pm$\,\textbf{12.71}&48.70\,$\pm$\,11.72&46.60\,$\pm$\,11.92\\
&Replay (Poor)&\cellcolor{lightgray}\textbf{48.90}\,$\pm$\,\textbf{12.75}&36.80\,$\pm$\,11.49&44.50\,$\pm$\,4.85&44.78\,$\pm$\,10.23&45.10\,$\pm$\,12.29&45.10\,$\pm$\,11.72\\
&Replay (Medium)&\cellcolor{lightgray}\textbf{50.20}\,$\pm$\,\textbf{10.92}&39.90\,$\pm$\,11.04&46.40\,$\pm$\,5.13&48.10\,$\pm$\,11.85&45.70\,$\pm$\,10.45&44.70\,$\pm$\,10.56\\
&Replay (Good)&\cellcolor{lightgray}\textbf{51.50}\,$\pm$\,\textbf{12.78}&41.70\,$\pm$\,10.38&45.60\,$\pm$\,5.50&48.90\,$\pm$\,11.81&46.30\,$\pm$\,10.54&45.20\,$\pm$\,10.19\\

\hline 
\multirow{5}{*}{\textbf{SMAC: 1o2r\_vs\_4r}}&Expert&83.60\,$\pm$\,17.68&\cellcolor{lightgray}\textbf{83.60}\,$\pm$\,\textbf{15.04}&79.30\,$\pm$\,12.71&80.70\,$\pm$\,19.69&45.60\,$\pm$\,25.81& 43.70\,$\pm$\,25.58 \\
&Noisy&\cellcolor{lightgray}\textbf{81.10}\,$\pm$\,\textbf{15.08}&76.70\,$\pm$\,21.21&77.40\,$\pm$\,10.59&79.90\,$\pm$\,18.75&57.90\,$\pm$\,26.88&41.50\,$\pm$\,25.26\\
&Replay (Poor)&\cellcolor{lightgray}\textbf{85.70}\,$\pm$\,\textbf{15.61}&83.60\,$\pm$\,19.86&82.10\,$\pm$\,10.76&84.70\,$\pm$\,15.56&40.30\,$\pm$\,25.85&45.40\,$\pm$\,25.34\\
&Replay (Medium)&79.60\,$\pm$\,18.17&\cellcolor{lightgray}\textbf{80.50}\,$\pm$\,\textbf{20.29}&77.70\,$\pm$\,10.86&79.20\,$\pm$\,18.55&50.20\,$\pm$\,26.39&45.20\,$\pm$\,25.71\\
&Replay (Good)&\cellcolor{lightgray}\textbf{80.10}\,$\pm$\,\textbf{15.26}&75.90\,$\pm$\,23.41&77.90\,$\pm$\,10.78&19.60\,$\pm$\,21.23&48.80\,$\pm$\,26.75&43.80\,$\pm$\,26.68\\

\hline 
\multirow{5}{*}{\textbf{SMAC: 1o10b\_vs\_1r}}&Expert&\cellcolor{lightgray}\textbf{84.10}\,$\pm$\,\textbf{21.36}&77.30\,$\pm$\,21.48&82.80\,$\pm$\,12.41&74.30\,$\pm$\,23.08&11.90\,$\pm$\,16.75&9.34\,$\pm$\,18.37\\
&Noisy&\cellcolor{lightgray}\textbf{84.60}\,$\pm$\,\textbf{17.77}&82.40\,$\pm$\,18.89&78.70\,$\pm$\,10.19&58.20\,$\pm$\,26.63&13.30\,$\pm$\,19.76&15.60\,$\pm$\,18.11\\
&Replay (Poor)&\cellcolor{lightgray}\textbf{82.90}\,$\pm$\,\textbf{17.19}&78.70\,$\pm$\,21.93&77.70\,$\pm$\,10.05&80.90\,$\pm$\,20.91&7.70\,$\pm$\,11.57&9.79\,$\pm$\,18.18\\
&Replay (Medium)&82.90\,$\pm$\,21.26&82.80\,$\pm$\,18.54&81.70\,$\pm$\,8.90&\cellcolor{lightgray}\textbf{85.40}\,$\pm$\,\textbf{16.14}&9.19\,$\pm$\,14.65&7.77\,$\pm$\,13.97\\
&Replay (Good)&\cellcolor{lightgray}\textbf{85.60}\,$\pm$\,\textbf{17.79}&80.90\,$\pm$\,20.71&81.10\,$\pm$\,10.19&80.30\,$\pm$\,23.97&10.30\,$\pm$\,17.21&6.44\,$\pm$\,12.16\\

\hline 
\multirow{5}{*}{\textbf{TJ: easy}}&Expert&\cellcolor{lightgray}\textbf{96.70}\,$\pm$\,\textbf{10.04}&68.60\,$\pm$\,29.36&89.70\,$\pm$\,7.51&96.70\,$\pm$\,12.93&92.00\,$\pm$\,16.31&91.20\,$\pm$\,18.64\\
&Noisy&\cellcolor{lightgray}\textbf{97.90}\,$\pm$\,\textbf{9.93}&93.60\,$\pm$\,12.65&96.20\,$\pm$\,4.39&96.40\,$\pm$\,11.47&95.80\,$\pm$\,10.24&95.10\,$\pm$\,16.29\\
&Replay (Poor)&91.10\,$\pm$\,19.53&75.60\,$\pm$\,27.95&95.30\,$\pm$\,5.22&92.20\,$\pm$\,16.53&90.20\,$\pm$\,20.76&\cellcolor{lightgray}\textbf{95.50}\,$\pm$\,\textbf{13.71}\\
&Replay (Medium)&87.30\,$\pm$\,21.93&71.90\,$\pm$\,28.52&93.60\,$\pm$\,7.89&89.80\,$\pm$\,18.95&\cellcolor{lightgray}\textbf{98.10}\,$\pm$\,\textbf{9.69}&96.80\,$\pm$\,12.41\\
&Replay (Good)&90.80\,$\pm$\,20.92&75.60\,$\pm$\,27.01&89.20\,$\pm$\,12.01&\cellcolor{lightgray}\textbf{96.70}\,$\pm$\,\textbf{10.23}&94.40\,$\pm$\,17.71&96.30\,$\pm$\,13.93\\

\hline 
\multirow{5}{*}{\textbf{TJ: medium}}&Expert&\cellcolor{lightgray}\textbf{88.60}\,$\pm$\,\textbf{19.98}&15.80\,$\pm$\,24.56&78.70\,$\pm$\,14.59&75.50\,$\pm$\,31.17&81.50\,$\pm$\,28.94&87.90\,$\pm$\,22.49\\
&Noisy&72.70\,$\pm$\,29.63&48.20\,$\pm$\,32.49&71.30\,$\pm$\,17.58&56.40\,$\pm$\,34.31&76.10\,$\pm$\,28.35&\cellcolor{lightgray}\textbf{78.10}\,$\pm$\,\textbf{26.14}\\
&Replay (Poor)&\cellcolor{lightgray}\textbf{99.60}\,$\pm$\,\textbf{3.04}&69.90\,$\pm$\,30.91&96.30\,$\pm$\,5.30&98.10\,$\pm$\,6.89&97.50\,$\pm$\,9.80&98.10\,$\pm$\,9.24\\
&Replay (Medium)&89.90\,$\pm$\,19.21&49.20\,$\pm$\,33.36&83.70\,$\pm$\,11.69&87.80\,$\pm$\,21.61&\cellcolor{lightgray}\textbf{91.20}\,$\pm$\,\textbf{17.81}&80.10\,$\pm$\,27.11\\
&Replay (Good)&92.60\,$\pm$\,14.81&72.20\,$\pm$\,30.12&90.30\,$\pm$\,6.89&85.90\,$\pm$\,23.69&\cellcolor{lightgray}\textbf{96.10}\,$\pm$\,\textbf{12.88}&91.10\,$\pm$\,20.63\\

\hline 
\end{tabular}
}
\end{table*}

\subsubsection{Communication Performance}
\begin{figure}[h]
	\centering
	\begin{subfigure}{0.95\linewidth}
	    \centering
	    \includegraphics[width=\linewidth]{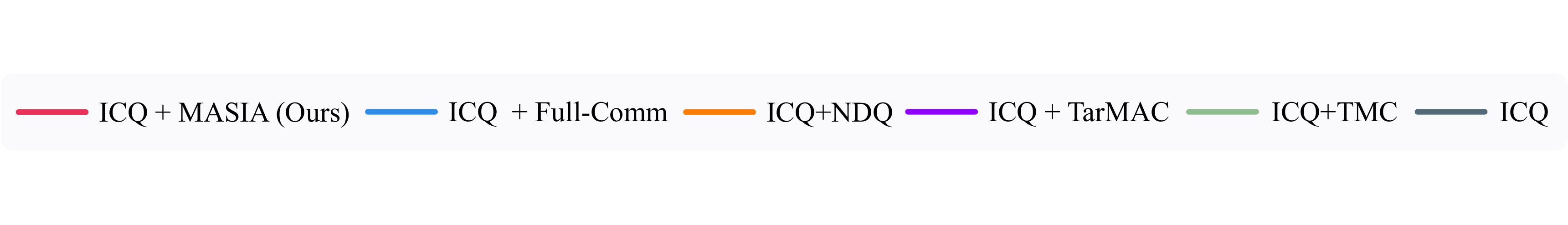}
	\end{subfigure}
	\centering
     	\begin{subfigure}{0.45\linewidth}
     		\centering
     		\includegraphics[width=\linewidth]{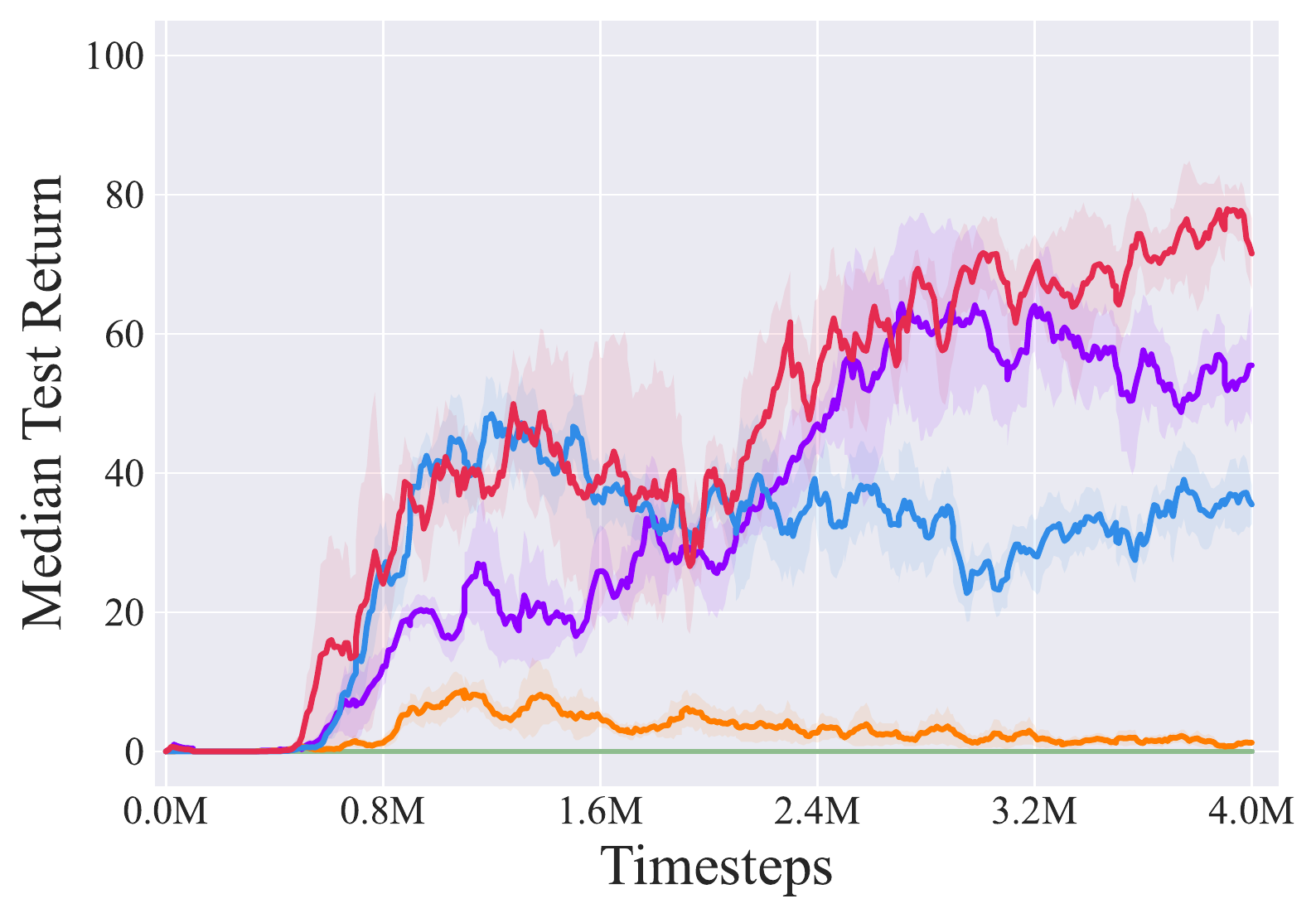}
     	\caption{Replay (Poor)}
     	\end{subfigure}
	\centering
     	\begin{subfigure}{0.45\linewidth}
     		\centering
     		\includegraphics[width=\linewidth]{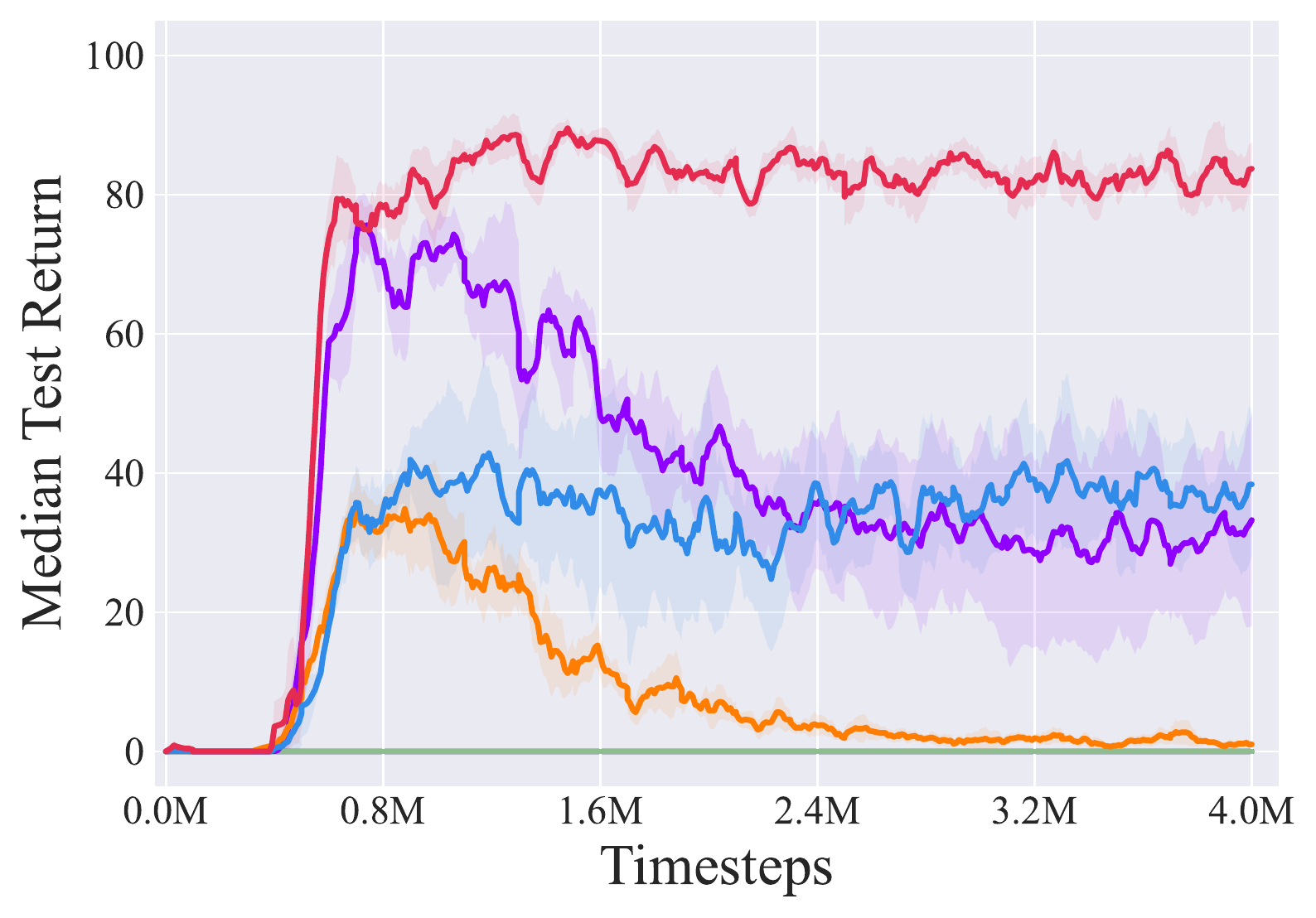}
     	\caption{Replay (Medium)}
     	\end{subfigure}
	\centering
     	\begin{subfigure}{0.45\linewidth}
     		\centering
     		\includegraphics[width=\linewidth]{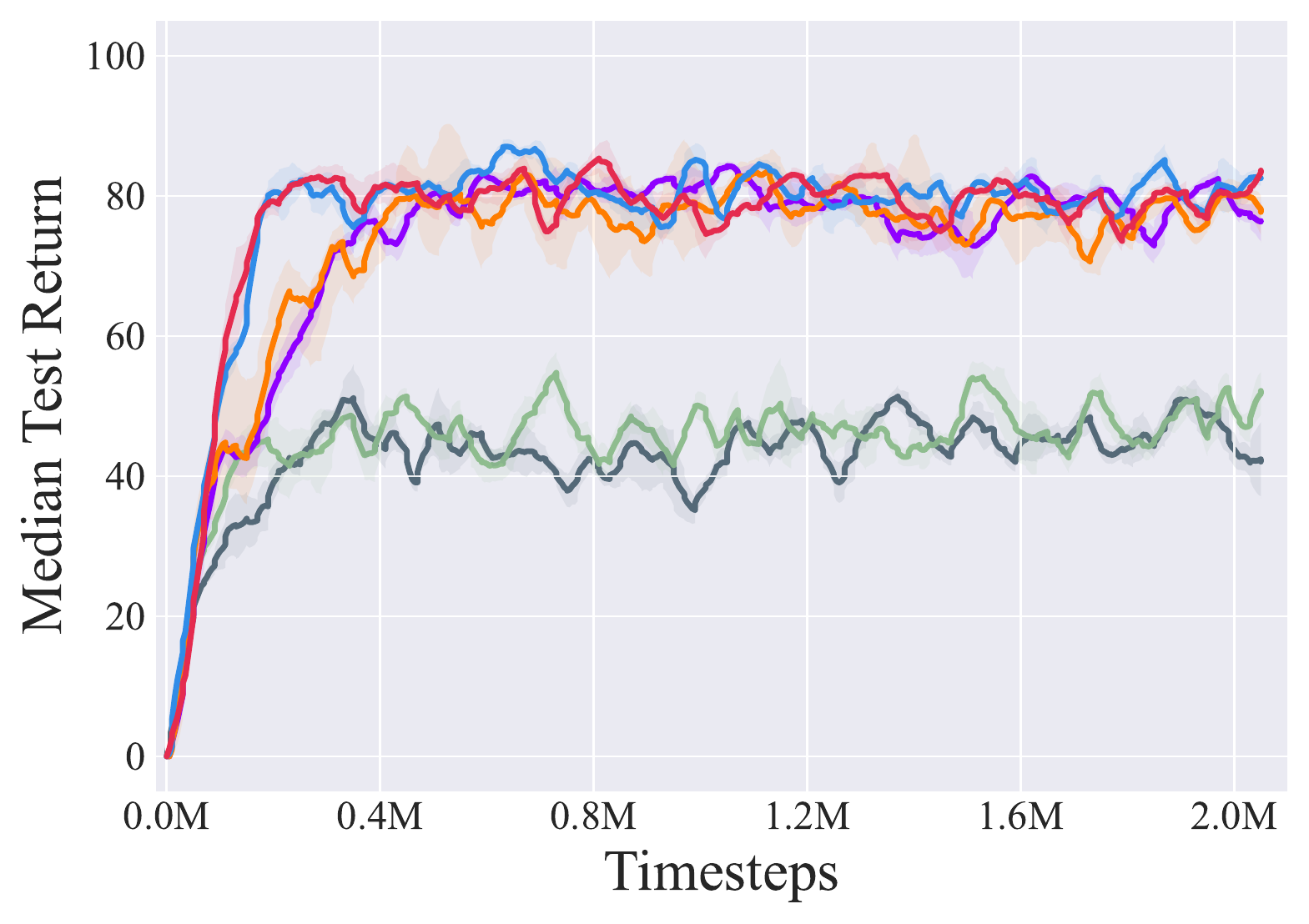}
     	\caption{Replay (Medium)}
     	\end{subfigure}
	\centering
     	\begin{subfigure}{0.45\linewidth}
     		\centering
     		\includegraphics[width=\linewidth]{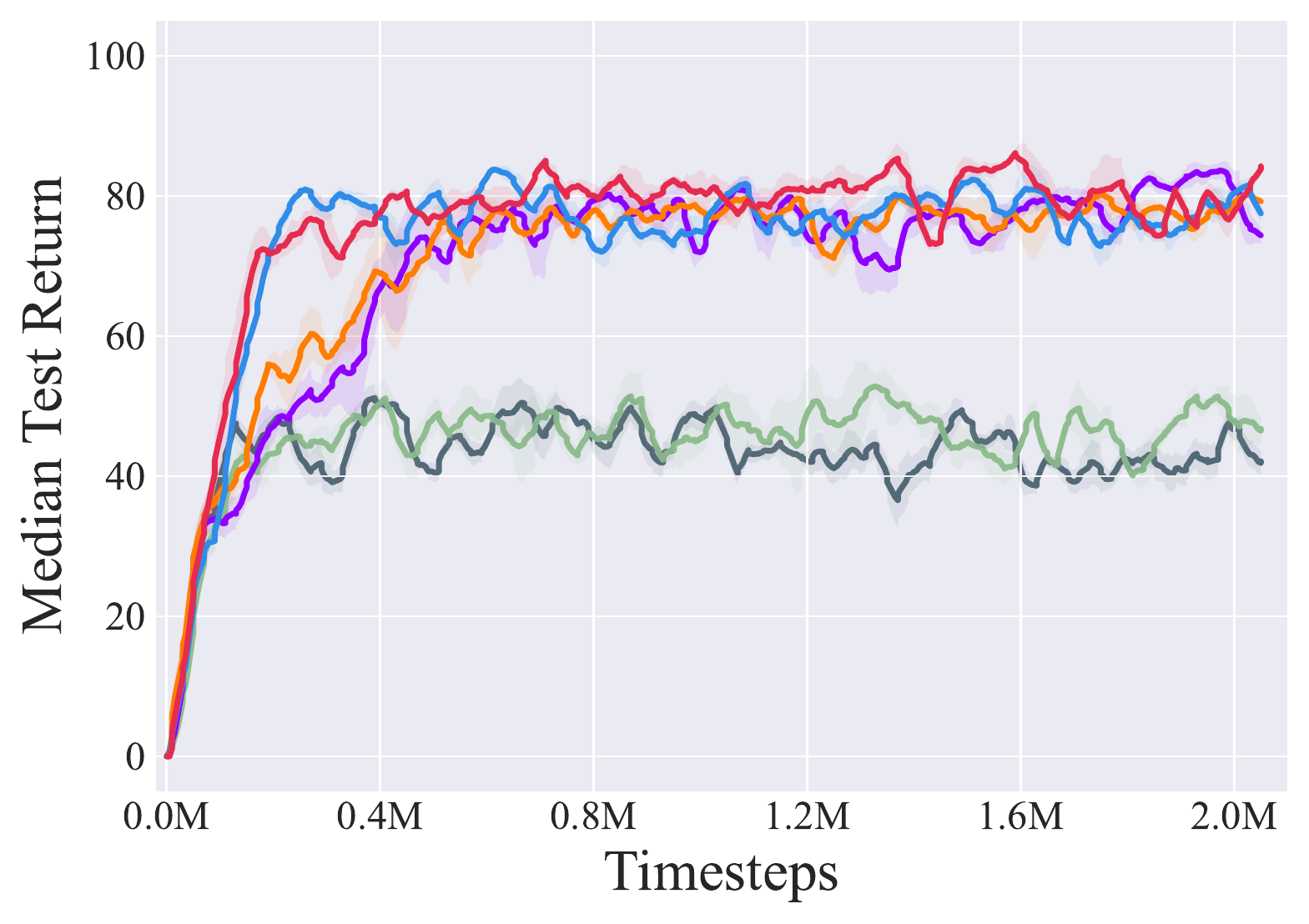}
     	\caption{Replay (Good)}
     	\end{subfigure}
	\caption{Learning curves for offline experiments. (a) and (b) are conducted on Hallway: 3x5-4x6x10, while (c) and (d) are conducted on SMAC: 1o2r\_vs\_4r.}
	\label{fig:offline_exp_curves}
    \vspace{-5mm}
\end{figure}
Considering the Q-divergence problem which is well-studied~\cite{levine2020offline} in offline learning, we propose to combine these communication algorithms with an offline MARL algorithm, ICQ~\cite{icq}, which adopts a conservative learning paradigm to alleviate the Q-divergence issue. Specifically, we integrate the design of different communication algorithms into the actor network structure, thus letting the agents learn message-based policy. That is to say agents make decisions based on communication messages. We also add the experimental results of the ICQ algorithm itself for an ablation, thus to validate the effectiveness of agent communication. The overall results are listed in Table~\ref{tab:offline}. Although Off-Policy Evaluation (OPE)~\cite{voloshin1empirical} is commonly applied in offline RL to test the final policy performance, in order to provide more accurate evaluation, we directly evaluate the policy in the environment and report the evaluation results.
Actually, it is exciting to find that \name~obtains the best communication performance on most tasks under different offline dataset settings. For example, when learning with the \textbf{Expert} offline dataset, \name~achieves the best performance in all scenarios except for SMAC: 1o2r\_vs\_4r. While on the task SMAC: 1o2r\_vs\_4r, \name~achieves as good average performance as Full-Comm with a slightly larger variance. Again taking the task of Hallway: 3x5-4x6x10 as an example, \name~attains the highest success rates on all modes of offline dataset, especially with a huge performance advantage over all other baselines on \textbf{Replay} dataset. The good performance of \name~on \textbf{Replay (Poor)} and \textbf{Replay (Medium)} datasets demonstrates the robustness of \name~as it can still learn good communication policies with low quality data.

To further compare the learning trends of \name~and other baselines on different tasks, we also selectively show the learning curves for offline experiment on part of datasets of Hallway: 3x5-4x6x10 and SMAC: 1o2r\_vs\_4r in Figure~\ref{fig:offline_exp_curves}. As we can see from Figure~\ref{fig:offline_exp_curves}a and \ref{fig:offline_exp_curves}b, \name~exhibits better learning speed and convergence in the learning curves of Hallway: 3x5-4x6x10. \name~is the only method that achieves over 60\% median test success rates within 4M learning samples on \textbf{Replay (Poor)} dataset. Similarly, on \textbf{Replay (Medium)} dataset of Hallway: 3x5-4x6x10, only \name~steadily converges to a success rate of over 80\%, while all other methods fail. Again on SMAC: 1o2r\_vs\_4r, \name~shows faster convergence rate and better convergence performance, on par with Full-Comm algorithm.

\subsubsection{Ablation Studies for Offline Learning}
\label{sec:offline_ablation}
\begin{figure}[htbp]
	\centering
	\begin{subfigure}{0.95\linewidth}
	    \centering
	    \includegraphics[width=\linewidth]{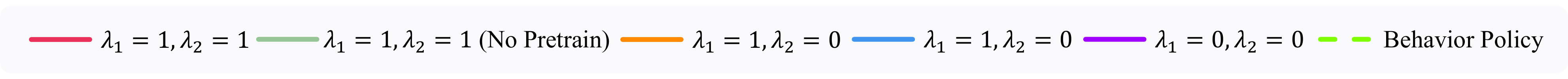}
	\end{subfigure}
	\centering
     	\begin{subfigure}{0.45\linewidth}
     		\centering
     		\includegraphics[width=\linewidth]{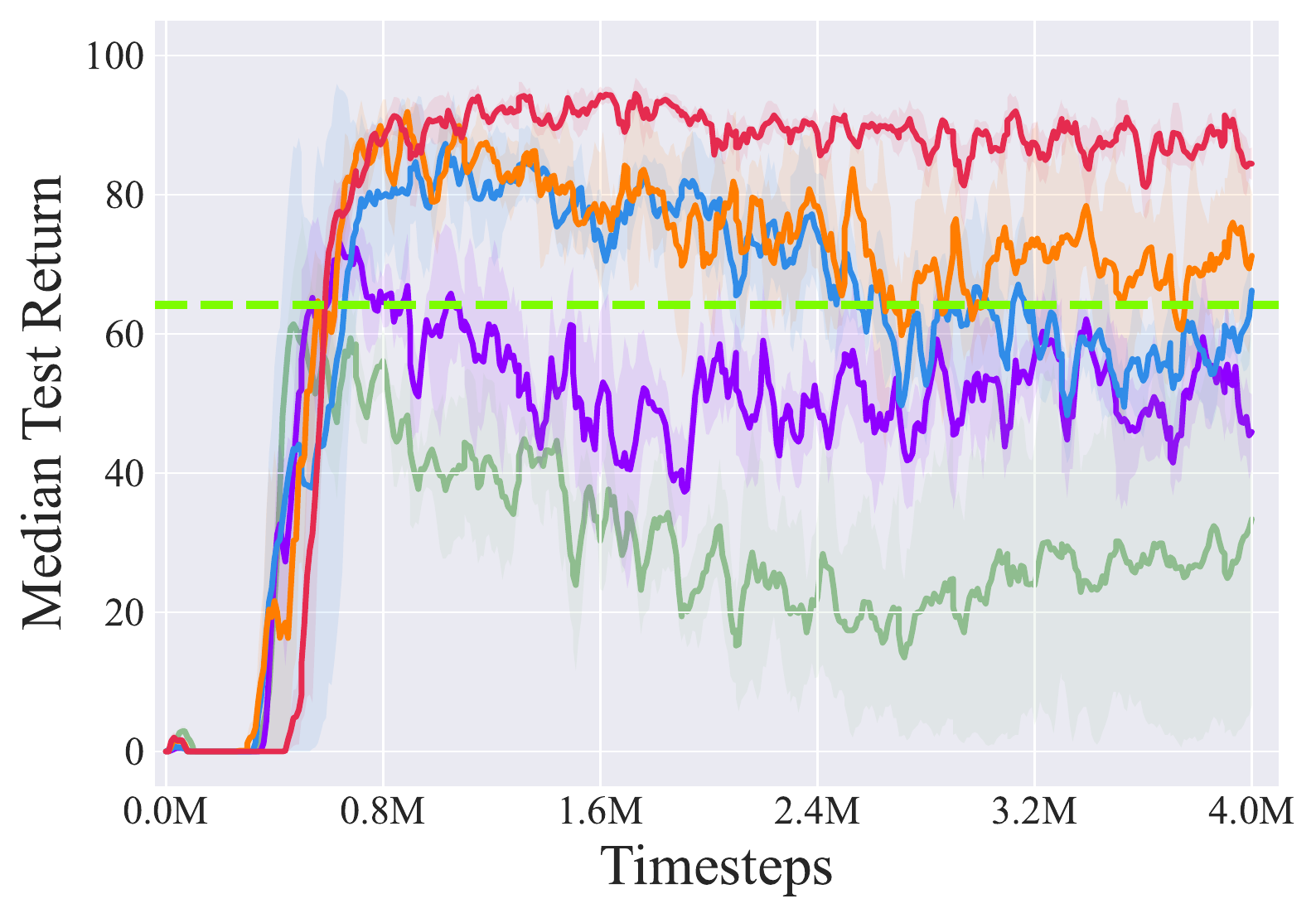}
     	\caption{Replay (Medium) }
     	\label{fig:Hallway(Medium)}
     	\end{subfigure}
	\centering
     	\begin{subfigure}{0.45\linewidth}
     		\centering
     		\includegraphics[width=\linewidth]{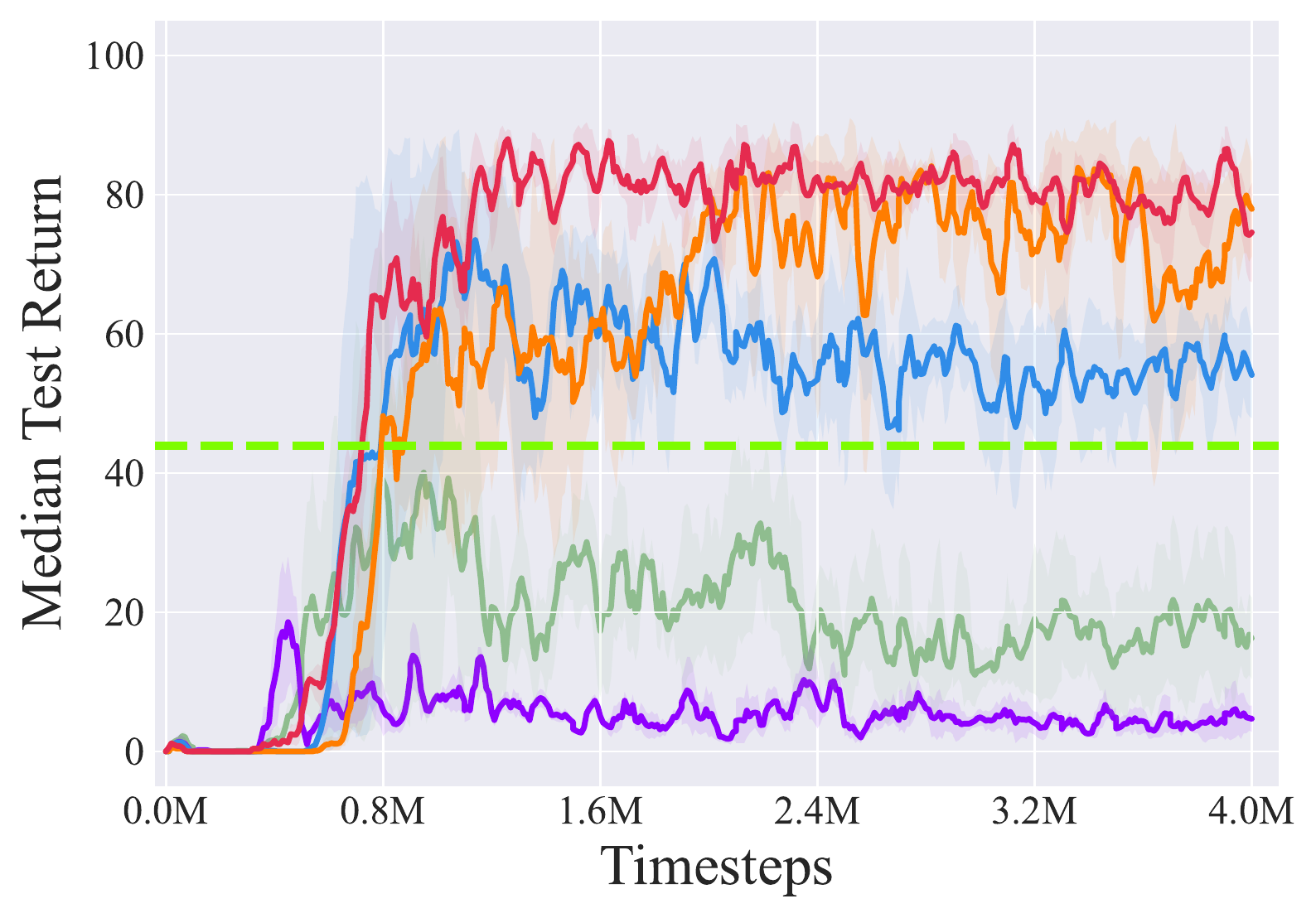}
     	\caption{Replay (Poor) }
     	\label{fig:Hallway(Poor)}
     	\end{subfigure}
	\caption{Ablation experiments for offline learning on the task of Hallway: 3x5-4x6x10. }
	\label{fig:offline_hallway_abations}
\end{figure}
In the previous experiments, we have proved the importance of our proposed two unsupervised representation objectives under online setting. To study how these two objectives work in the offline setting, we further conduct ablation studies for the offline experiments. The ablation results on task of Hallway: 3x5-4x6x10 with dataset \textbf{Replay (Medium)} and \textbf{Replay (Poor)} are depicted in Figure~\ref{fig:offline_hallway_abations}.

Similar to the ablation studies for online experiments, here we ablate the two unsupervised objectives, respectively. From the results we can see that, ablating any of these two objectives will result in a drop in the final communication performance, indicating that the designed two unsupervised objectives play an indispensable role in the offline setting. Actually, it can be seem that the impact of ablating the unsupervised learning objectives in the offline setting is relatively larger than that in the online experiments. The main reason for this phenomenon is that it is more challenging to learn a good communication policy in the offline setting as no exploration in the environment is allowed. Besides, the problem of the dataset quality, especially for dataset \textbf{Replay (Poor)}, strengthens the problem.

Moreover, we also add the ablation experiments for the representation pre-training practice in the offline experiments. From the results we can see that the ablation of the representation pre-training also causes great harm to the performance, which justifies the effectiveness of this practice. More ablation experiments and analyses can be found in Appendix~\ref{apdx:ablation}.

\section{Related Work}
\textbf{Multi-Agent Reinforcement Learning (MARL).} MARL has made prominent progress these years. Having emerged under the CTDE paradigm, many methods are designed to relieve the non-stationarity issue, and have made noticeable progress these years. Most of them can be roughly divided into policy-based and value-based methods. Typical policy gradient-methods involve MADDPG~\cite{maddpg}, COMA \cite{coma}, MAAC \cite{MAAC}, SQDDPG \cite{sqddpg}, FOP \cite{fop}, and  HAPPO~\cite{kuba2021trust} which explore the optimization of multi-agent policy gradient methods, while value-based methods mainly focus on the factorization of the global value function~\cite{wang2021towards,dou2022understanding}. VDN \cite{vdn} applies a simple additive factorization to decompose the joint value function into agent-wise value functions. 
QMIX \cite{qmix2018} structurally enforces the learned joint value function to be monotonic to the agent's utilities, which can represent a more affluent class of value functions. QPLEX \cite{qplex} further takes a duplex dueling network architecture to factorize the joint value function, achieving a full expressiveness power of Individual Global Maximization (IGM) \cite{qtran}. 

\textbf{Multi-agent Communication.} Communication plays a promising role in multi-agent coordination under partial observability~\cite{eccles2019biases,zhu2022survey}. Extensive research have been made on learning communication protocols to improve performance on cooperative tasks \cite{giles2002learning, foerster2016learning, lazaridou2020emergent, xue2021mis, du2021learning, lin2021learning, DBLP:conf/aaai/YuanWZWZ0Z22, ijcai2022p82,guan2022efficient}. Previous works can be divided into two categories. One focuses on generating a meaningful message for the message senders. The simplest way is to treat the raw local observation, or the local information history as message \cite{foerster2016learning, communication16}. VBC~\cite{zhang2019efficient} and TMC~\cite{zhang2020succinct} apply techniques, such as variance-based control and temporal smoothing, in the sender end to make the generated messages meaningful and valuable for policy learning. NDQ \cite{ndq} generates minimized messages for different teammates to learn nearly decomposable value functions, and optimize the message generator based on two different information-theory-based regularizers to achieve expressive communication. 
On the contrary, other works try to learn efficiently to extract the most useful message on the receiver end, and they design mechanisms to differentiate the importance of messages. I2C \cite{i2c} and ACML \cite{acml} employ the gate mechanism to be selective on received messages. There are also works inspired by the broad application of the attention mechanism \cite{chaudhari2019attentive, de2021attention}.
TarMAC \cite{tarmac} achieves targeted communication via a simple signature-based soft-attention mechanism, where the sender broadcasts a key encoding the properties of the agents, then the receiver attends to all received messages for a weighted sum of messages for decision marking.
SARNet \cite{RangwalaW20} and MAGIC \cite{niu2021multi} further remove the signature in TarMAC and leverage attention-based networks to learn efficient and interpretable relations between entities, decide when and with whom to communicate.

\textbf{Offline MARL.} 
Offline reinforcement learning~\cite{levine2020offline} attracts tremendous attention for its data-driven training paradigm without interactions with the environment \cite{prudencio2022survey}. Previous work \cite{bcq} discusses the distribution shift issue in offline learning and considers learning behavior-constrained policies to relieve extrapolation error from unseen data estimations \cite{brac, bear, cql}. Offline MARL is a promising research direction \cite{DBLP:journals/tac/ZhangYLZB21} that trains policies from a static dataset. Following online MARL methods that either extend policy gradient algorithms to multi-agent cases \cite{maddpg, coma, dop} or adopt Q-learning paradigms with value decomposition \cite{vdn, qmix2018, qtran, qplex}, existing offline MARL methods try to exploit offline data with policy constraints. ICQ \cite{icq} effectively alleviates the extrapolation error by only trusting offline data. MABCQ \cite{mabcq} introduces a fully decentralized offline MARL setting and utilizes techniques of value deviation and transition normalization for efficient learning. OMAR \cite{omar} combines first-order policy gradients and zeroth-order optimization methods to avoid the uncoordinated local optima. MADT~\cite{meng2021offline} leverages transformer’s modelling ability of sequence modelling and integrates it seamlessly with both offline and online MARL tasks.
\cite{tian2022learning} investigates offline
MARL with explicit consideration on the diversity of agent-wise trajectories and proposes a novel framework called
Shared Individual Trajectories (SIT) to address this problem. \cite{tseng2022offline} proposes to first train a teacher policy who has the privilege to access every agent’s observations, actions, and rewards. After the teacher policy has identified and recombined the "good" behavior in the dataset, they create separate student policies and distill not only the teacher policy’s features but also its structural relations among different agents’ features to student policies. ODIS~\cite{zhang2023discovering} proposes a novel Offline MARL algorithm to Discover coordInation Skills (ODIS) from multi-task data.
\cite{marlofflinebenchmark} recently releases a framework
Off-the-Grid MARL (OG-MARL) for generating offline MARL datasets and algorithms without communication, which releases an initial set of datasets and baselines
for cooperative offline MARL, along with a standardised evaluation
protocol. 


To the best of our knowledge, none of the existing MARL communication methods explicitly consider how the multiple received messages can be optimized for efficient policy learning. Agents may be confused by redundant information from teammates, and simply augmenting the local policy with the raw message may burden the learning. Meanwhile, there is no testbed for offline multi-agent communication setting. Our proposed method applies a message aggregation module to learn a compact information representation and extracts the most relevant part for decision-making in online and offline settings.

\section{Conclusion and Future Work}
In this paper, we investigate the information representation for multi-agent communication. Previous works either focus on generating meaningful messages or designing a mechanism to select the most relevant message in a raw way, ignoring the aggregation of the message, resulting in low sample efficiency in complex scenarios. Our approach improves communication efficiency by learning a compact information representation to ground the true state and optimizing it in a self-supervised way. Also, we apply a focusing network to extract the most relevant part for decision-making. We conduct sufficient experiments in various benchmarks to verify the efficiency of the proposed methods, both in online and offline settings, and more visualization results further reveal why our approach works. We further release the newly built offline benchmark for multi-agent communication, hoping to facilitate the real-world application of MARL communication.
For future work, more results on image input and solving the scalability issue when facing environments with hundreds or thousands of agents by techniques like agent grouping would be of great interest. Also, how to obtain a robust communication policy when suffering from a distribution shift in online policy deployment is an urgent topic.

\section*{Acknowledgments}
This work is supported by the National Key Research
and Development Program of China (2020AAA0107200), the National Science Foundation of China (61921006, 61876119, 62276126), the Natural Science Foundation of Jiangsu (BK20221442), and the program B for Outstanding
Ph.D. candidate of Nanjing University. We thank Lichao Zhang and Chuneng Fan for their useful support, suggestions, and discussions.

\bibliographystyle{plain}
\bibliography{all_reference}


%
\newpage

\appendix
\section{Details about Benchmarks and Algorithms Involved}
\label{benchmarkdt}
We introduce four types of testing environments as shown in Figure~\ref{fig:env} and Table~\ref{tab:properties}, including Hallway~\cite{ndq}, Level-Based Foraging (LBF)~\cite{papoudakis2021benchmarking}, Traffic Junction (TJ)~\cite{tarmac}, and two maps named 1o2r\_vs\_4r and 1o10b\_vs\_1r requiring communication from StarCraft Multi-Agent Challenge (SMAC)~\cite{ndq}. 
\begin{table*}[!htbp]
\centering
\caption{Properties for each environment.}
\label{tab:properties}
\scalebox{0.75}{
\begin{tabular}{|c|c|c|c|c|c|c|c|c|c|}
\hline
\diagbox{Environment}{Property}& narrow & undirected & multi-task & sparse rewards & suboptimal & realistic & partial observability \\ 
\hline
Hallway: 4x6x10 & \checkmark & \checkmark & & \checkmark & & & \checkmark \\
\hline
Hallway: 3x5-4x6x10 & \checkmark & \checkmark & & \checkmark & & & \checkmark \\
\hline
LBF: 11x11-6p-4f-s1 & \checkmark &  & \checkmark & &\checkmark & & \checkmark \\
\hline
LBF: 20x20-10p-6f-s1 & \checkmark &  & \checkmark & &\checkmark & & \checkmark \\
\hline
SMAC: 1o2r\_vs\_4r & \checkmark &  & & & \checkmark & & \checkmark \\
\hline
SMAC: 1o10b\_vs\_1r & \checkmark &  & & & \checkmark & & \checkmark \\
\hline
TJ: easy & \checkmark &  & & & \checkmark & & \checkmark \\
\hline
TJ: medium & \checkmark &  & & & \checkmark & & \checkmark \\
\hline
\end{tabular}
}
\end{table*}

\textbf{Hallway}:
We design two instances of the Hallway environment. In the first instance, we apply three hallways with lengths of $4, 6$, and $10$, respectively. That means we let three agents $a, b, c$ respectively initialized randomly at states $a_1$ to $a_4$, $b_1$ to $b_6$, and $c_1$ to $c_{10}$, and require them to arrive at state $g$ simultaneously. In the second instance, we divide $5$ agents into two groups. The first group has hallways with lengths of $3$ and $5$, and the second group has hallways with lengths of $4, 6$, and $10$. A reward of $1$ will be given if one group arrives at the goal $g$ simultaneously. However, if both groups reach the goal simultaneously, a penalty of $-0.5$ will be given.

\textbf{Level-Based Foraging (LBF)}:
We use a variant version of the original environment used in~\cite{papoudakis2021benchmarking}, where we define the \texttt{state} to be a data structure that can represent the true global state instead of concatenating the observations of all agents directly. On this basis, we use two environment instances with different configurations, of which one is an $11\times 11$ grid world with $6$ agents, $4$ foods, and the other is a $20\times 20$ grid world with $10$ agents, $6$ foods. In both instances, the observation of agents is a $3\times3$ field of view around it.

\textbf{Traffic Junction (TJ)}:
We use the \textit{medium} and \textit{hard} versions of the Traffic Junction environments. The $medium$ version has an agent number limit of $10$, and the road dimension is $14$, while the $hard$ version has an agent number limit of $20$, and the road dimension is $18$. In both of these two instances, the sight of the agent is limited to $0$, which means each agent can only observe a $1\times 1$ field of view around it.

\textbf{StarCraft Multi-Agent Challenge (SMAC)}:
We use two maps named 1o2r\_vs\_4r and 1o10b\_vs\_1r in SMAC, which are introduced in NDQ~\cite{ndq}. In 1o2r\_vs\_4r, an Overseer finds $4$ Reapers, and the ally units, $2$ Roaches, need to reach enemies and kill them. Similarly, 1o10b\_vs\_1r is a map full of cliffs, where an Overseer detects a Roach, and the randomly spawned ally units, $10$ Banelings, are required to reach and kill the enemy.
\begin{table*}[h]
\small
  \centering
  \caption{Comparison of various algorithms used in this paper.}
  \label{table:comparison of algo}
  \newcommand{\tabincell}[2]{\begin{tabular}[c]{@{}#1@{}}#2\end{tabular}}
  \scalebox{0.95}{
  \begin{tabular}{c|c|c|c}
        \toprule
        Name & \tabincell{c}{Type of communication\\ } & \tabincell{c}{Where to process the information\\ (the Sender/Receiver)} & \tabincell{c}{Matched scenarios} \\
        \midrule
        MASIA(ours) & Full & Receiver & No Restrictions \\ 
        Full-Comm & Full & No & Without redundant information \\
        NDQ & Full & Sender & Value function is nearly decomposable   \\
        TarMAC+QMIX & Full & Receiver & Message with relative importance\\
        TMC & Time-Partial & Sender~\&~Receiver & Message with transmission loss\\
        QMIX & No & No & Full observation or easy coordination \\
        
        \bottomrule
  \end{tabular}
  }
\end{table*}


Several algorithms are involved in our work, including Full-Comm, NDQ~\cite{ndq}, TarMAC~\cite{tarmac}, TMC~\cite{zhang2020succinct}, and QMIX~\cite{qmix2018}. All these methods follow the setting of Dec-POMDP in our experiments, which means that each agent can only have access to its individual partial observation at each timestep. Some algorithms among them (Full-Comm, NDQ, TarMAC, and TMC) do communication to base each agent's decision-making on richer information, while QMIX has no communication and always let agents make decisions based on their local observations (or observation histories). In Table~\ref{table:comparison of algo}, we offer a comparison between these baselines and our method from different dimensions.

\section{Implementation Details}
\label{implementationdt}

\subsection{Network Architecture and Hyper-parameters}
\label{apdx:network_arch}

\begin{figure}[t]
    \centering
    \includegraphics[width=0.8\linewidth]{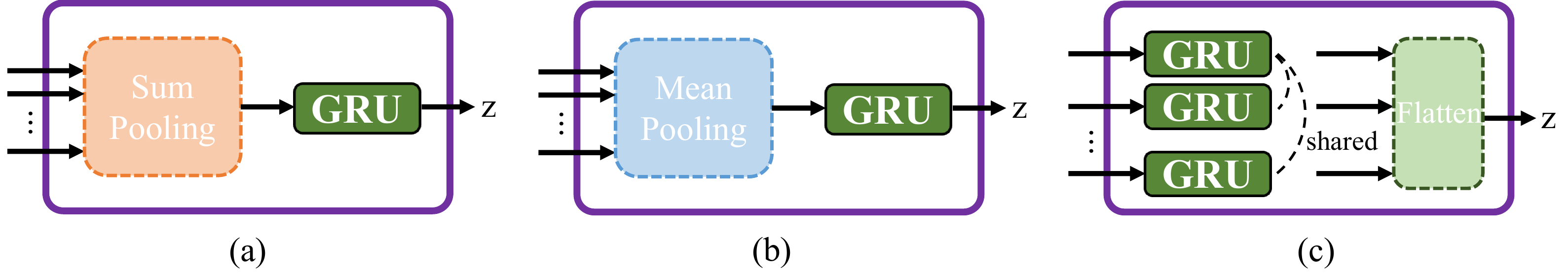}
	\caption{Three versions of implementation for integration network. The dotted lines in (c) indicate that these GRU networks share the same parameters.}
	\label{fig:integration network}
\end{figure}

\textbf{Integration Network.} The Information Aggregation Encoder (IAE) in our approach consists of a self-attention network and an integration network. Here we describe the details of the integration network. We introduce $3$ different kinds of integration networks as shown in Figure~\ref{fig:integration network}.

Among these three versions of integration networks, $(a)$ and $(b)$ apply \textit{mean} and \textit{sum} pooling operations, respectively, on the output vectors of the self-attention network, while in the third version $(c)$, we let the self-attention network's output vectors pass through a shared GRU network separately and finally flatten them as the obtained aggregation representation. In $(a)$ and $(b)$, no matter how we permute the agents, we will always obtain the same aggregation representation. In $(c)$, the permutation of the agents will only affect the order of some dimensions of the aggregation representation instead of obtaining a totally different representation in some vanilla designs, such as networks with Multi-Layer Perceptions. In practice, we find that design $(c)$ achieves the best performance among these three integration networks, and all the experiment results shown in our paper are based on this implementation. 

\textbf{Hyper-parameters.} Our implementation of MASIA is based on the EPyMARL\footnote{\url{https://github.com/uoe-agents/epymarl}} \cite{samvelyan2019starcraft} with StarCraft 2.4.6.2.69232 and uses its default hyper-parameter settings. For example, we apply the default $\epsilon$-greedy action selection algorithm to each method, which means $\epsilon$ decays from $1$ to $0.05$ in $50$K timesteps. The selection of the additional parameters introduced in our approach is listed in Table~\ref{table:hyper-parameters}. We use this set of parameters in all experiments shown in this paper except for the ablations.
\begin{table}[h]
  \centering
  \caption{Hyper-parameters in experiments.}
  \label{table:hyper-parameters}
  \scalebox{0.88}{
  \begin{tabular}{c|c}
        \toprule
        name & value \\
        \midrule
        hidden dimension for query and key in self-attention module & 16 \\
        output dimension of self-attention module & 32 \\
        $\lambda_3$ (whether to warm up the representation learning) & 1 \\
        $\lambda_2$ (coefficient of latent model learning loss) & 1 \\
        $\lambda_1$ (coefficient of encoder-decoder learning loss) & 1 \\
        action embedding dimension in latent model & 8 \\
        dimension for each agent in the aggregation representation & 8 \\
        observation embedding dimension before concatenated with $z$ & 32 \\
        whether to predict the residuals of the next state & True \\
        prediction length K in latent model & 2 \\
        hidden dimension for hidden states in latent model & 64 \\
        \bottomrule
  \end{tabular}
  }
\end{table}
  
{\begin{algorithm}[htbp]
	\caption{Training Framework}
	\label{alg:training flow}
	\begin{algorithmic}[1]
	    \STATE Initialize replay buffer $\buffer$.
	    \STATE Initialize information aggregation encoder with random parameters $\theta$ and state prediction decoder with random parameters $\eta$.
	    \STATE Initialize $Q$ network with random parameters $\phi$ and latent model with random parameters $\psi$.
	    \STATE Initialize parameters of target encoder $\theta^{-} = \theta$, and target $Q$ network $\phi^{-} = \phi$.
	    
	    \FOR{$\mbox{episode}=1$ to $M$}
            \STATE Roll out one trajectory $\bm{\tau}$ with $\epsilon$-greedy policy in the environment.
            \STATE Store the trajectory $\bm{\tau}$ in $\buffer$.
	        \IF{$|\buffer|$ is larger than batch size $m$}
	            \STATE Sample a minibatch $\batch$ of $m$ trajectories from $\buffer$.
                \STATE Compute the encoder-decoder loss:
                $$
	                \loss_{ae}(\theta, \eta) = \sum_{\rm traj\in\batch}\sum_{t=1}^T \|g_\eta(z^t) - s^t\|_2^2,~z^t = f_\theta(\bm{o}^t).
	            $$
	            \STATE Compute the latent model loss:
	            \begin{align*}
	                &\loss_{m}(\theta, \psi) = 
	                \sum_{\rm traj\in\batch}\sum_{t=1}^{T-K} \sum_{k=1}^K\|\hat{z}^{t+k} - z^{t+k}\|_2^2,~\\
	               &\hat{z}^{k+1} = h_{\psi}(z^t, \bm{a}^t),~\\
                    &\hat{z}^{t+k}=h_{\psi}(\hat{z}^{t+k-1}, \bm{a}^{t+k-1}), k=2,\dots,K,~\\
	                &z^{t+k} = f_{\theta^{-}}(\bm{o}^{t+k}), k=0,\dots,K,
	            \end{align*}
	            \STATE Compute the reinforcement learning loss:
	                $$
                    \begin{aligned}
	                    &\loss_{rl}(\theta, \phi) =
	                    \sum_{\rm traj\in\batch}\sum_{t=1}^{T-1} \Big(r + \\ &\gamma\max_{\bm{a}^{\boldsymbol{\prime}}}Q_{\rm tot}(\bm{\tau}^{t+1},\bm{a}^{\boldsymbol{\prime}};\theta^{-}, \phi^{-})-Q_{\rm tot}(\bm{\tau}^t, \bm{a}^t;\theta, \phi)\Big)^2.
	            \end{aligned}   
                   $$
	            \STATE Update $\theta, \eta, \phi, \psi$ by minimizing $\loss_{rl}(\theta, \phi) + \lambda_1 \loss_{ae}(\theta, \eta) + \lambda_2 \loss_{m}(\theta, \psi)$.
	        \ENDIF
	        \STATE Update target network parameters $\theta^{-}, \phi^{-}$ with the EMA method. 
	    \ENDFOR 
    \end{algorithmic}
\end{algorithm}

\subsection{The Overall Flow for Training and Testing}
\label{apdx:overall flow}
Firstly, to offer a direct impression of our work, we first talk about how agents behave differently during decentralized execution vs.
centralized training in our method (MASIA). The main differences are that we would use state information to help compute the auto-encoder loss and estimate the $Q_{\rm{tot}}$ (if there is a mixing network in the $Q$-network, e.g. MASIA+QMIX) during training phase. Besides, multi-step prediction loss is also only computed and optimized in the training phase. During decentralized execution, the state decoder, the latent model, and the possible mixing network are all thrown away. We remain the IAE, the focusing network, and individual Q-networks to ensure the decentralized execution process. The representation loss terms we designed are aimed at training the encoder network well.

To illustrate the process of training, the overall training flow of MASIA is shown in Algorithm~\ref{alg:training flow}. Lines 5$\sim$16 express the whole training process, where we apply an off-policy learning algorithm and iteratively update the parameters of the model. Specifically, we compute encoder-decoder loss, latent model loss and temporal difference loss in Lines 10, 11 and 12, respectively, and do parameter updating together in Line 13.
Besides, the execution flow of MASIA is shown in Algorithm~\ref{alg:execution flow}. In the execution phase, the agents first broadcast their observations to each other, and then each agent calculates its own action $a_i^t$ by applying the focusing network and individual $Q$ network learned during training, which is described in Lines 4 and 5.
\begin{algorithm}[htbp]
	\caption{Execution Flow}
	\label{alg:execution flow}
	\begin{algorithmic}[1]
	    \REQUIRE information aggregation encoder with parameters $\theta$, individual $Q$ network with parameters $\phi_i$ for each $Q_i$, focusing networks with parameter $\omega_i$ for each agent and agent number $n$.

	    \FOR{$\mbox{step}=0$ to $\mbox{episode\_limit}$}
	        \STATE Each agent broadcasts its observational information $o_i^t$ at timestep $t$, and then each agent feed collected observations $\bm{o}^t=\{o_i^t\}_n$ into the information aggregation network, obtaining $z^t=f_\theta(\bm{o}^t)$.
	        \FOR{$i=1$ to $n$}
	            \STATE Agent $i$ calculates $w_i^t=F_{w_i}(o_i^t)$ by using the focusing network, and obtains extracted information $\bar{z}_i^t=w_i^t\cdot z^t$.
	            \STATE The $o_i^t$ and $\bar{z}_i^t$ are fed into the individual $Q$ network to calculate $Q_i(\tau_i^t, \cdot)$, and agent $i$ gets action $a_i^t=\mathop{\arg\max}_{a}Q_i(\tau_i^t, a)$.
	        \ENDFOR
	        \STATE The agent system interacts with the environment by executing actions $\{a_i^t\}_n$.
	    \ENDFOR 
    \end{algorithmic}
\end{algorithm}}
\subsection{Experimental Details}
Our experiments were performed on a desktop machine with 4 NVIDIA GTX 3090 GPUs. For all the performance curves in our paper, we pause training every $M$ timesteps and evaluate for $N$ episodes with decentralized greedy action selection. The $(M, N)$ in Hallway~\cite{ndq}, Level-Based Foraging~\cite{papoudakis2021benchmarking}, Traffic Junction~\cite{tarmac}, and SMAC~\cite{ndq} are $(10\rm{K}, 100), (50\rm{K}, 100), (10\rm{K}, 40)$, and $(50\rm{K}, 100)$, respectively. We evaluate the test win rate, the percentage of episodes in which the agents win the game within the time limit in $N$ testing episodes for all tasks.

\section{More Results about offline learning}
\subsection{Details about Offline Dataset}
Due to the huge cost and limitations of interacting with the environment in the real world, offline learning has been seen as a promising direction for getting reinforcement learning off the ground. Besides, since many real-world problems can be modeled as multi-agent systems, especially multi-agent communication systems, advances in research concerning offline communication learning are urgently needed. However, in the field of multi-agent communicative reinforcement learning, there is still no good evaluation criteria for offline learning, which largely hinders the development of this field. Considering this situation, we construct an offline dataset, which involves multiple multi-agent communication environments and different dataset settings, to support the research of offline multi-agent communication learning.

In specific, our dataset involves four environments, each including two scenarios, for a total of eight scenarios. The included environments are consistent with the task environments in the online experiments, which are shown in Figure~\ref{fig:env}. As we analyzed in the experimental section (Section~\ref{sec:exp}), these environments represent communication scenarios with different characteristics. To offer more insights of these four environments in the offline setting, we additionally provide a description of these environments in Table~\ref{tab:properties} with reference to several dataset properties proposed in D4RL~\cite{fu2020d4rl}. Among these properties, all environments are \textbf{narrow}, not \textbf{realistic} and of \textbf{partial observability}. The latter two are because the data are all collected in the simulator without real human data and the agents' sight ranges  are limited due to the communication setting.
The data for Hallway environment is \textbf{undirected} and of \textbf{sparse rewards} because the agents only obtain rewards when they reach target together and the data can be stitched in this environment. Besides, it is the only environment that the expert policy can generate optimal trajectories. In addition, the data for LBF environment is \textbf{multi-task} because the locations of the target fruits are randomly generated at the beginning of each episode, thus it can be seem as including data of multiple navigation tasks.

\begin{table*}[t!]
\centering\small
\caption{Properties of different offline datasets.}
\label{tab:data_qualities}
\begin{tabular}{|c|c|c|c|c|}
\hline
\textbf{Environment Scenarios} & \textbf{Dataset} & \textbf{Behavior Policy Test Success\%} & \textbf{\# Samples} & \textbf{Return Distribution}\\
\hline
    \multirow{5}{*}{\textbf{Hallway: 4x6x10}} & Expert & 100.00\,$\pm$\,0.00 & & 1.00\\
        &Noisy&100.00\,$\pm$\,0.00 & &0.40\\
        &Replay (Poor)&32.90\,$\pm$\,47.01 & 12000& 0.33\\
        &Replay (Medium)&62.50\,$\pm$\,48.41 & &0.62\\
        &Replay (Good)&87.50\,$\pm$\,33.07& &0.88\\
\hline
    \multirow{5}{*}{\textbf{Hallway: 3x5-4x6x10}}&Expert&99.80\,$\pm$\,0.25& & 2.00\\
        &Noisy&99.80\,$\pm$\,0.25& & 1.01\\
        &Replay (Poor)&43.90\,$\pm$\,33.31& 12000& 0.88\\
        &Replay (Medium)&64.20\,$\pm$\,30.41& &1.28\\
        &Replay (Good)&82.90\,$\pm$\,27.74& &1.66\\
\hline
    \multirow{5}{*}{\textbf{LBF: 11x11-6p-4f-s1}}&Expert&90.20\,$\pm$\,2.59& & 0.91\\
        &Noisy&90.20\,$\pm$\,2.59& &0.83\\
        &Replay (Poor)&46.60 \,$\pm$\,14.93& 12000& 0.48\\
        &Replay (Medium)&67.40\,$\pm$\,13.36& &0.67\\
        &Replay (Good)&83.60\,$\pm$\,22.71& &0.84\\
\hline
    \multirow{5}{*}{\textbf{20x20-10p-6f-s1}}&Expert&53.80\,$\pm$\,2.34& & 0.54\\
        &Noisy&53.80\,$\pm$\,2.34& &0.45\\
        &Replay (Poor)&38.70\,$\pm$\,14.55 & 12000& 0.41\\
        &Replay (Medium)&45.50\,$\pm$\,15.75& &0.65\\
        &Replay (Good)&49.10\,$\pm$\,22.67& &0.78\\
\hline
    \multirow{5}{*}{\textbf{SMAC: 1o2r\_vs\_4r}}&Expert&81.50\,$\pm$\,5.72& & 18.25\\
        &Noisy&81.50\,$\pm$\,5.72 & &14.71\\
        &Replay (Poor)&45.70\,$\pm$\,19.25 & 12000& 9.14\\
        &Replay (Medium)&63.80\,$\pm$\,29.03& &12.78\\
        &Replay (Good)&76.60\,$\pm$\,27.69& &16.53\\
\hline
    \multirow{5}{*}{\textbf{SMAC: 1o10b\_vs\_1r}}&Expert&82.80\,$\pm$\,2.66& & 18.52\\
        &Noisy&82.80\,$\pm$\,2.66& &12.27\\
        &Replay (Poor)&48.20\,$\pm$\,31.48 & 12000& 9.62\\
        &Replay (Medium)&67.30\,$\pm$\,34.65& &14.63\\
        &Replay (Good)&78.70\,$\pm$\,30.43& &16.80\\
\hline
    \multirow{5}{*}{\textbf{TJ: easy}}&Expert& 98.70\,$\pm$\,1.25& & -3.28\\
        &Noisy&98.70\,$\pm$\,1.25& &-13.48\\
        &Replay (Poor)&45.60\,$\pm$\,17.28& 10000& -25.27\\
        &Replay (Medium)&57.80\,$\pm$\,16.57& &-20.61\\
        &Replay (Good)&73.90\,$\pm$\,16.98& &-9.58\\
\hline
    \multirow{5}{*}{\textbf{TJ: medium}}&Expert& 94.30\,$\pm$\,1.08& & -7.08\\
        &Noisy&94.30\,$\pm$\,1.08& &-15.53\\
        &Replay (Poor)&42.70\,$\pm$\,14.86& 36000& -27.77\\
        &Replay (Medium)&52.40\,$\pm$\,21.79& &-23.32\\
        &Replay (Good)&67.10\,$\pm$\,18.16& &-17.17\\
\hline
\end{tabular}
\end{table*}

\begin{figure*}[htbp]
	\centering
	\begin{subfigure}{\linewidth}
		\centering
    	\begin{subfigure}{0.33\linewidth}
    		\centering
    		\includegraphics[width=\linewidth]{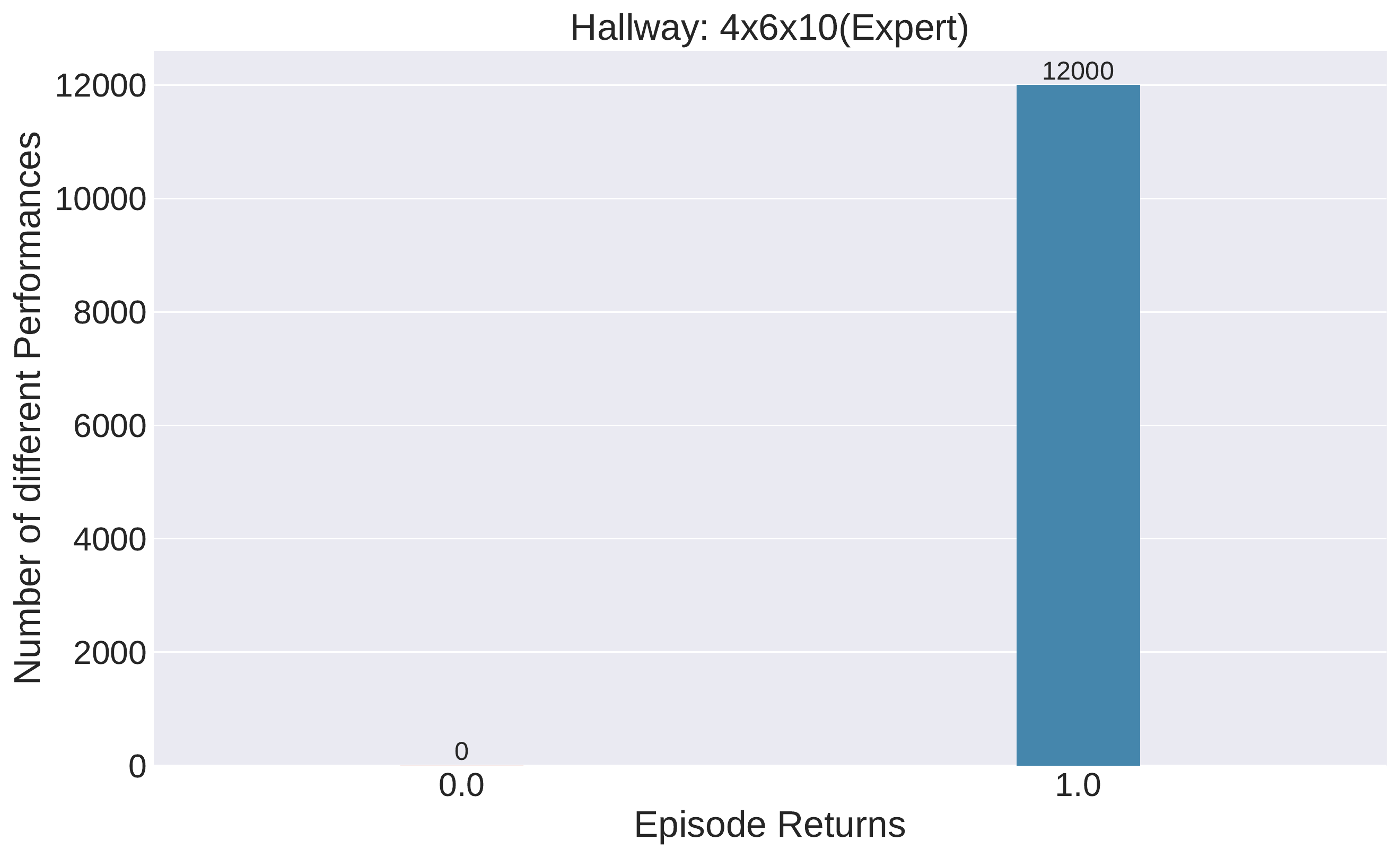}
    	\label{fig:Hallway(Expert)}
    	\end{subfigure}
    	\hspace{-0.7em}
    	\begin{subfigure}{0.33\linewidth}
    		\centering
    		\includegraphics[width=\linewidth]{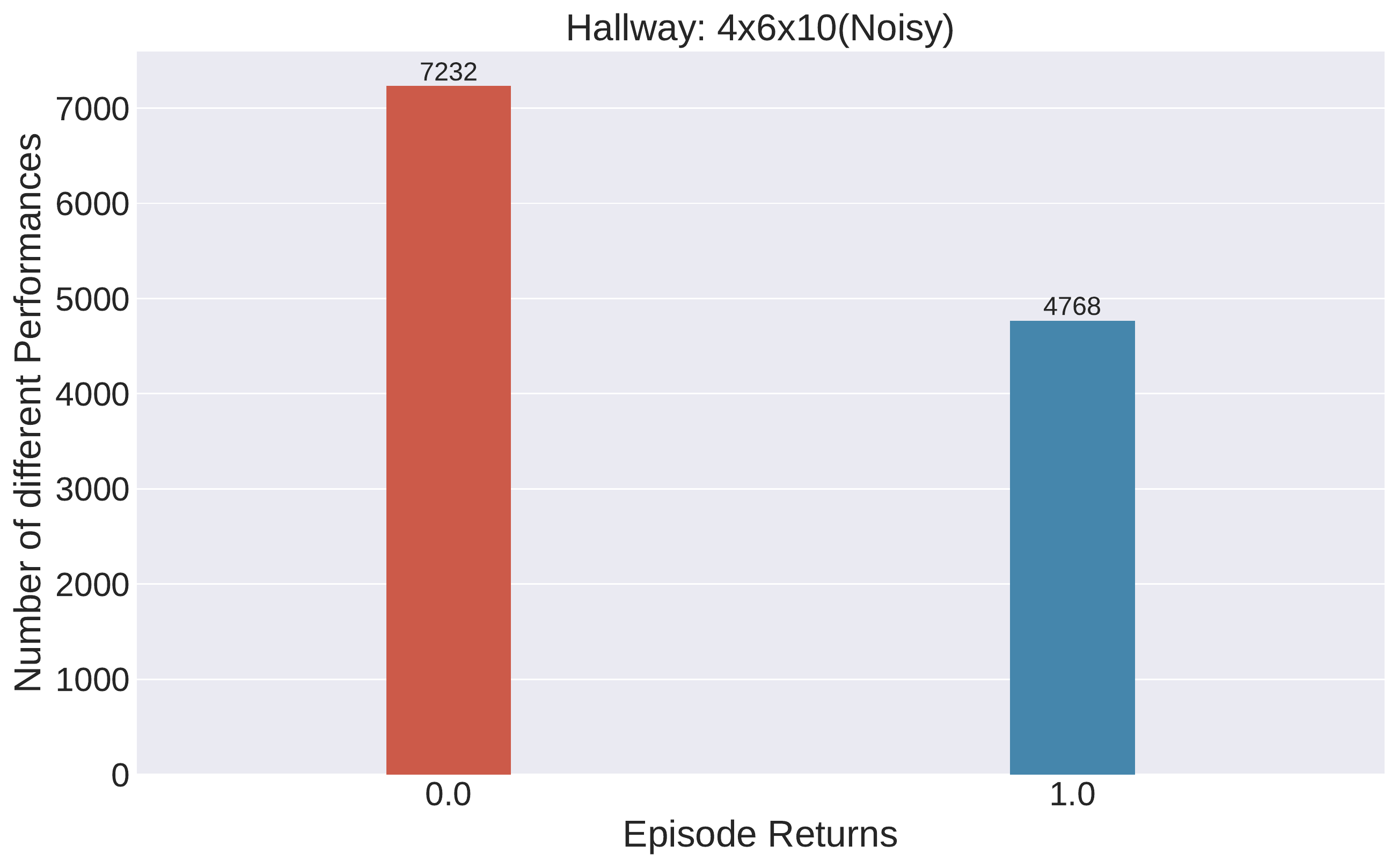}
    	\label{fig:Hallway(Noisy)}
    	\end{subfigure}
    	\hspace{-0.7em}
    	\begin{subfigure}{0.33\linewidth}
    		\centering
    		\includegraphics[width=\linewidth]{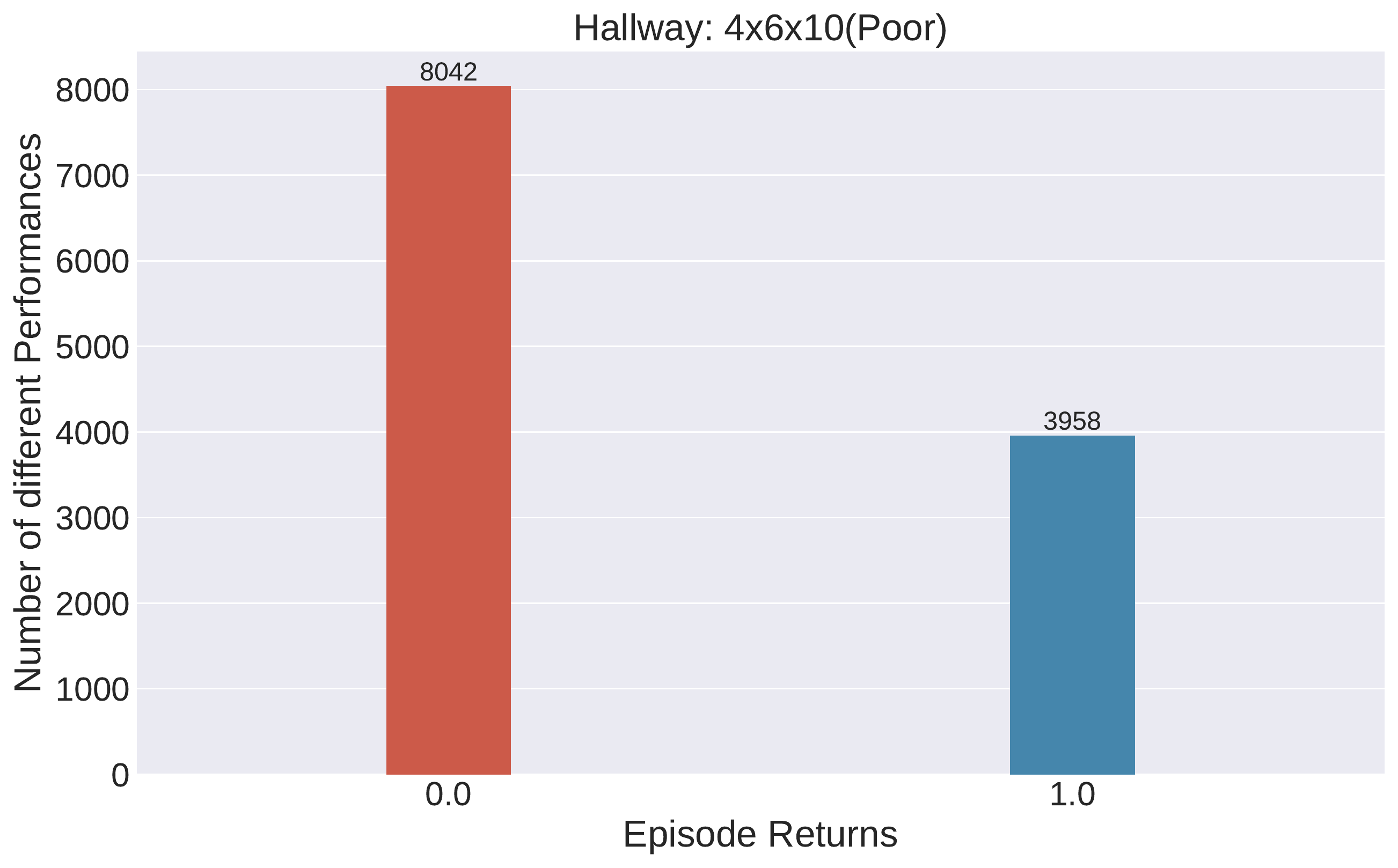}
    	\label{fig:Hallway(Poor)}
    	\end{subfigure}
	\end{subfigure}
	\begin{subfigure}{\linewidth}
        \centering
         \hspace{-0.7em}
    	\begin{subfigure}{0.33\linewidth}
    		\centering
    		\includegraphics[width=\linewidth]{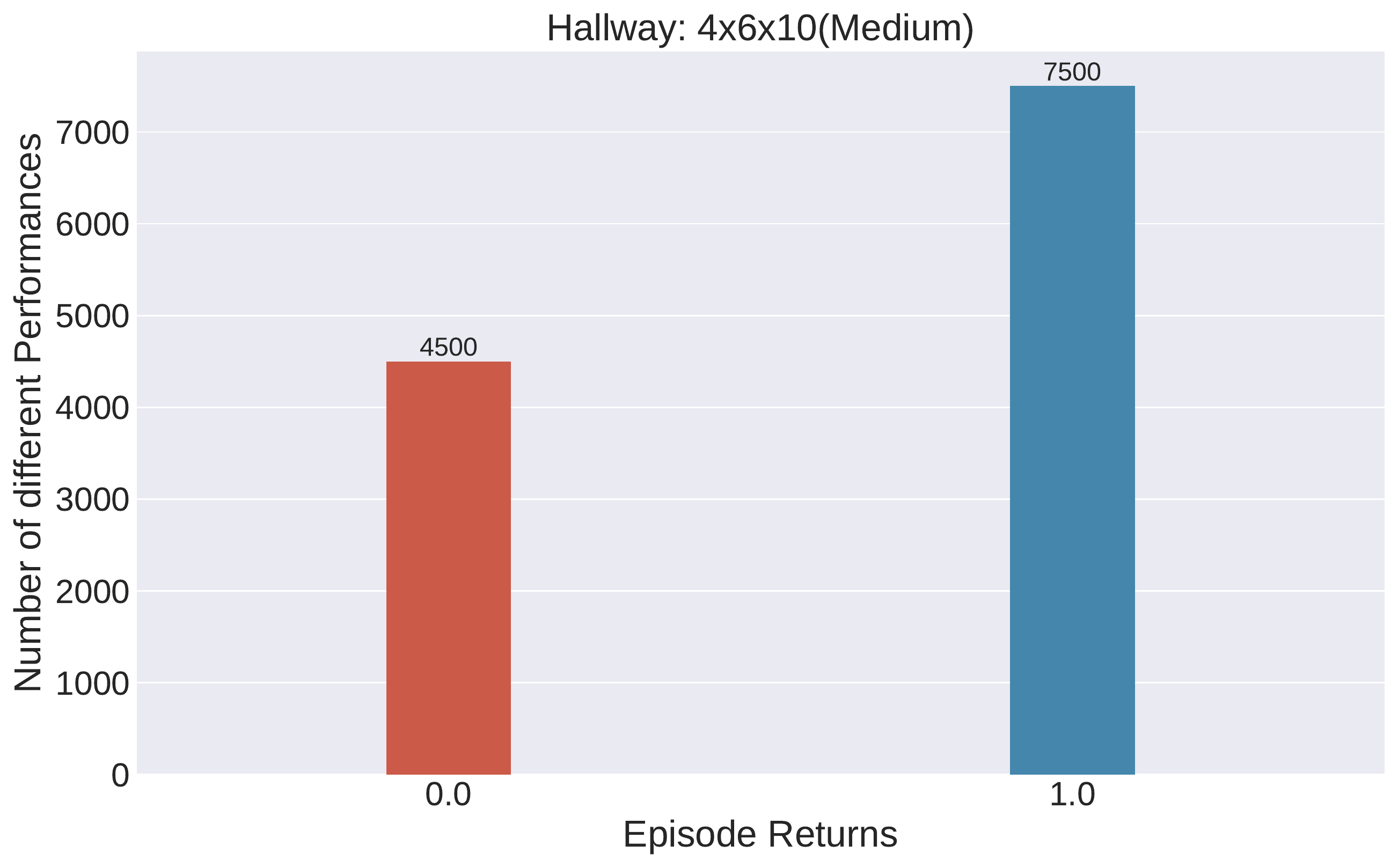}
    	\label{fig:Hallway(Medium)}
    	\end{subfigure}
    	\hspace{-0.7em}
    	\begin{subfigure}{0.33\linewidth}
    		\centering
    		\includegraphics[width=\linewidth]{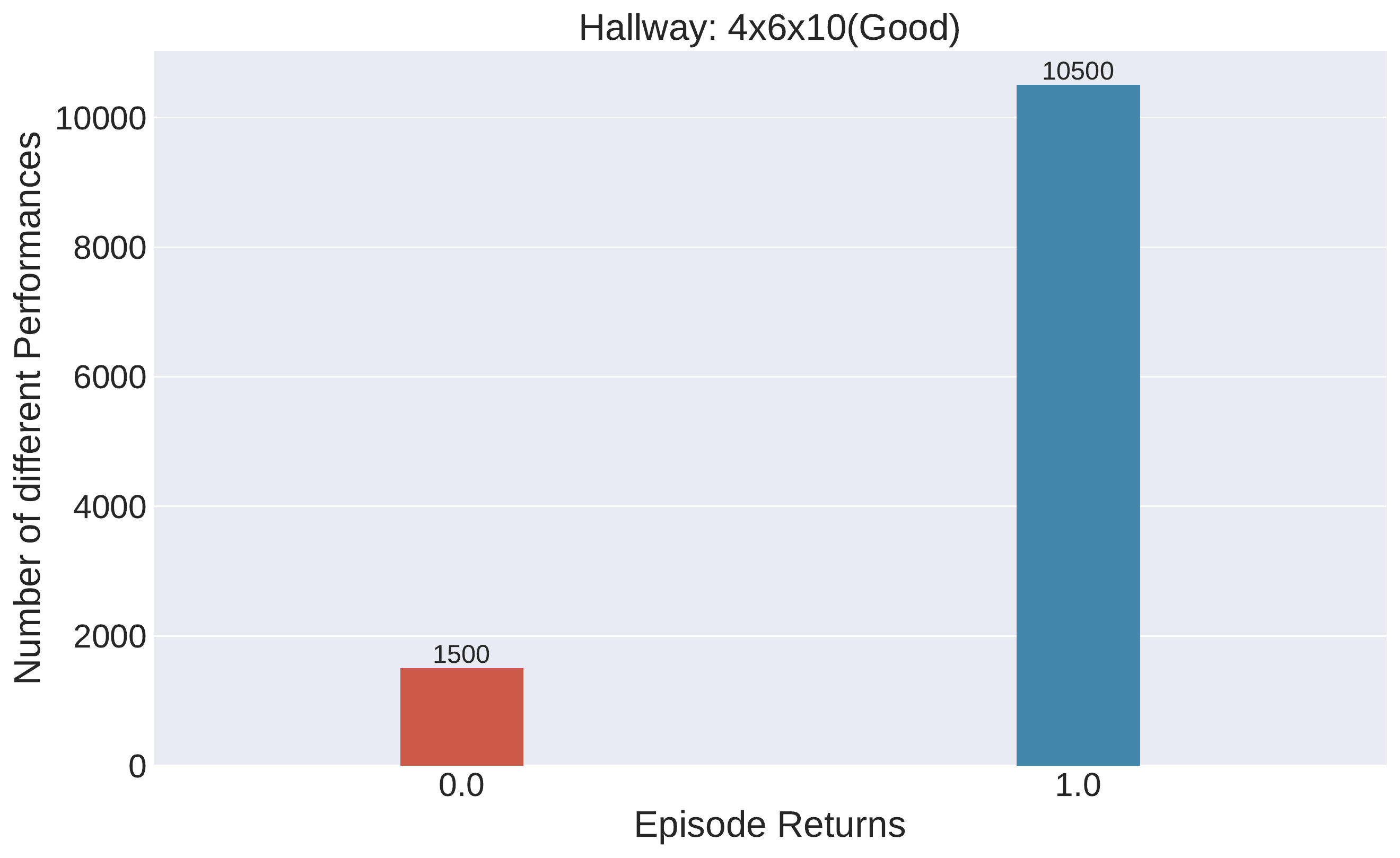}
    	\label{fig:Hallway(Good)}
    	\end{subfigure}
    	\hspace{-0.7em}
    	\begin{subfigure}{0.33\linewidth}
    		\centering
    		\includegraphics[width=\linewidth]{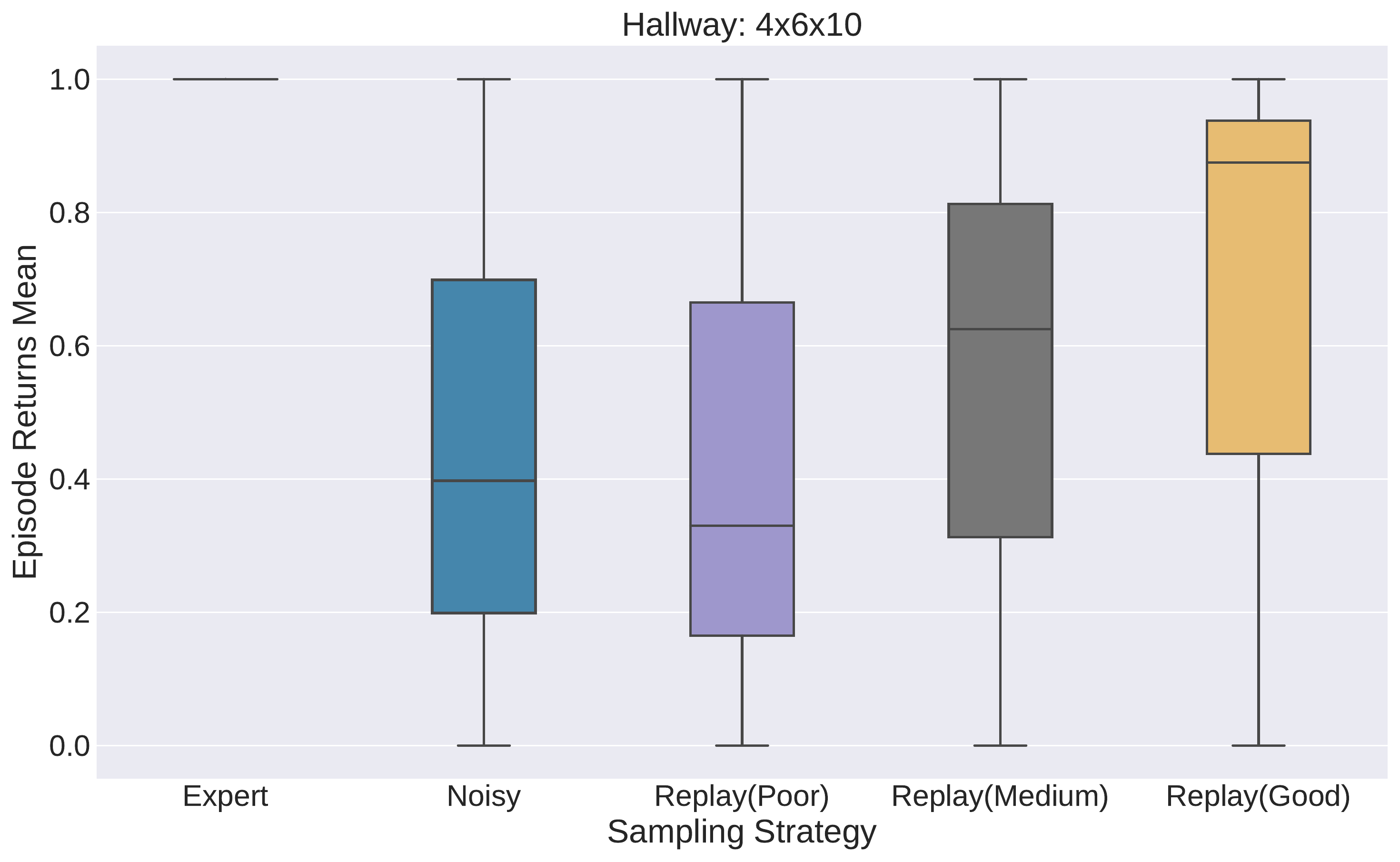}
    	\label{fig:Hallway(ALL)}
    	\end{subfigure}
	\end{subfigure}
	\caption{Hallway: 4x6x10 offline dataset distribution. }
	\label{fig:hallway_data_dist}
	\vspace*{-5mm}
\end{figure*}

\begin{figure*}[htbp]
	\centering
	\begin{subfigure}{\linewidth}
		\centering
    	\begin{subfigure}{0.33\linewidth}
    		\centering
    		\includegraphics[width=\linewidth]{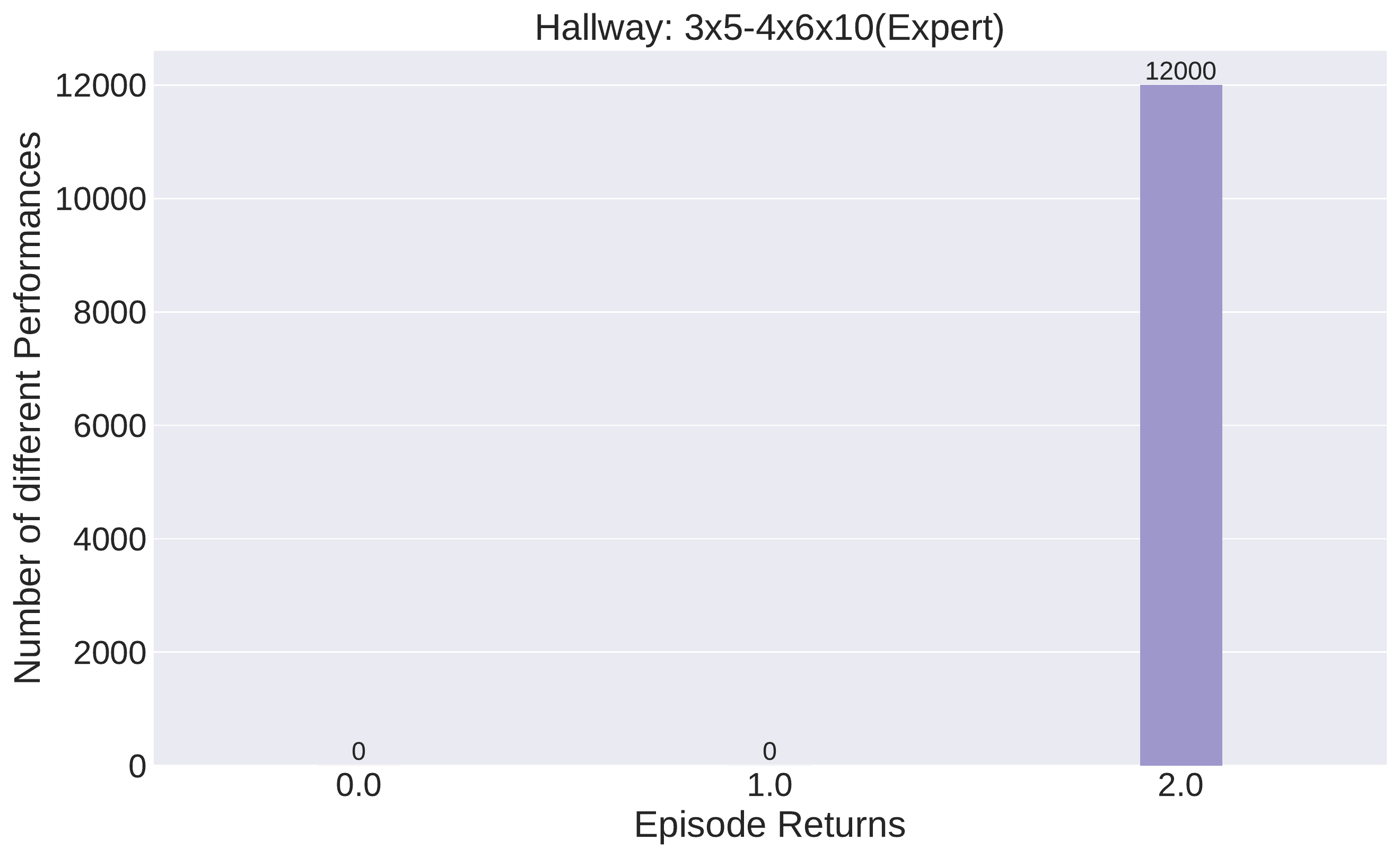}
    	\label{fig:Hallway_group(Expert)}
    	\end{subfigure}
    	\hspace{-0.7em}
    	\begin{subfigure}{0.33\linewidth}
    		\centering
    		\includegraphics[width=\linewidth]{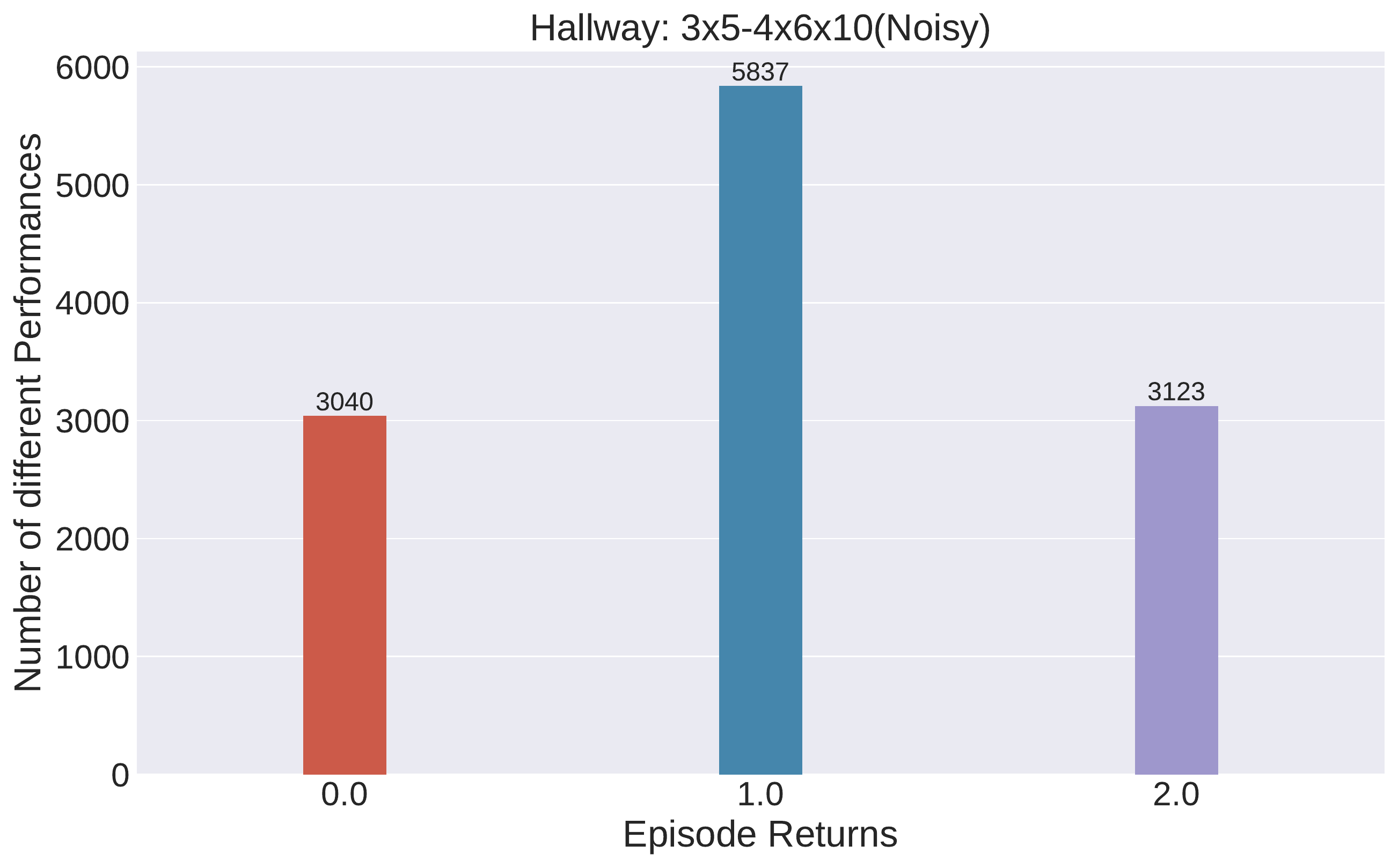}
    	\label{fig:Hallway_group(Noisy)}
    	\end{subfigure}
    	\hspace{-0.7em}
    	\begin{subfigure}{0.33\linewidth}
    		\centering
    		\includegraphics[width=\linewidth]{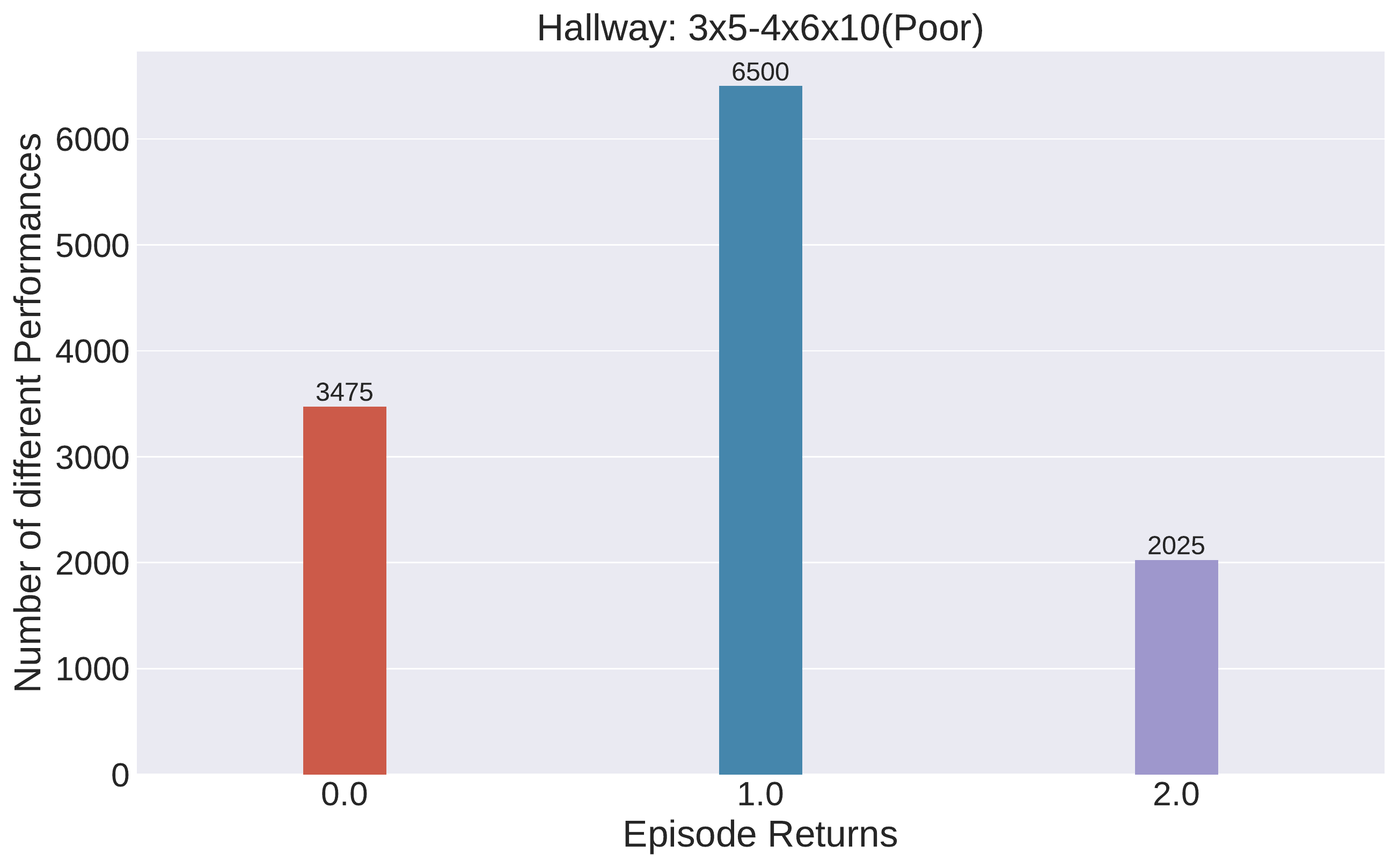}
    	\label{fig:Hallway_group(Poor)}
    	\end{subfigure}
	\end{subfigure}
	\begin{subfigure}{\linewidth}
        \centering
         \hspace{-0.7em}
    	\begin{subfigure}{0.33\linewidth}
    		\centering
    		\includegraphics[width=\linewidth]{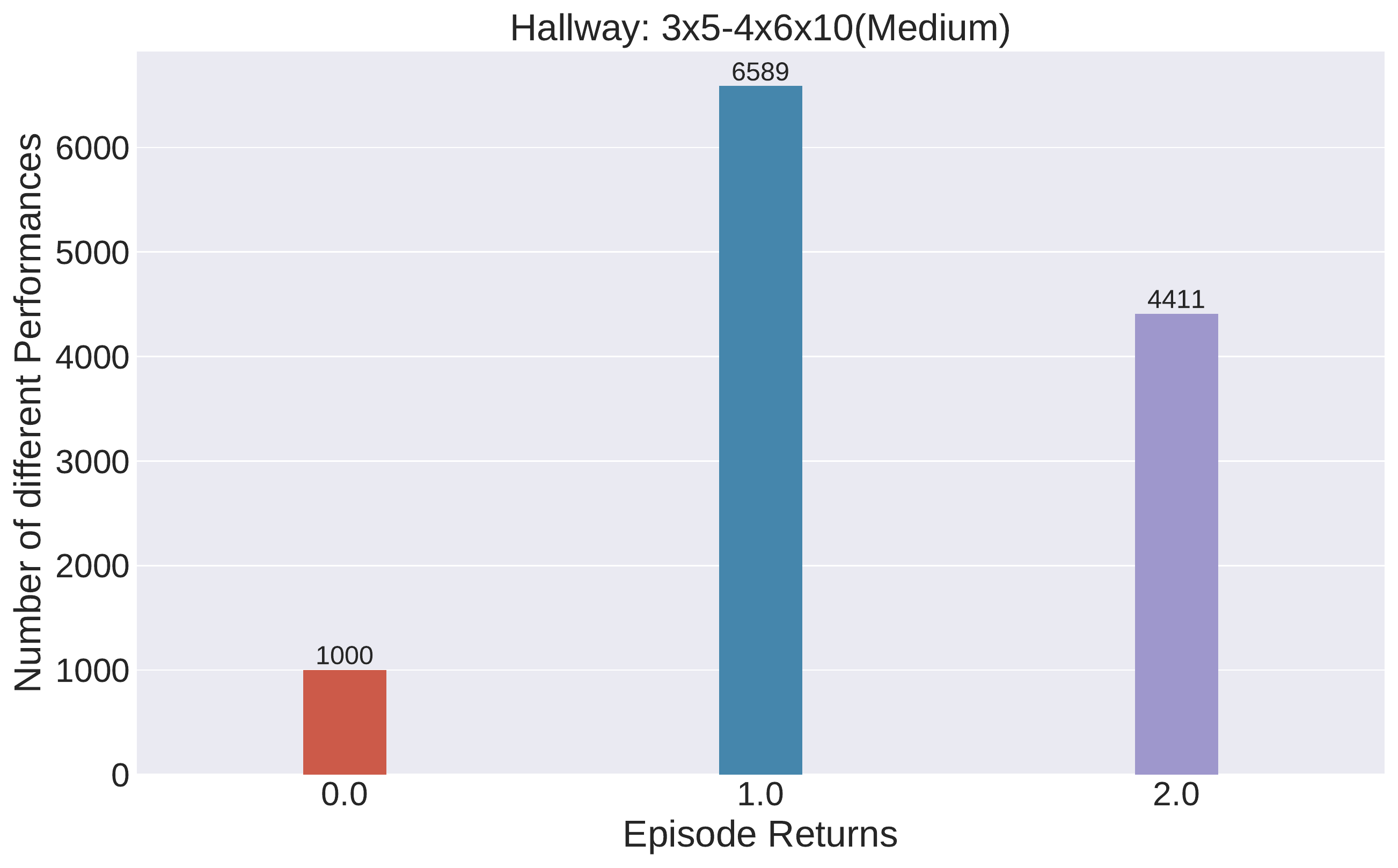}
    	\label{fig:Hallway_group(Medium)}
    	\end{subfigure}
    	\hspace{-0.7em}
    	\begin{subfigure}{0.33\linewidth}
    		\centering
    		\includegraphics[width=\linewidth]{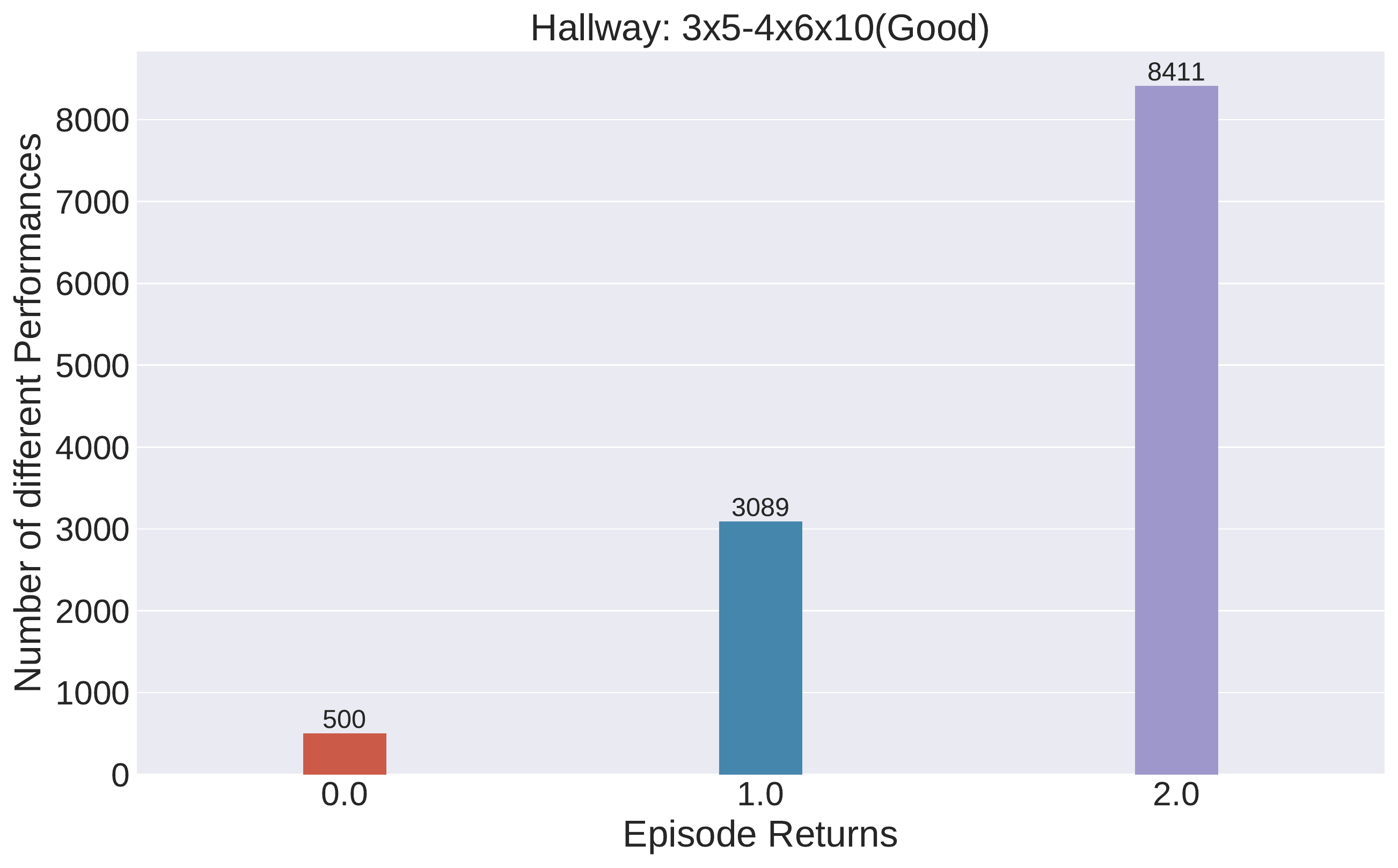}
    	\label{fig:Hallway_group(Good)}
    	\end{subfigure}
    	\hspace{-0.7em}
    	\begin{subfigure}{0.33\linewidth}
    		\centering
    		\includegraphics[width=\linewidth]{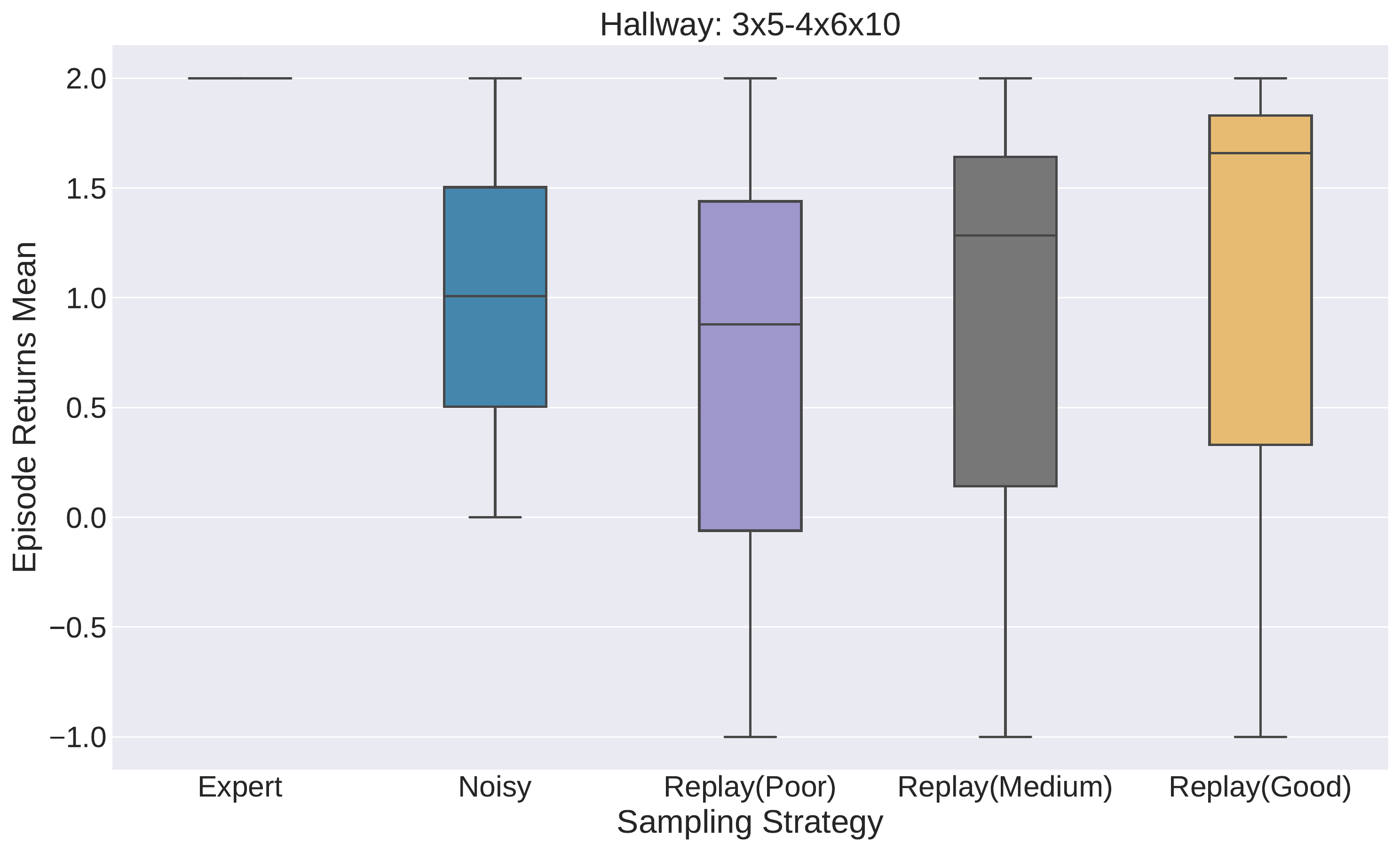}
    	\label{fig:Hallway_group(ALL)}
    	\end{subfigure}
	\end{subfigure}
	\caption{Hallway: 3x5-4x6x10 offline dataset distribution. }
	\label{fig:hallway_group_data_dist}
	\vspace*{-5mm}
\end{figure*}

\begin{figure*}[htbp]
	\centering
	\begin{subfigure}{\linewidth}
		\centering
    	\begin{subfigure}{0.33\linewidth}
    		\centering
    		\includegraphics[width=\linewidth]{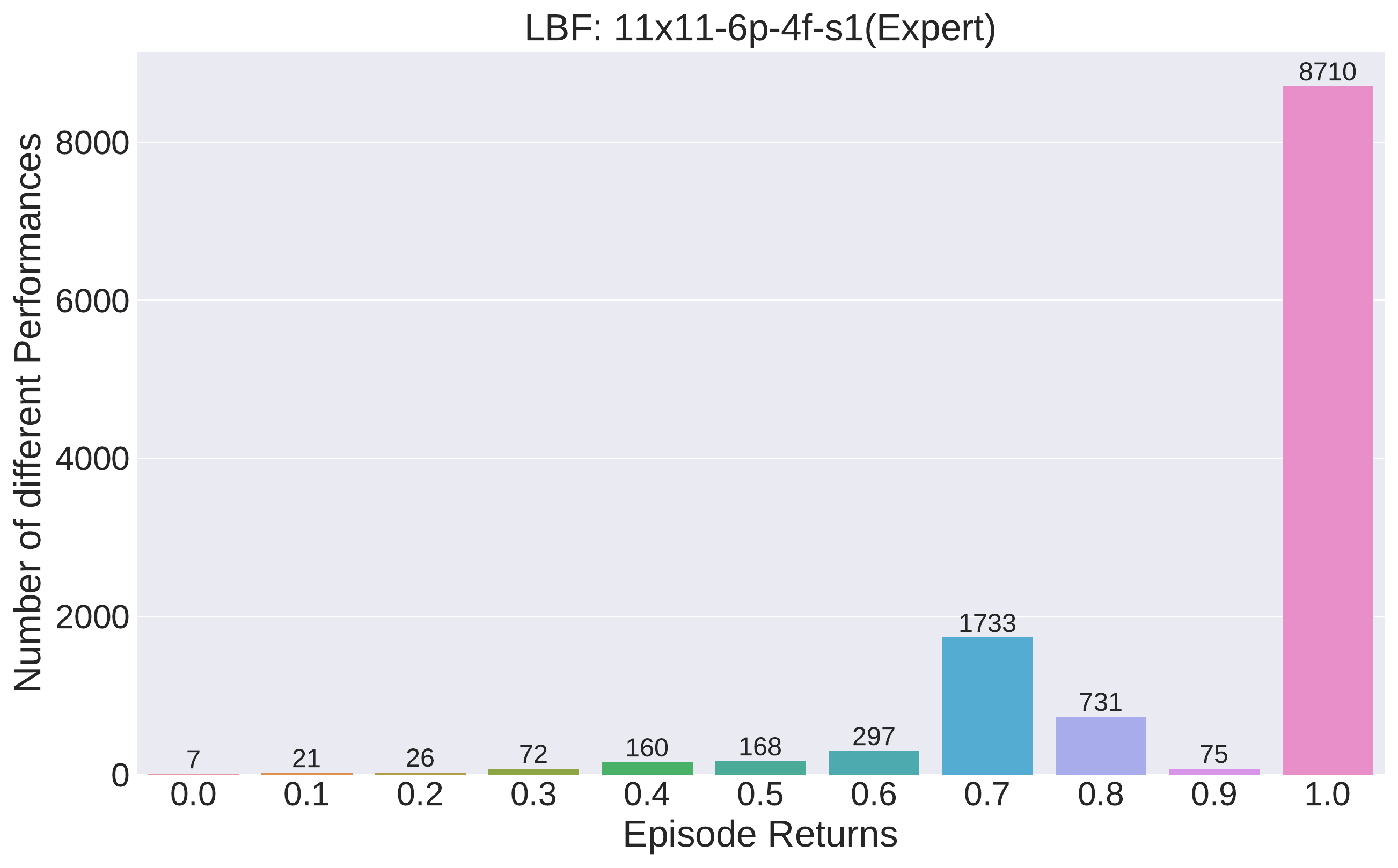}
    	\label{fig:LBF11x11(Expert)}
    	\end{subfigure}
    	\hspace{-0.7em}
    	\begin{subfigure}{0.33\linewidth}
    		\centering
    		\includegraphics[width=\linewidth]{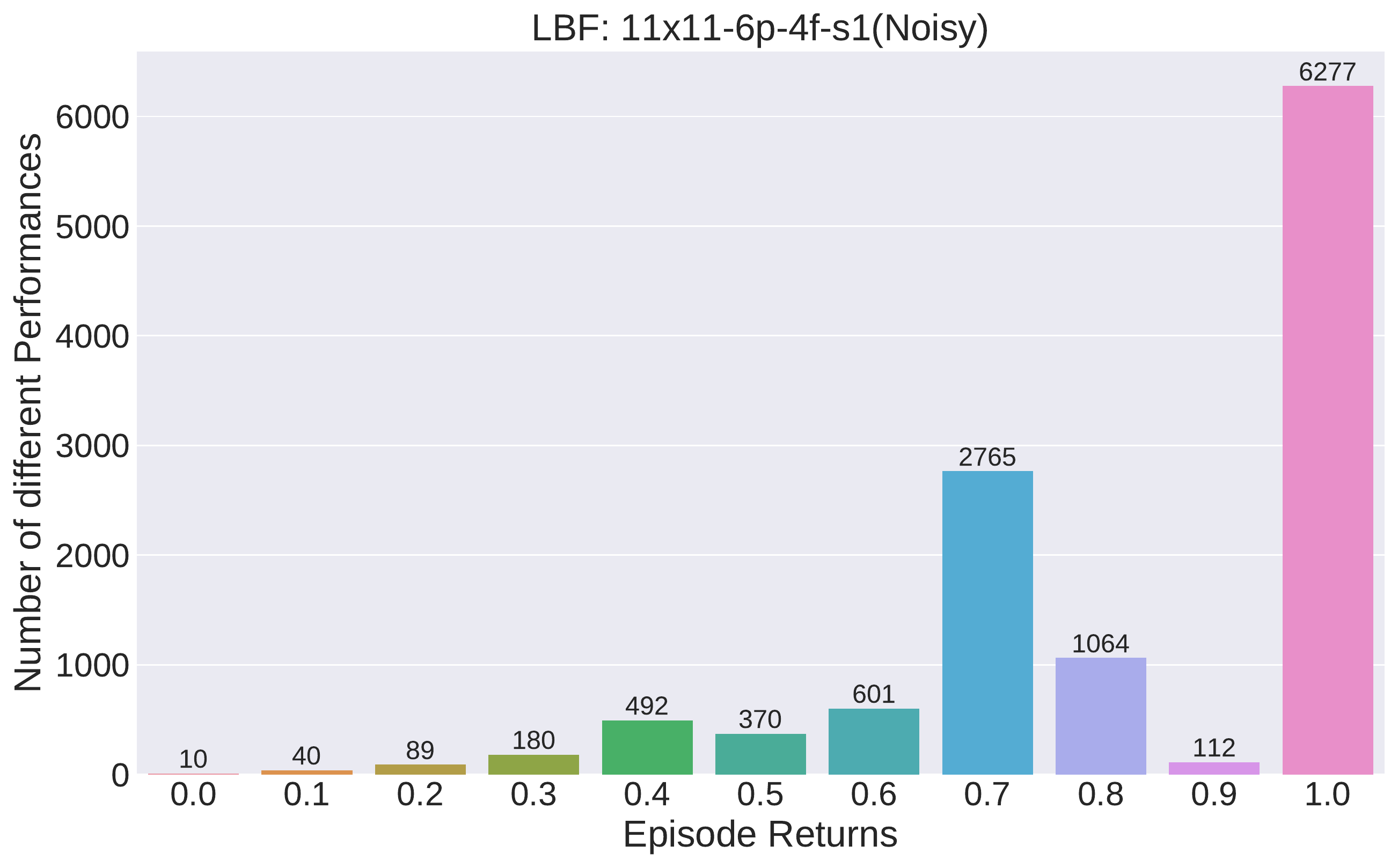}
    	\label{fig:LBF11x11(Noisy)}
    	\end{subfigure}
    	\hspace{-0.7em}
    	\begin{subfigure}{0.33\linewidth}
    		\centering
    		\includegraphics[width=\linewidth]{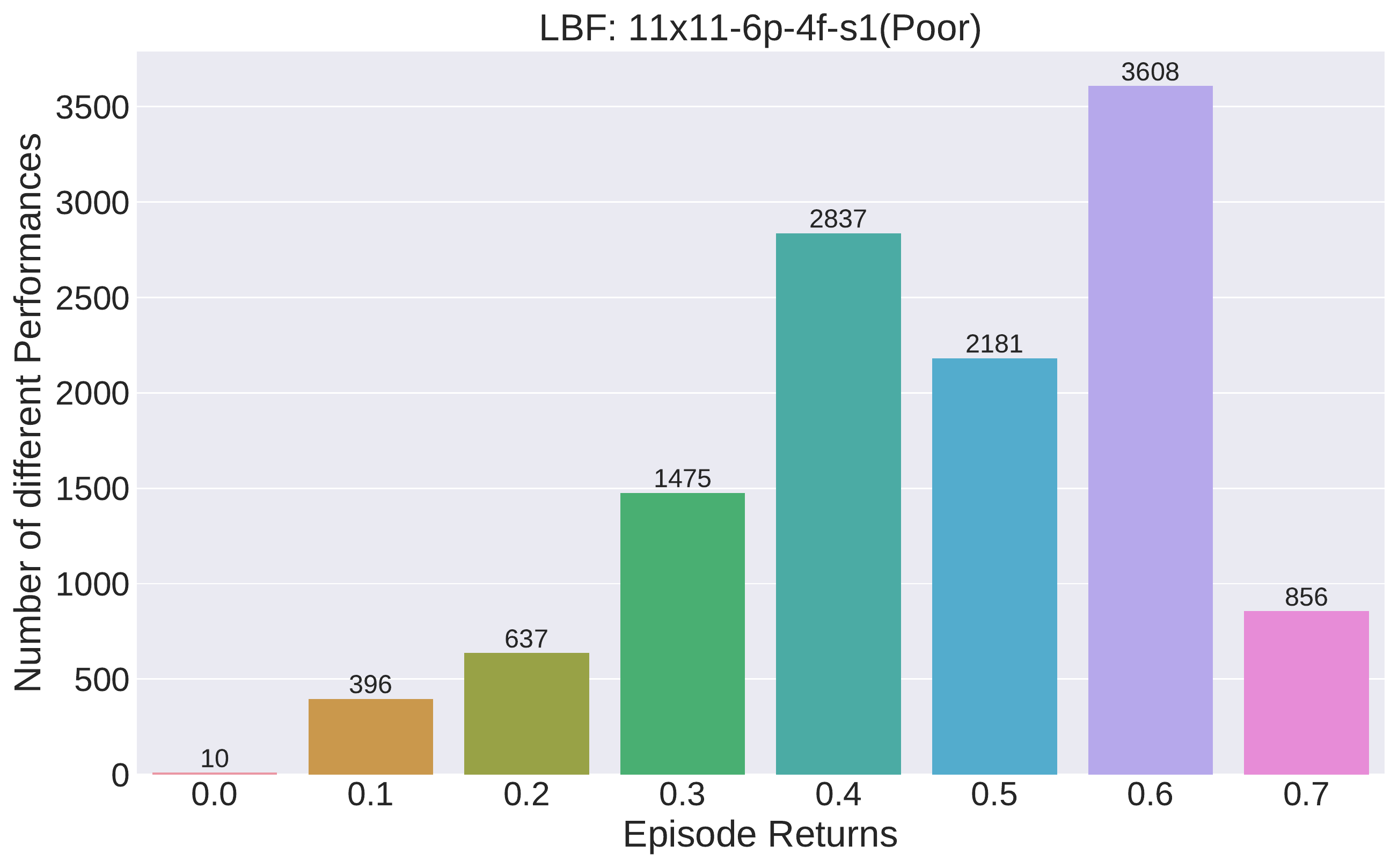}
    	\label{fig:LBF11x11(Poor)}
    	\end{subfigure}
	\end{subfigure}
	\begin{subfigure}{\linewidth}
        \centering
         \hspace{-0.7em}
    	\begin{subfigure}{0.33\linewidth}
    		\centering
    		\includegraphics[width=\linewidth]{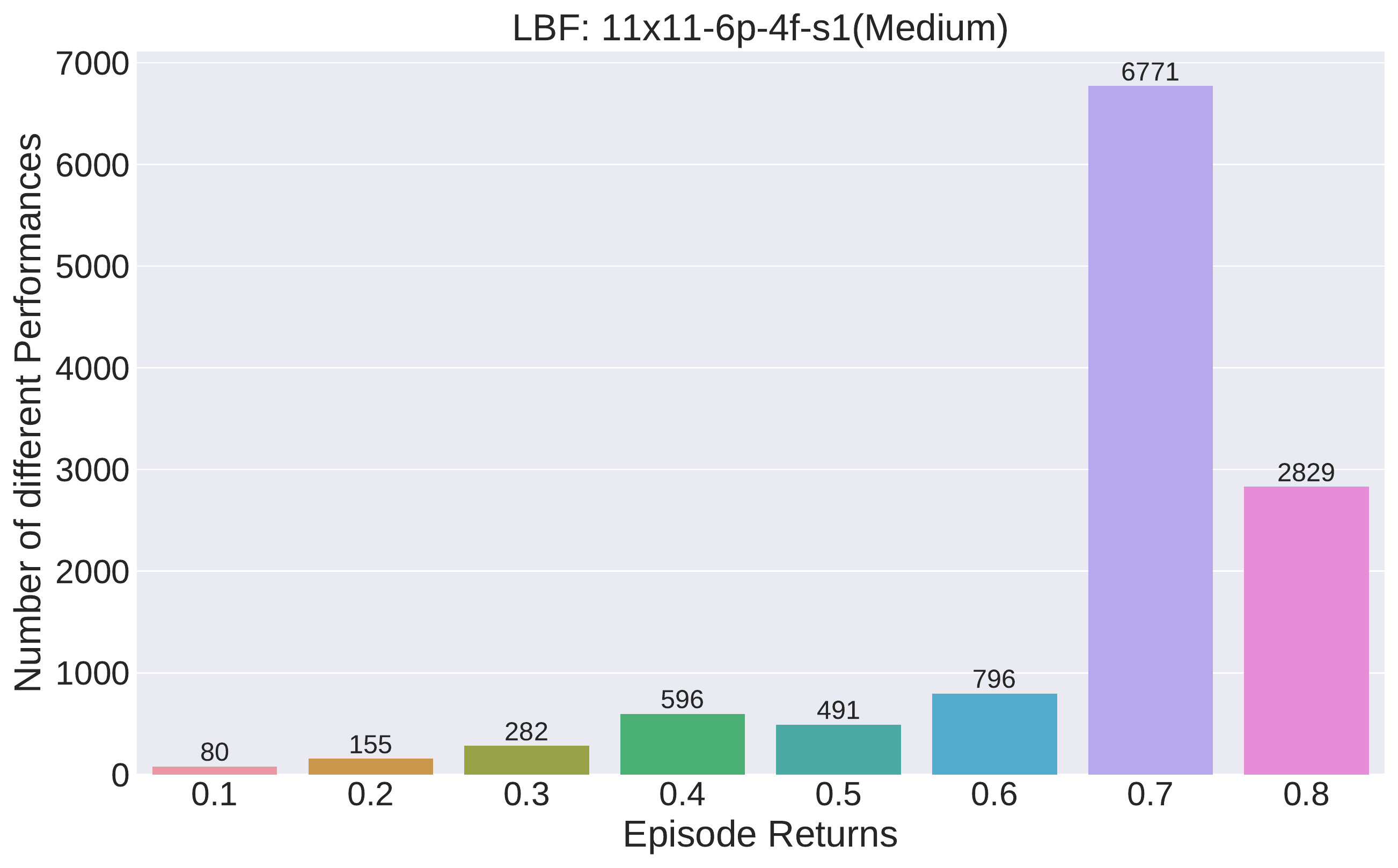}
    	\label{fig:LBF11x11(Medium)}
    	\end{subfigure}
    	\hspace{-0.7em}
    	\begin{subfigure}{0.33\linewidth}
    		\centering
    		\includegraphics[width=\linewidth]{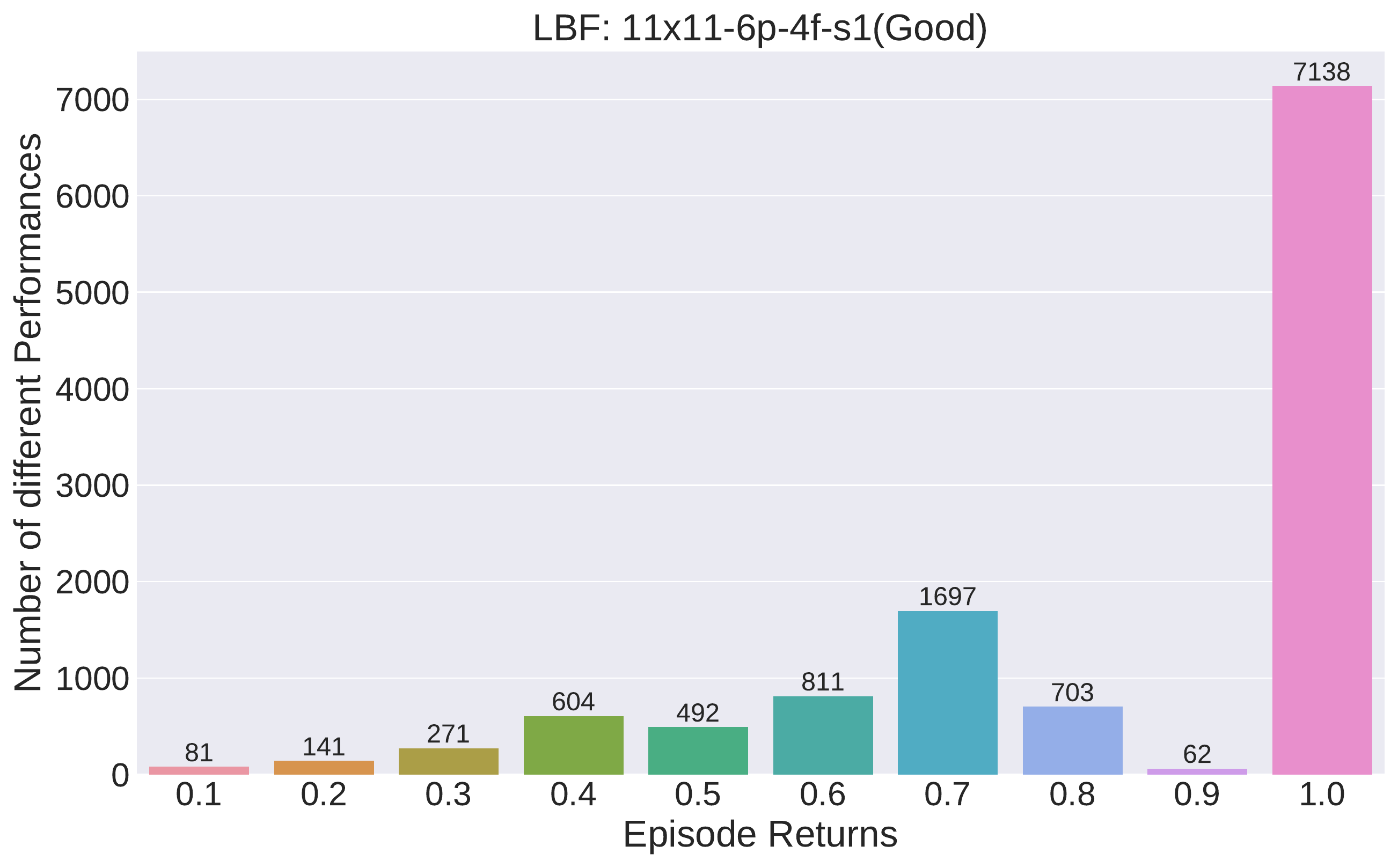}
    	\label{fig:LBF11x11(Good)}
    	\end{subfigure}
    	\hspace{-0.7em}
    	\begin{subfigure}{0.33\linewidth}
    		\centering
    		\includegraphics[width=\linewidth]{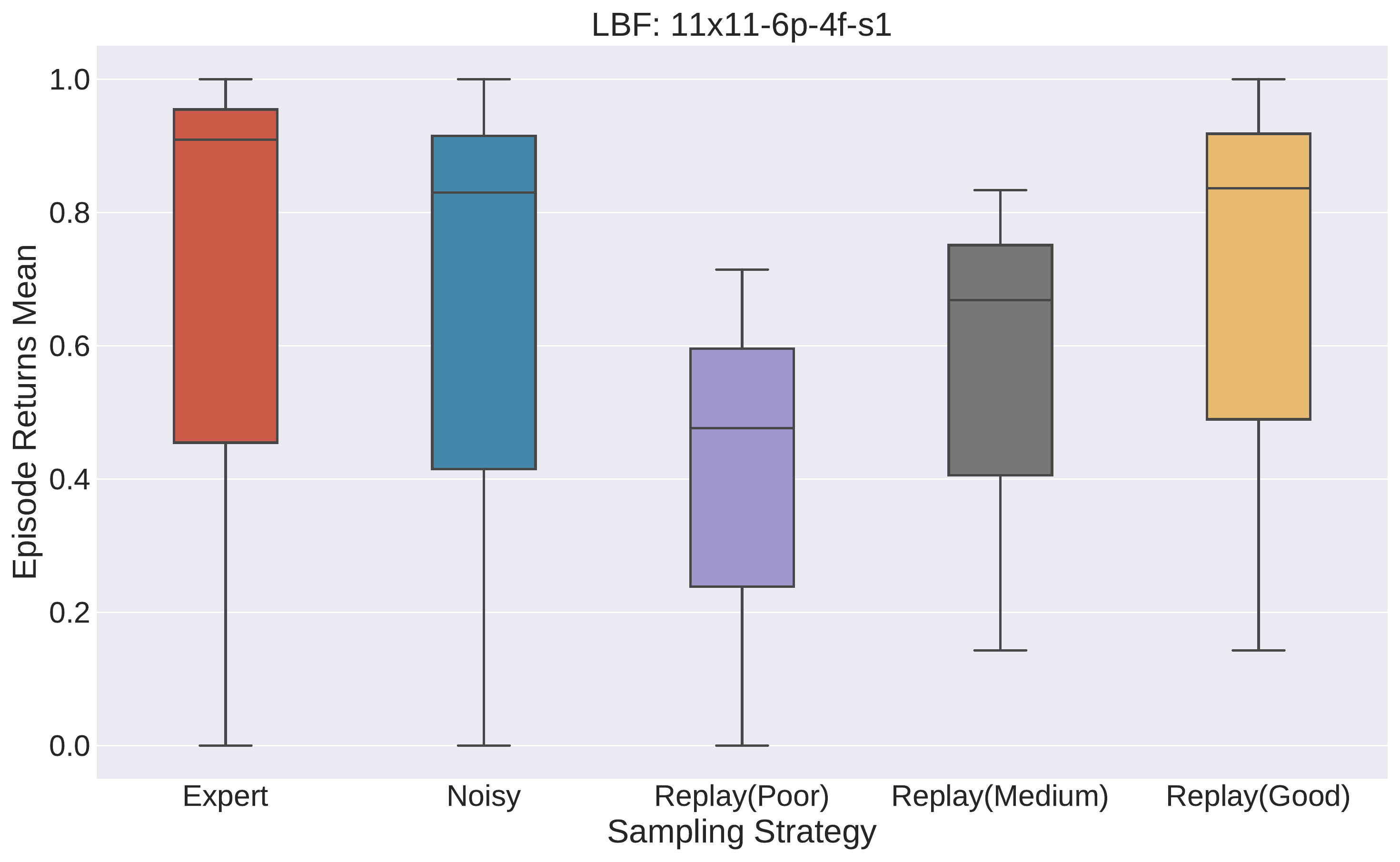}
    	\label{fig:LBF11x11(ALL)}
    	\end{subfigure}
	\end{subfigure}
	\caption{LBF: 11x11-6p-4f-s1 offline dataset distribution. }
	\label{fig:lbf1_data_dist}
	\vspace*{-5mm}
\end{figure*}

\begin{figure*}[htbp]
	\centering
	\begin{subfigure}{\linewidth}
		\centering
    	\begin{subfigure}{0.33\linewidth}
    		\centering
    		\includegraphics[width=\linewidth]{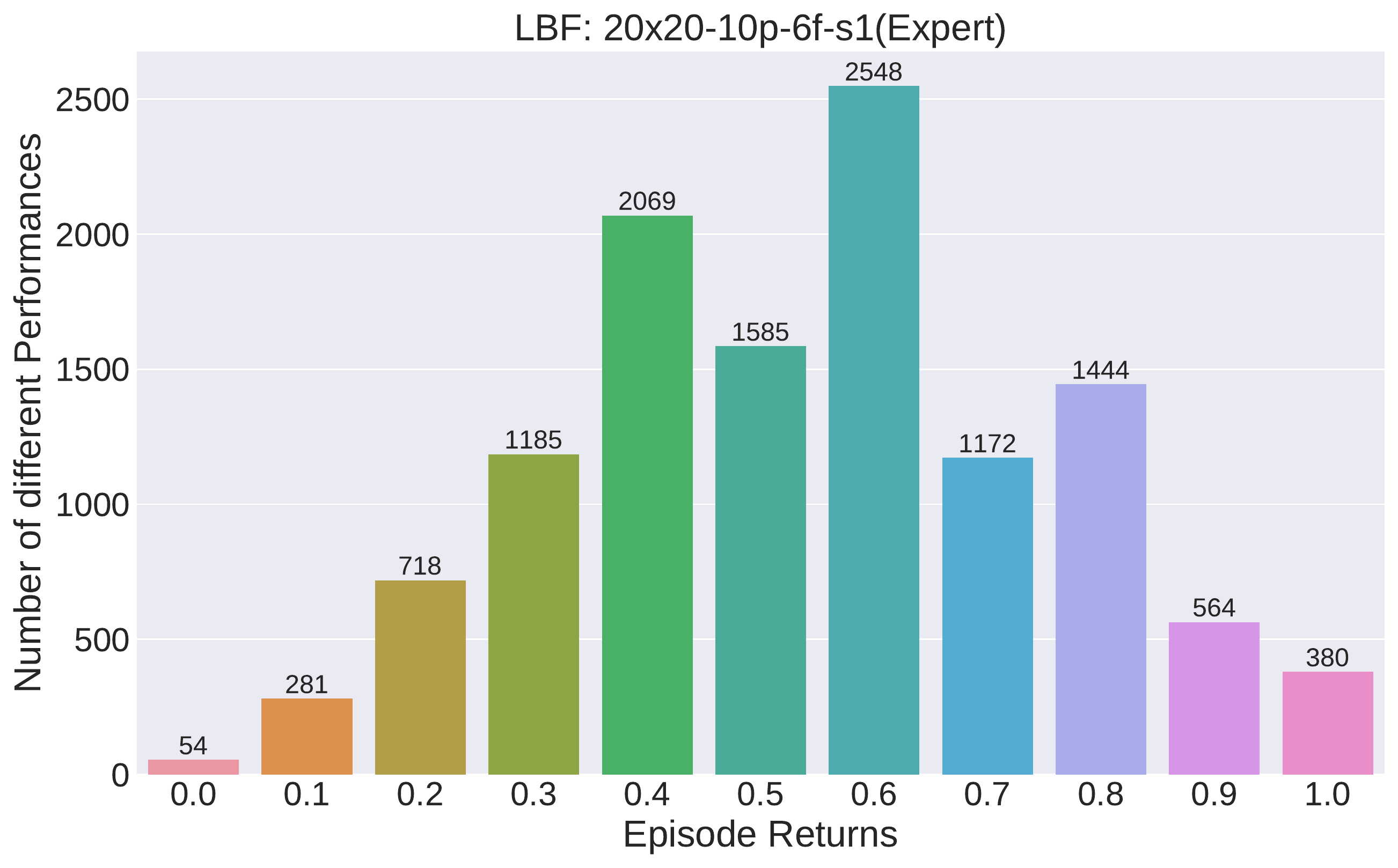}
    	\label{fig:LBF20x20(Expert)}
    	\end{subfigure}
    	\hspace{-0.7em}
    	\begin{subfigure}{0.33\linewidth}
    		\centering
    		\includegraphics[width=\linewidth]{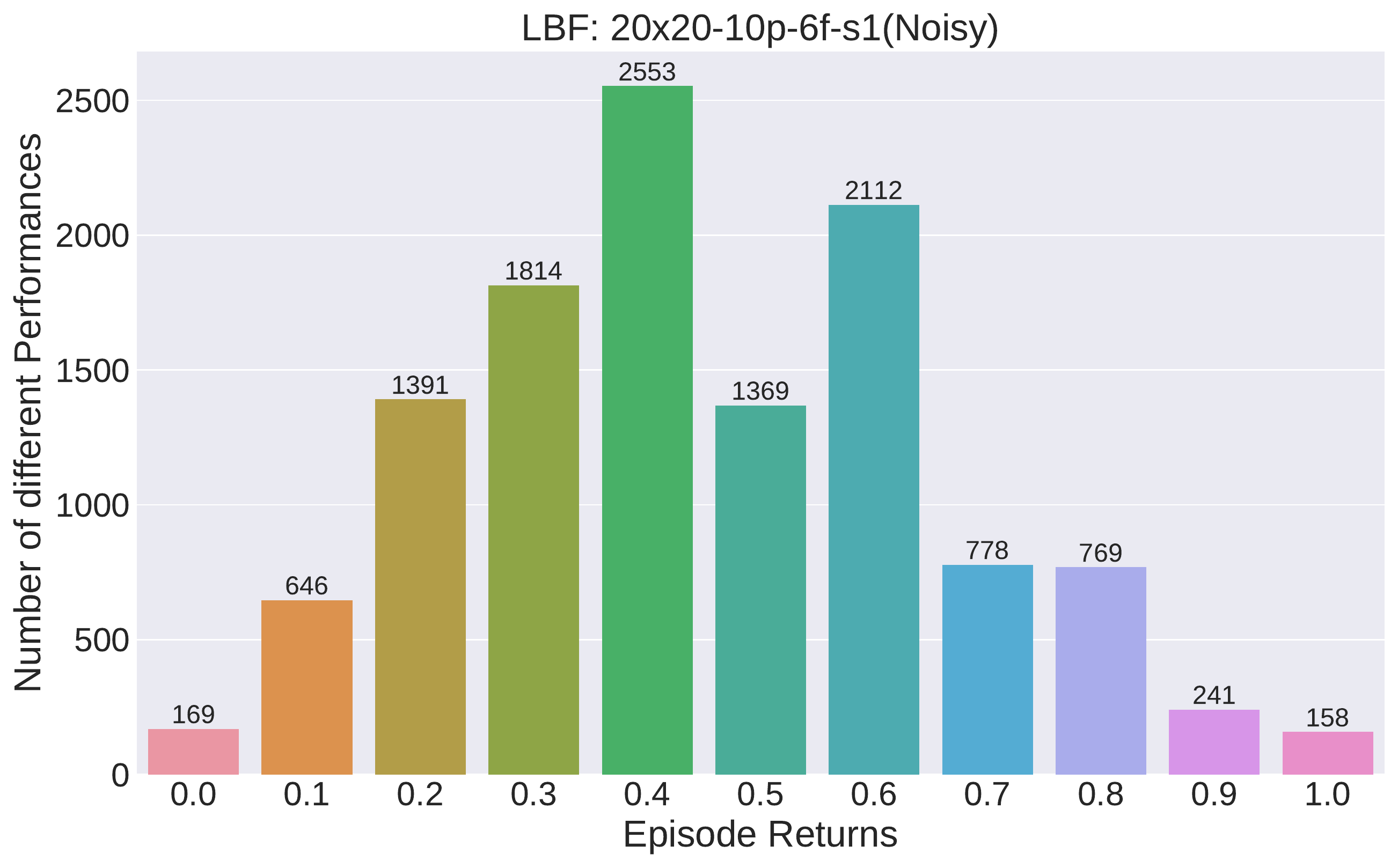}
    	\label{fig:LBF20x20(Noisy)}
    	\end{subfigure}
    	\hspace{-0.7em}
    	\begin{subfigure}{0.33\linewidth}
    		\centering
    		\includegraphics[width=\linewidth]{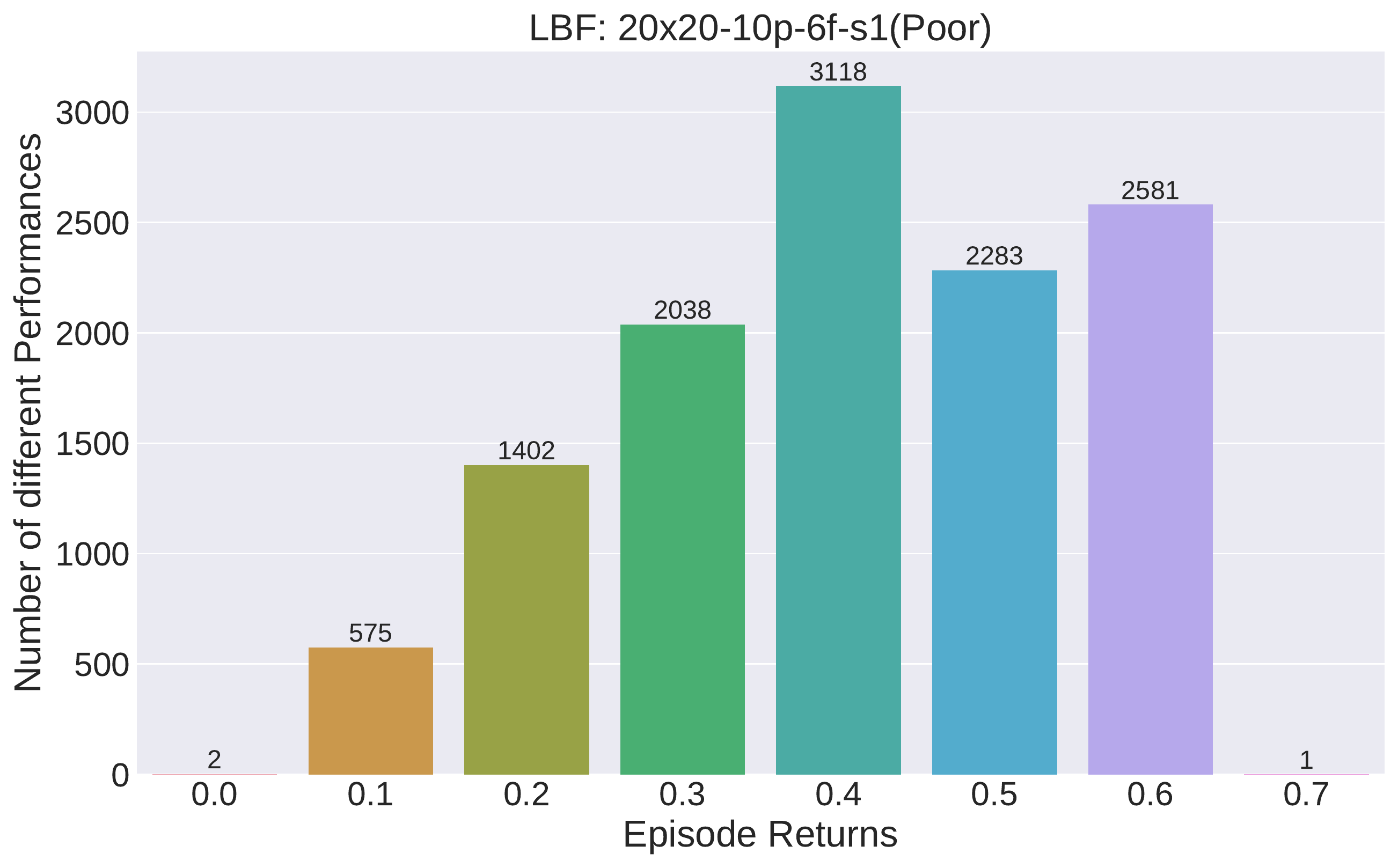}
    	\label{fig:LBF20x20(Poor)}
    	\end{subfigure}
	\end{subfigure}
	\begin{subfigure}{\linewidth}
        \centering
         \hspace{-0.7em}
    	\begin{subfigure}{0.33\linewidth}
    		\centering
    		\includegraphics[width=\linewidth]{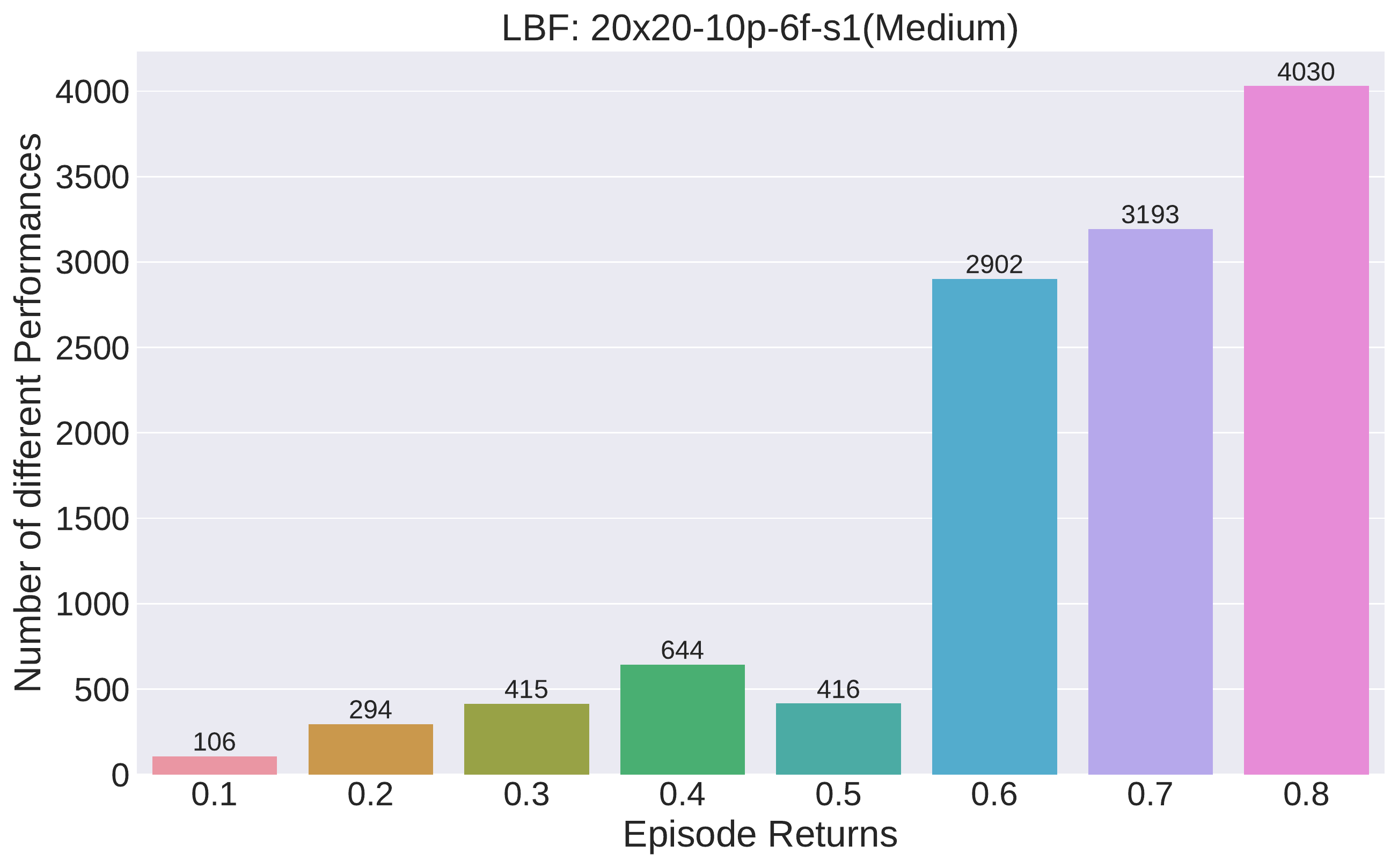}
    	\label{fig:LBF20x20(Medium)}
    	\end{subfigure}
    	\hspace{-0.7em}
    	\begin{subfigure}{0.33\linewidth}
    		\centering
    		\includegraphics[width=\linewidth]{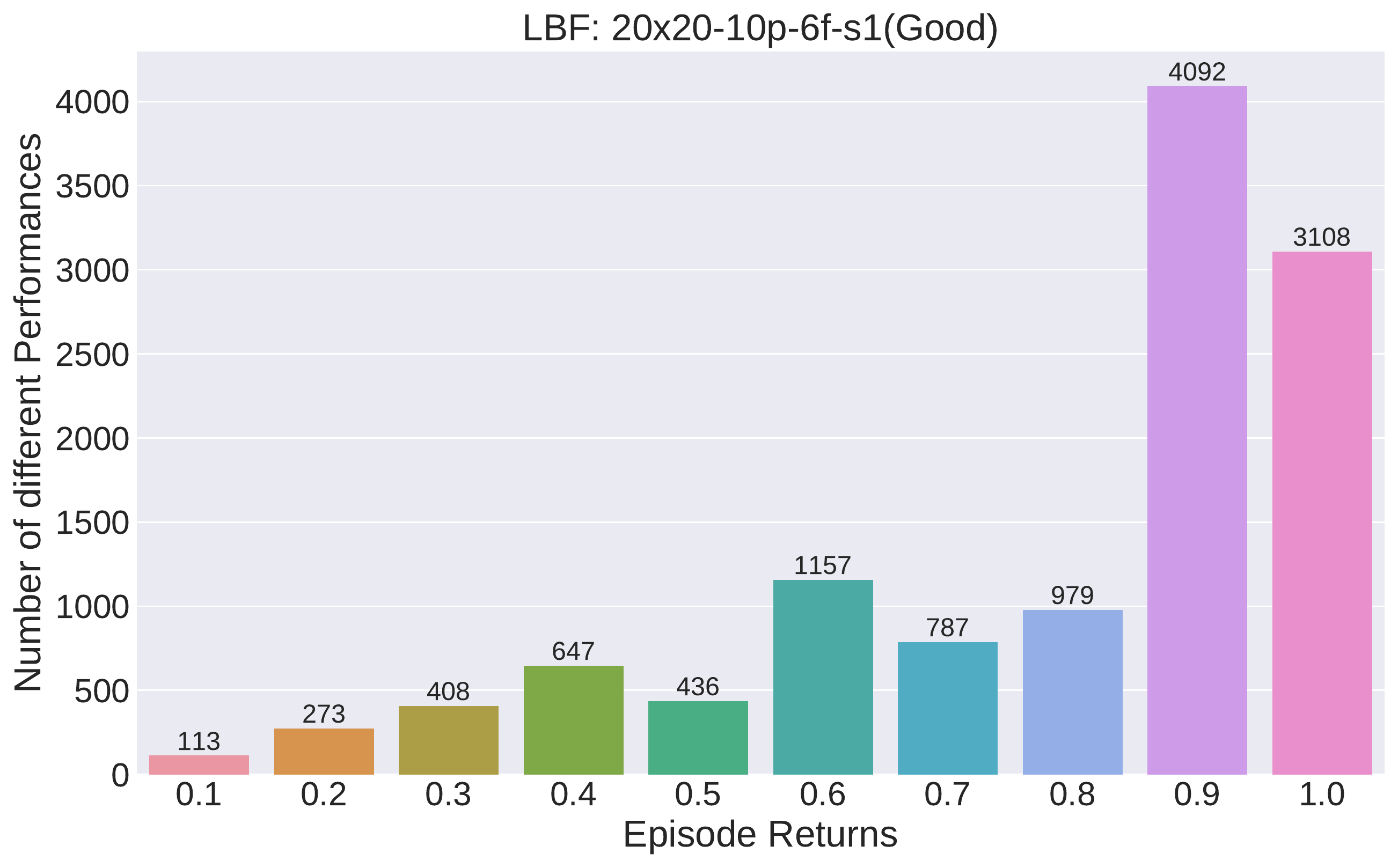}
    	\label{fig:LBF20x20(Good)}
    	\end{subfigure}
    	\hspace{-0.7em}
    	\begin{subfigure}{0.33\linewidth}
    		\centering
    		\includegraphics[width=\linewidth]{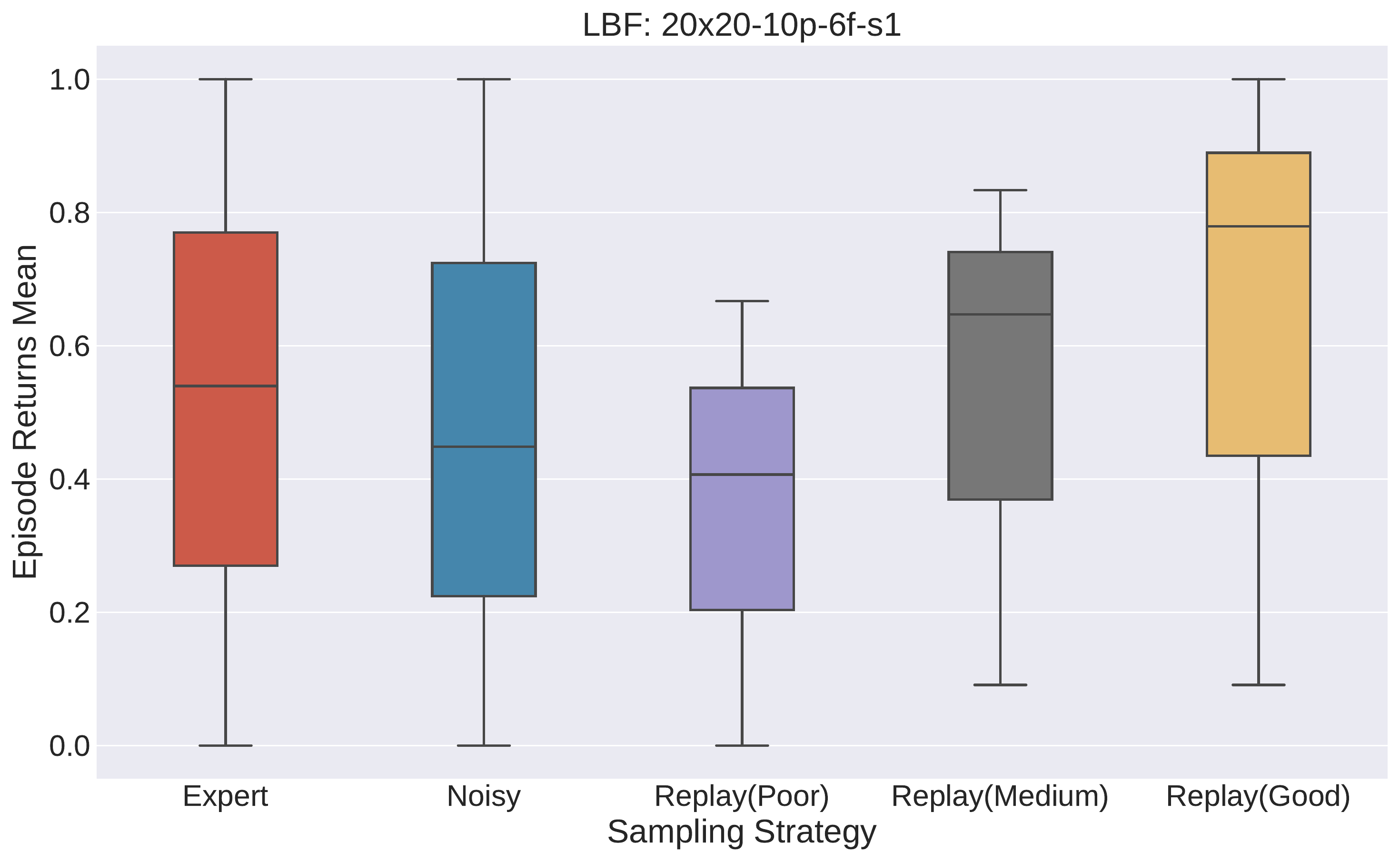}
    	\label{fig:LBF20x20(ALL)}
    	\end{subfigure}
	\end{subfigure}
	\caption{LBF: 20x20-10p-6f-s1 offline dataset distribution. }
	\label{fig:lbf2_data_dist}
	\vspace*{-5mm}
\end{figure*}

\begin{figure*}[htbp]
	\centering
	\begin{subfigure}{\linewidth}
		\centering
    	\begin{subfigure}{0.33\linewidth}
    		\centering
    		\includegraphics[width=\linewidth]{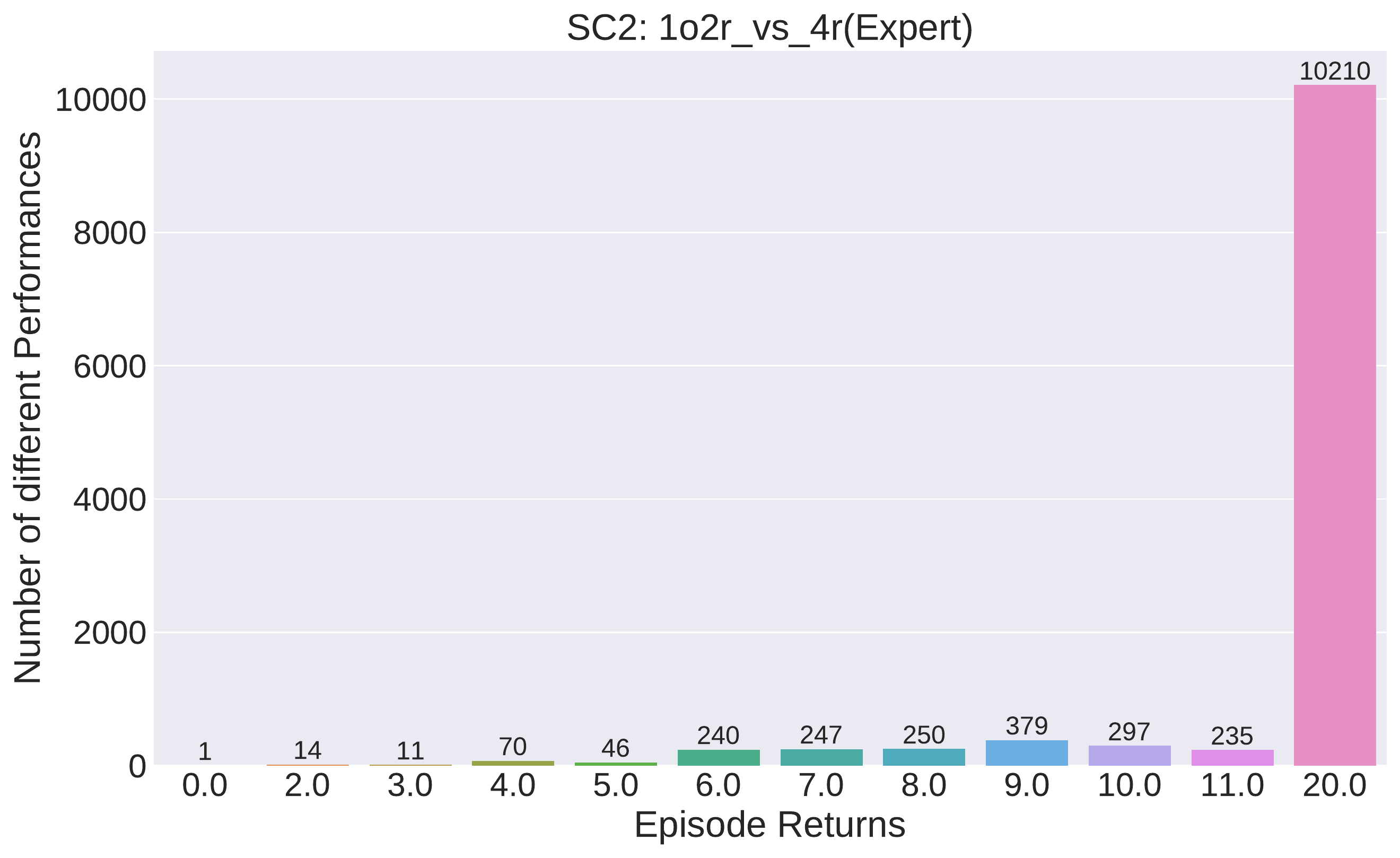}
    	\label{fig:SC2_1o2rVS4r(Expert)}
    	\end{subfigure}
    	\hspace{-0.7em}
    	\begin{subfigure}{0.33\linewidth}
    		\centering
    		\includegraphics[width=\linewidth]{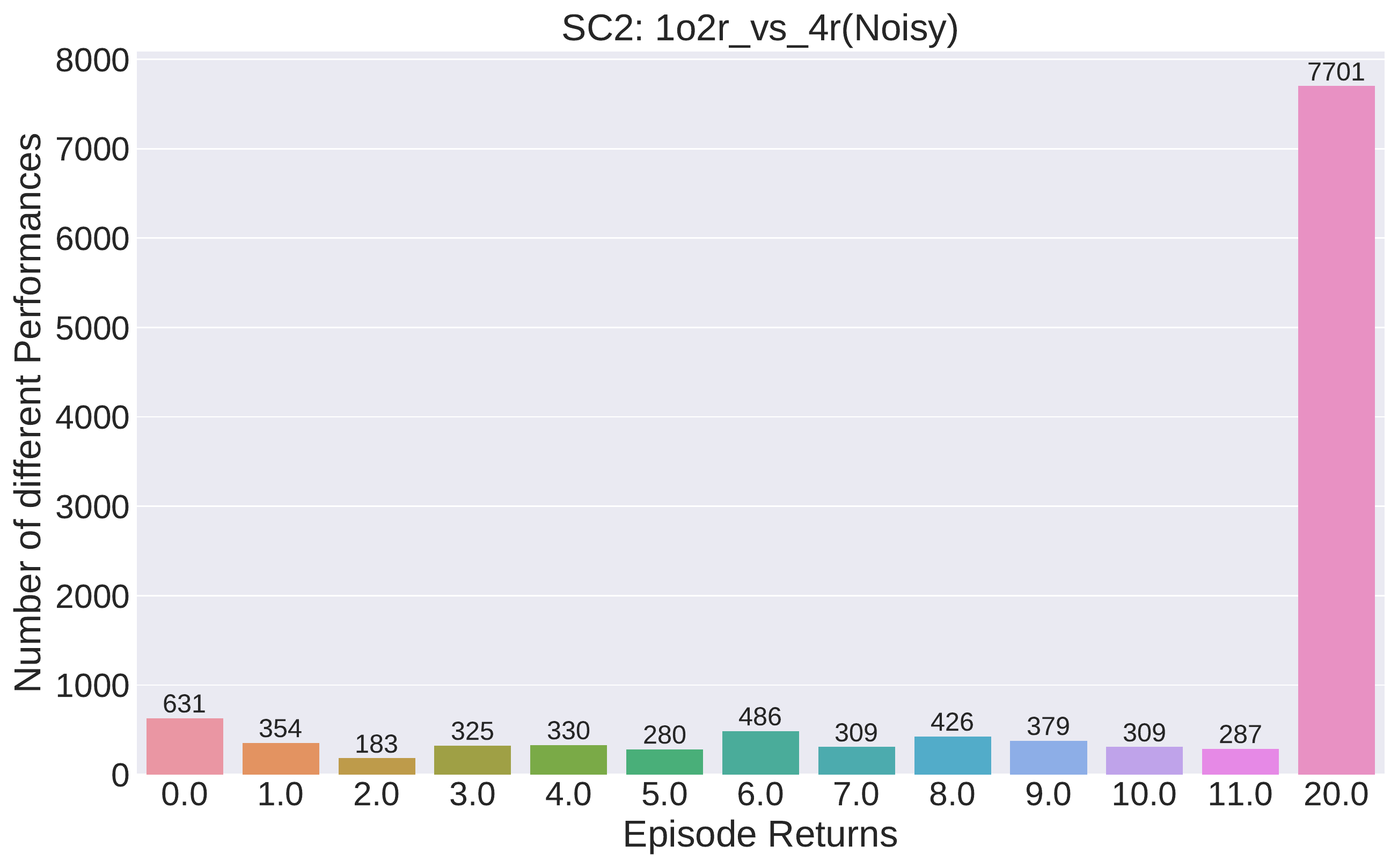}
    	\label{fig:SC2_1o2rVS4r(Noisy)}
    	\end{subfigure}
    	\hspace{-0.7em}
    	\begin{subfigure}{0.33\linewidth}
    		\centering
    		\includegraphics[width=\linewidth]{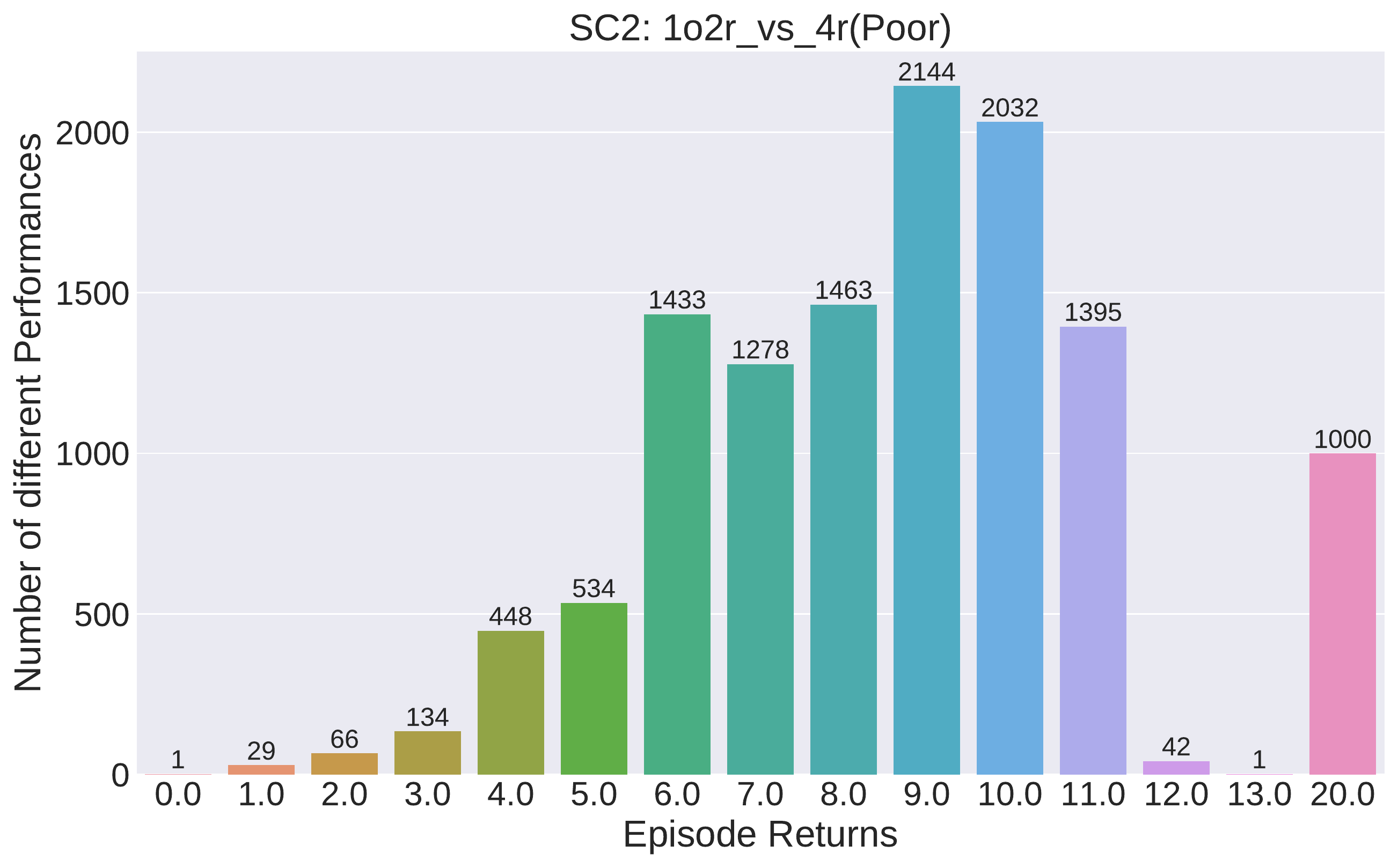}
    	\label{fig:SC2_1o2rVS4r(Poor)}
    	\end{subfigure}
	\end{subfigure}
	\begin{subfigure}{\linewidth}
        \centering
         \hspace{-0.7em}
    	\begin{subfigure}{0.33\linewidth}
    		\centering
    		\includegraphics[width=\linewidth]{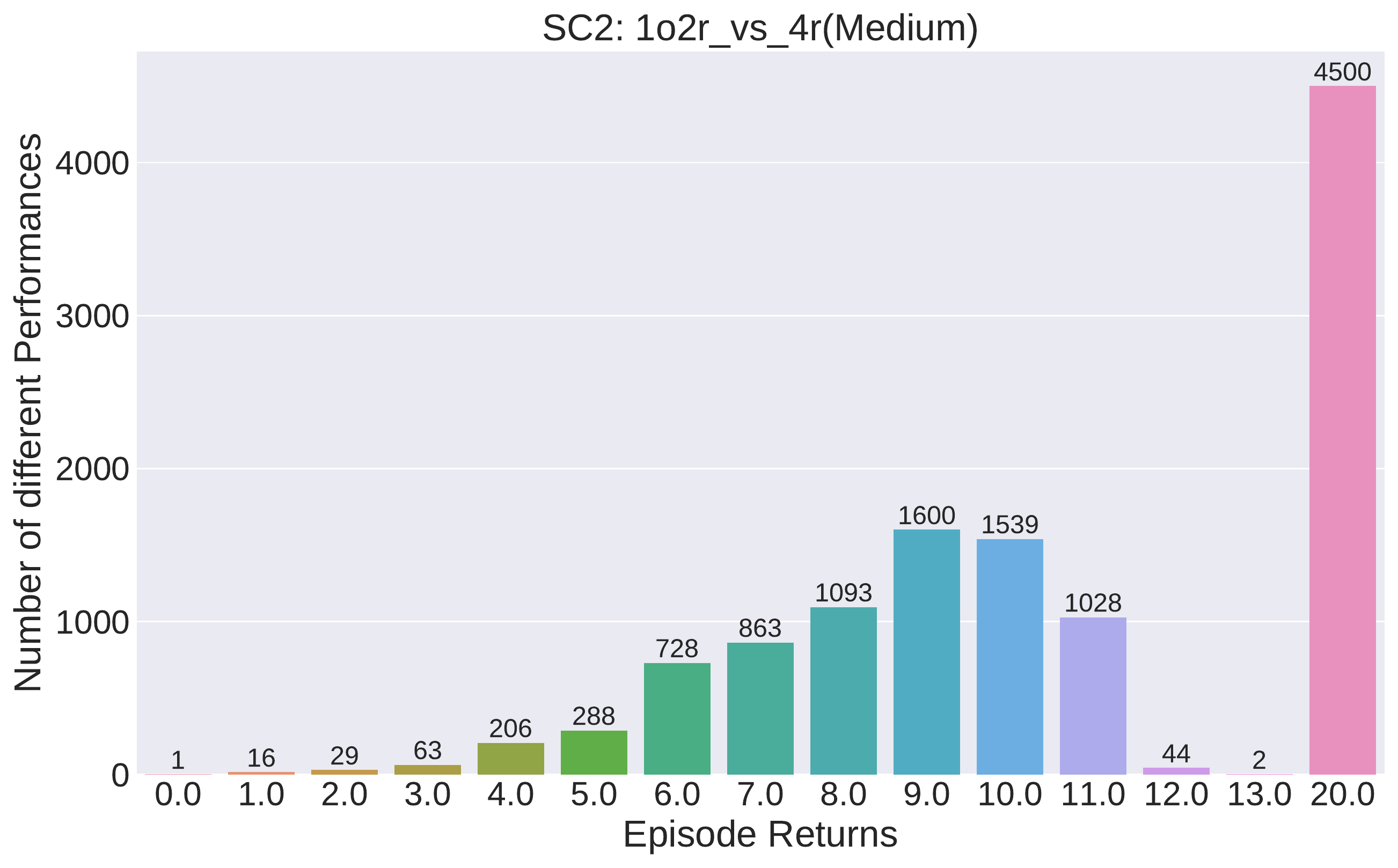}
    	\label{fig:SC2_1o2rVS4r(Medium)}
    	\end{subfigure}
    	\hspace{-0.7em}
    	\begin{subfigure}{0.33\linewidth}
    		\centering
    		\includegraphics[width=\linewidth]{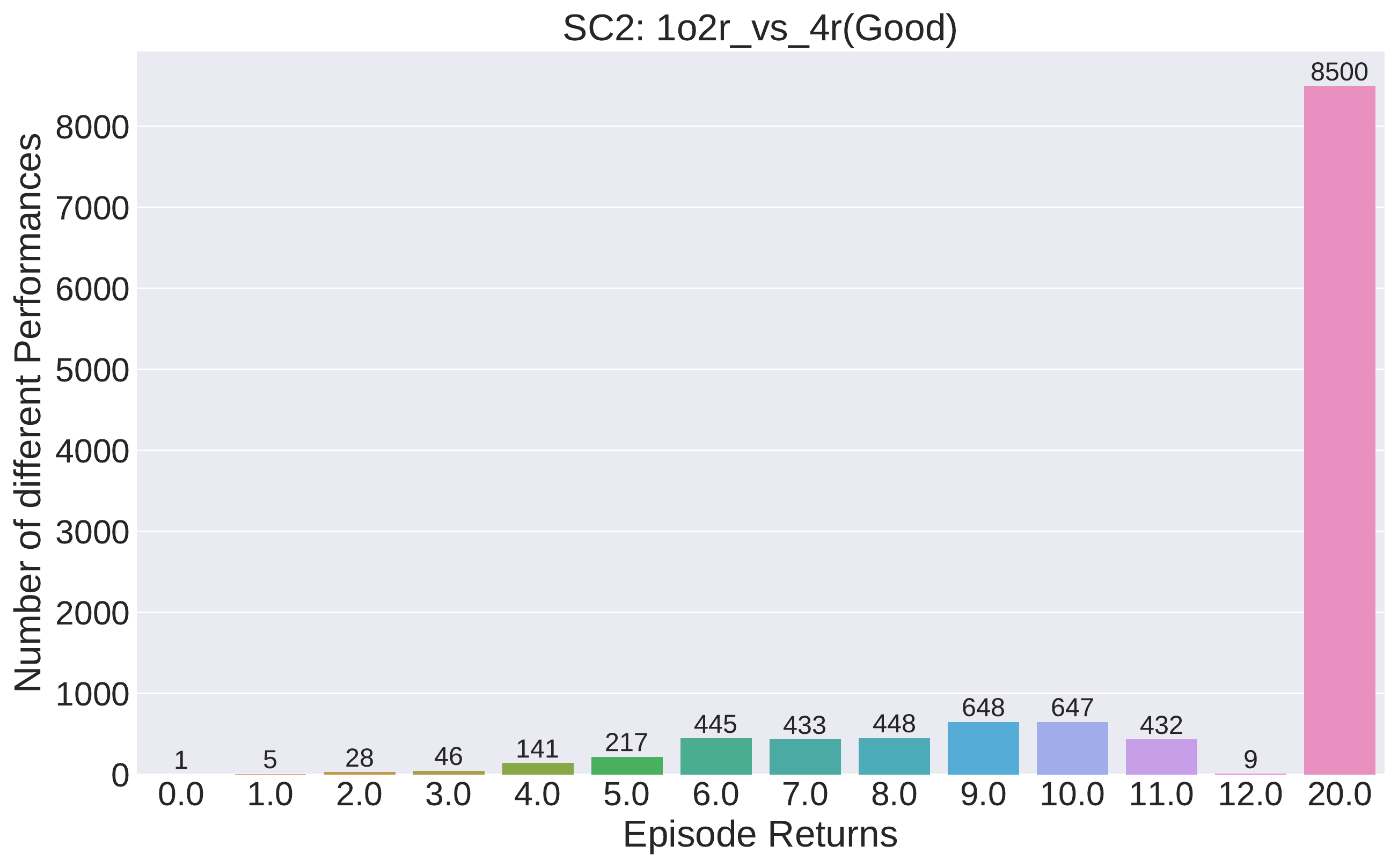}
    	\label{fig:SC2_1o2rVS4r(Good)}
    	\end{subfigure}
    	\hspace{-0.7em}
    	\begin{subfigure}{0.33\linewidth}
    		\centering
    		\includegraphics[width=\linewidth]{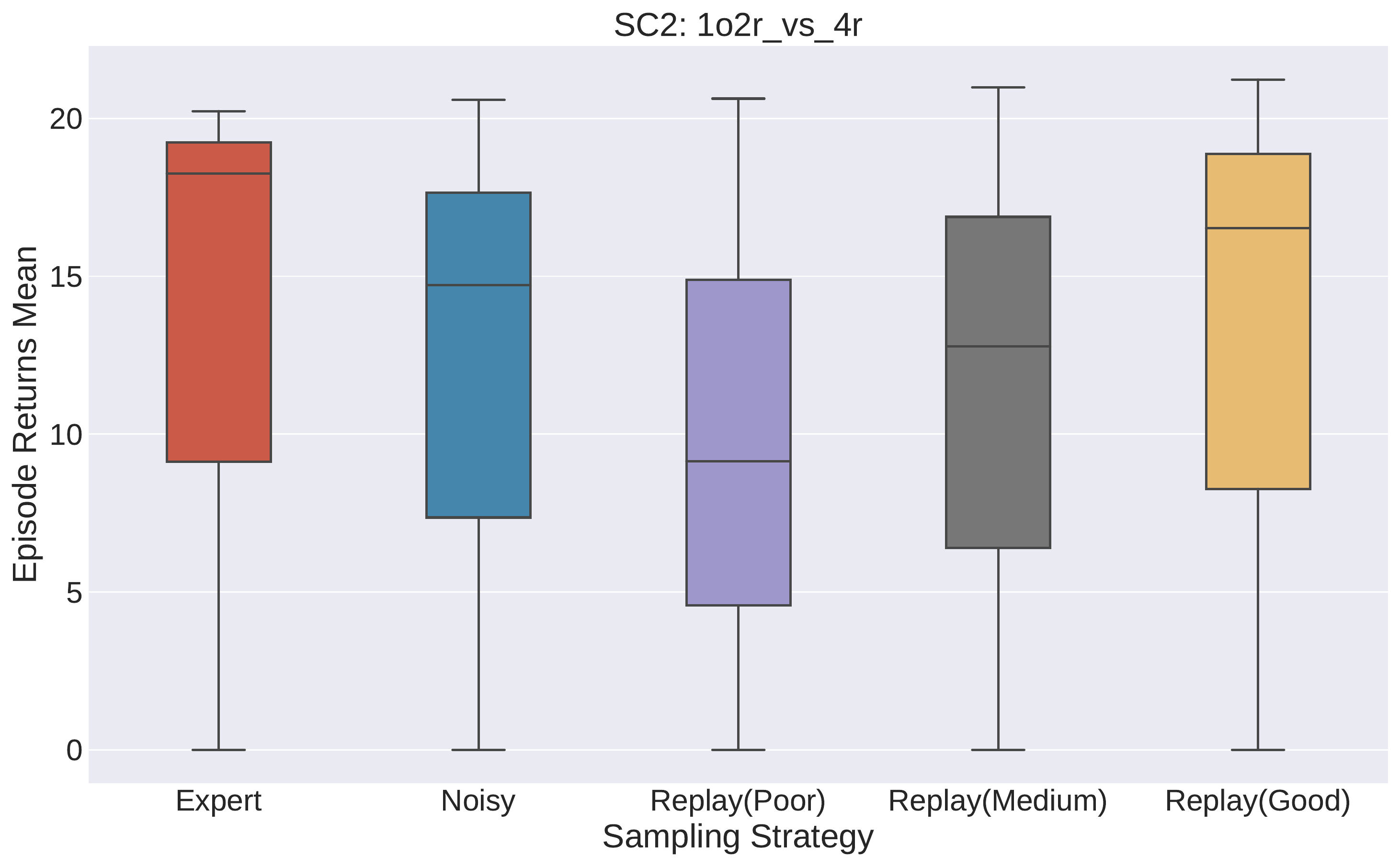}
    	\label{fig:SC2_1o2rVS4r(ALL)}
    	\end{subfigure}
	\end{subfigure}
	\caption{SMAC: 1o2r\_vs\_4r offline dataset distribution. }
	\label{fig:sc21_data_dist}
	\vspace*{-5mm}
\end{figure*}

\begin{figure*}[htbp]
	\centering
	\begin{subfigure}{\linewidth}
		\centering
    	\begin{subfigure}{0.33\linewidth}
    		\centering
    		\includegraphics[width=\linewidth]{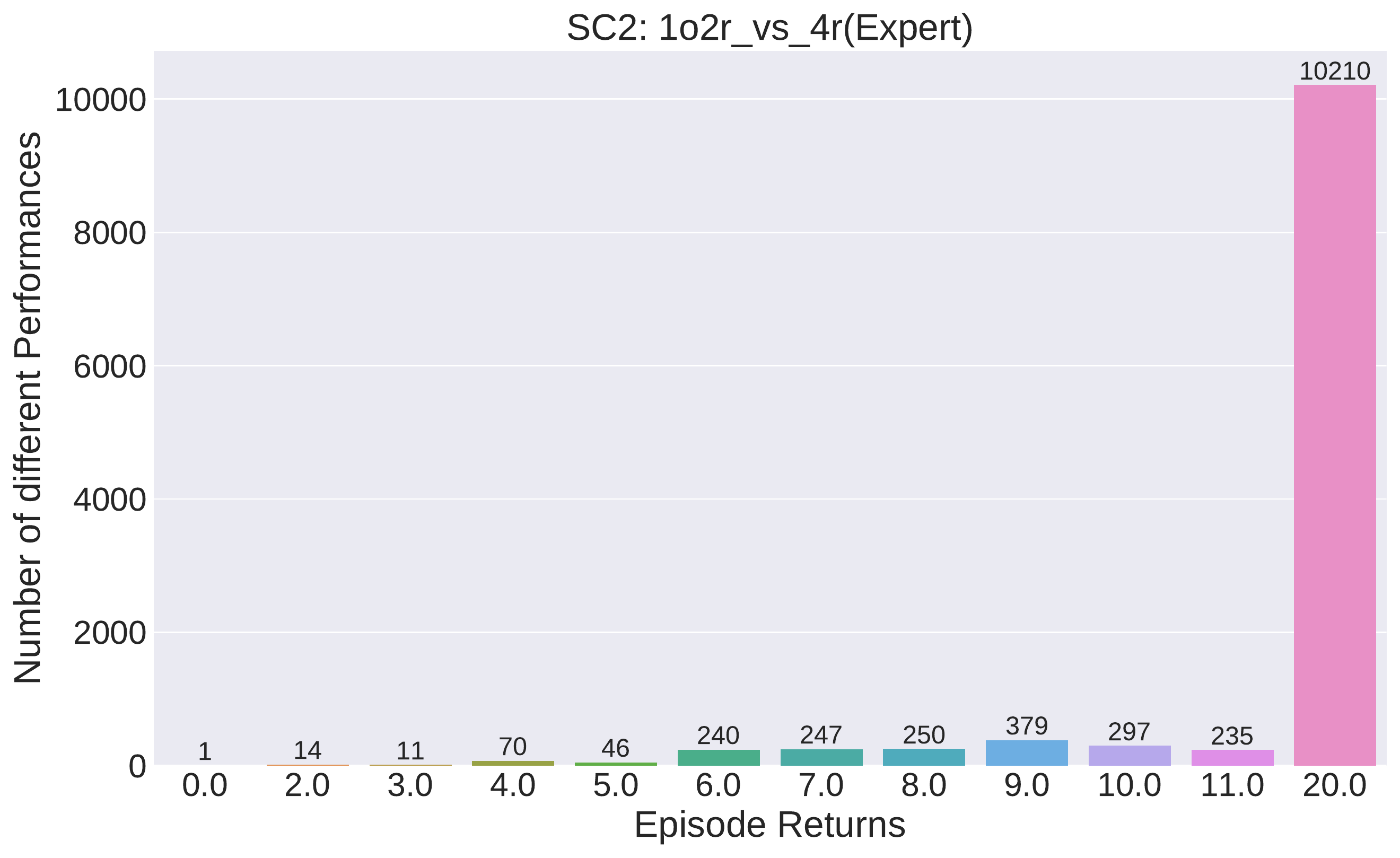}
    	\label{fig:SC2_1o10bVS1r(Expert)}
    	\end{subfigure}
    	\hspace{-0.7em}
    	\begin{subfigure}{0.33\linewidth}
    		\centering
    		\includegraphics[width=\linewidth]{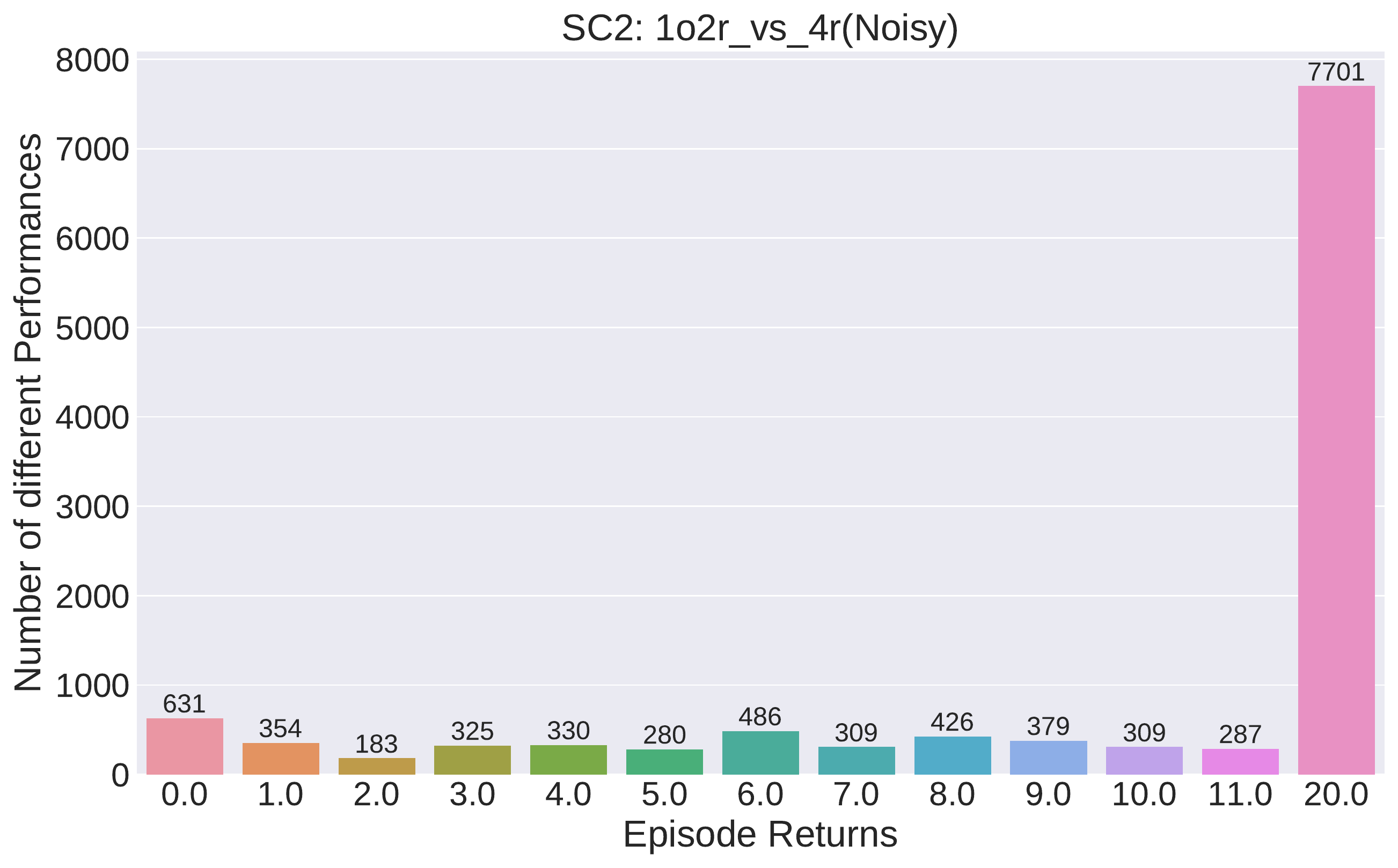}
    	\label{fig:SC2_1o10bVS1r(Noisy)}
    	\end{subfigure}
    	\hspace{-0.7em}
    	\begin{subfigure}{0.33\linewidth}
    		\centering
    		\includegraphics[width=\linewidth]{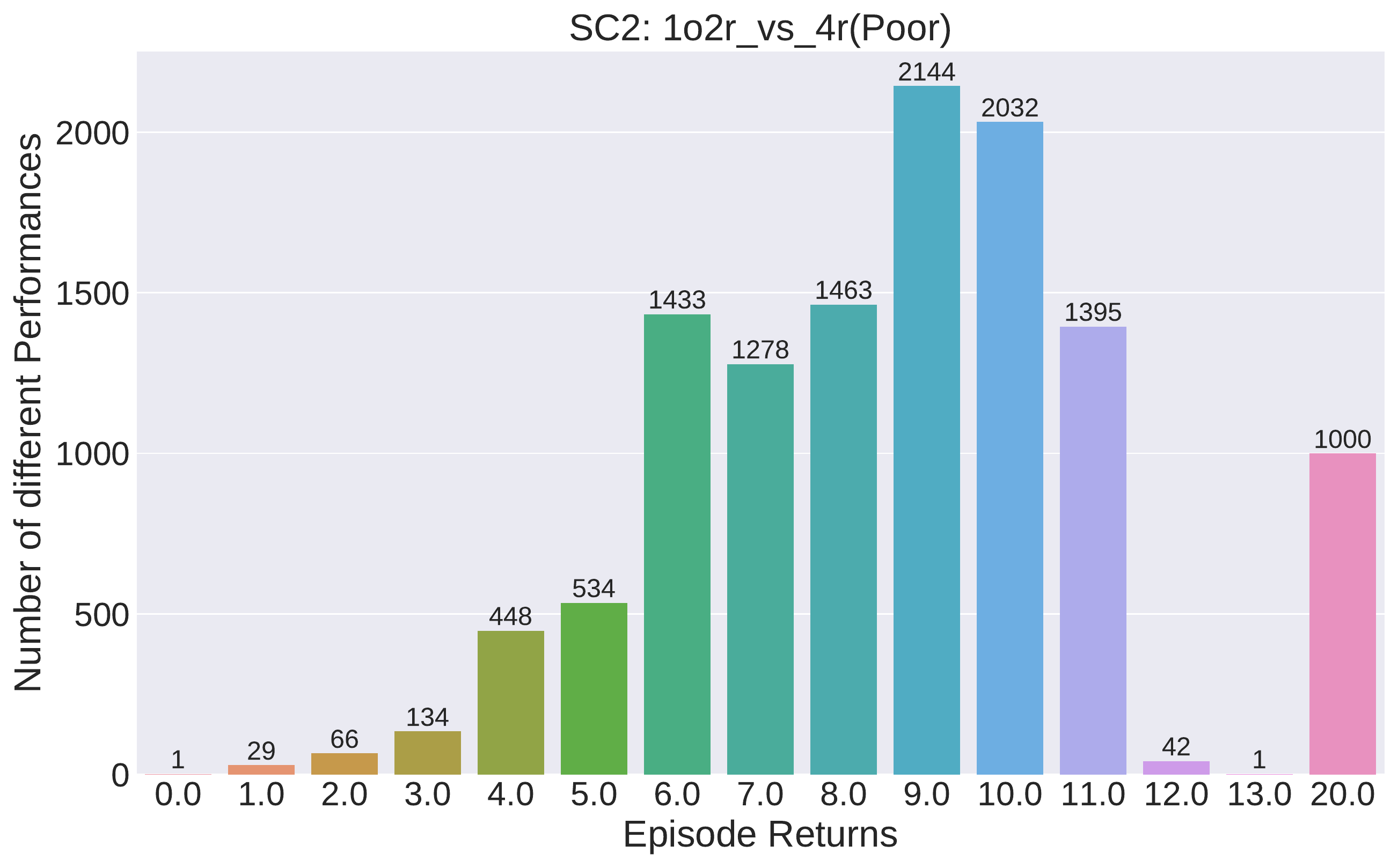}
    	\label{fig:SC2_1o10bVS1r(Poor)}
    	\end{subfigure}
	\end{subfigure}
	\begin{subfigure}{\linewidth}
        \centering
         \hspace{-0.7em}
    	\begin{subfigure}{0.33\linewidth}
    		\centering
    		\includegraphics[width=\linewidth]{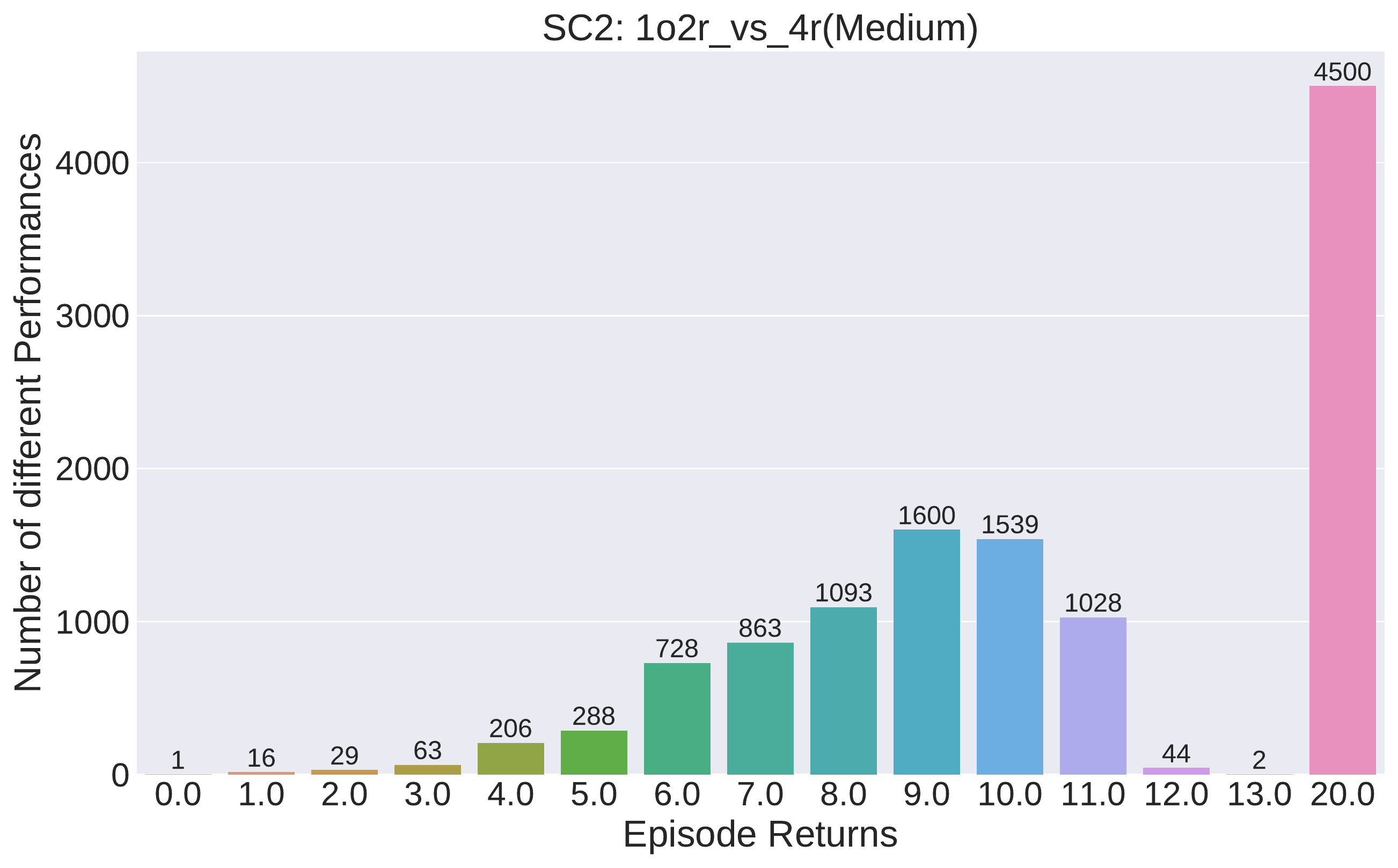}
    	\label{fig:SC2_1o10bVS1r(Medium)}
    	\end{subfigure}
    	\hspace{-0.7em}
    	\begin{subfigure}{0.33\linewidth}
    		\centering
    		\includegraphics[width=\linewidth]{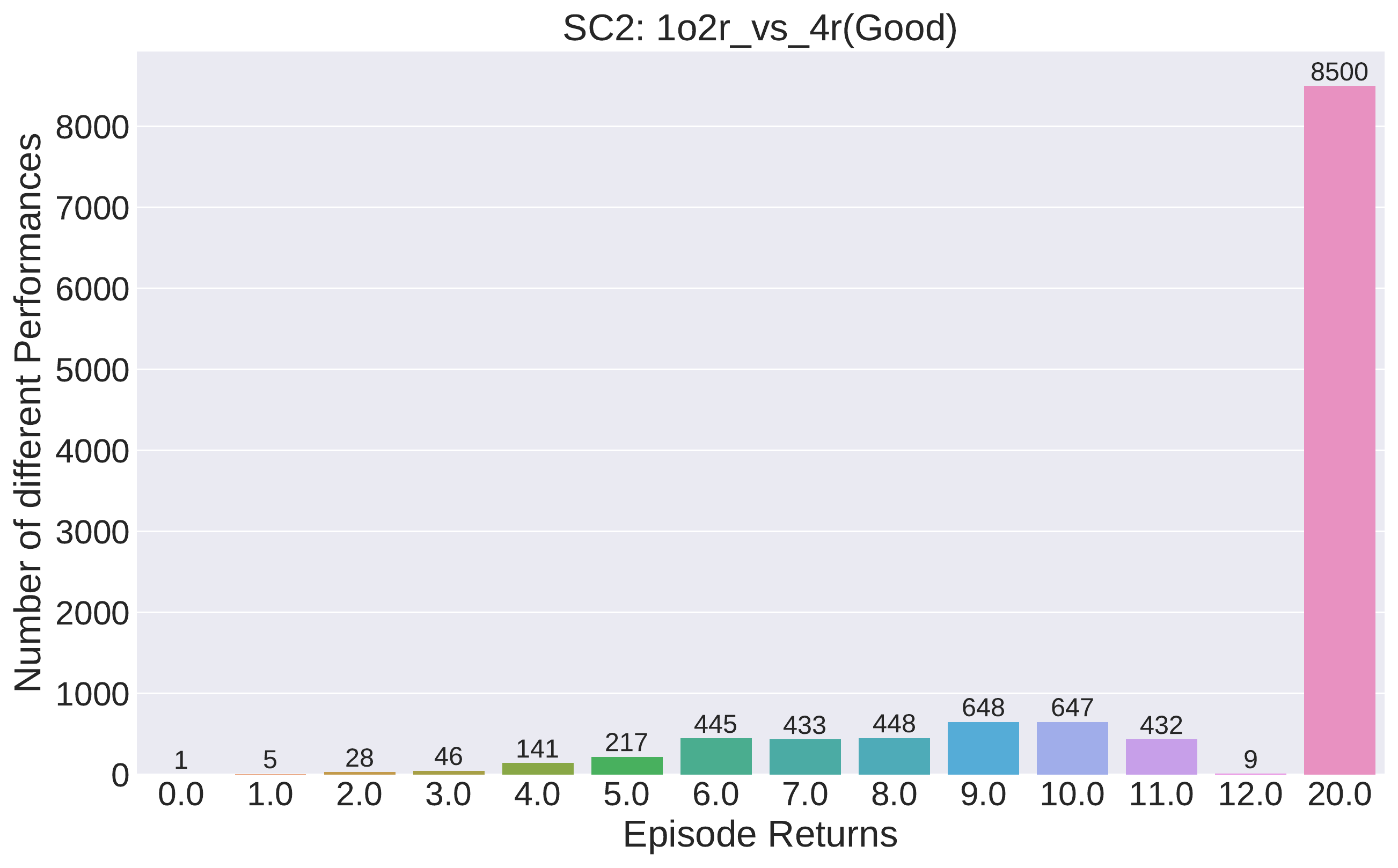}
    	\label{fig:SC2_1o10bVS1r(Good)}
    	\end{subfigure}
    	\hspace{-0.7em}
    	\begin{subfigure}{0.33\linewidth}
    		\centering
    		\includegraphics[width=\linewidth]{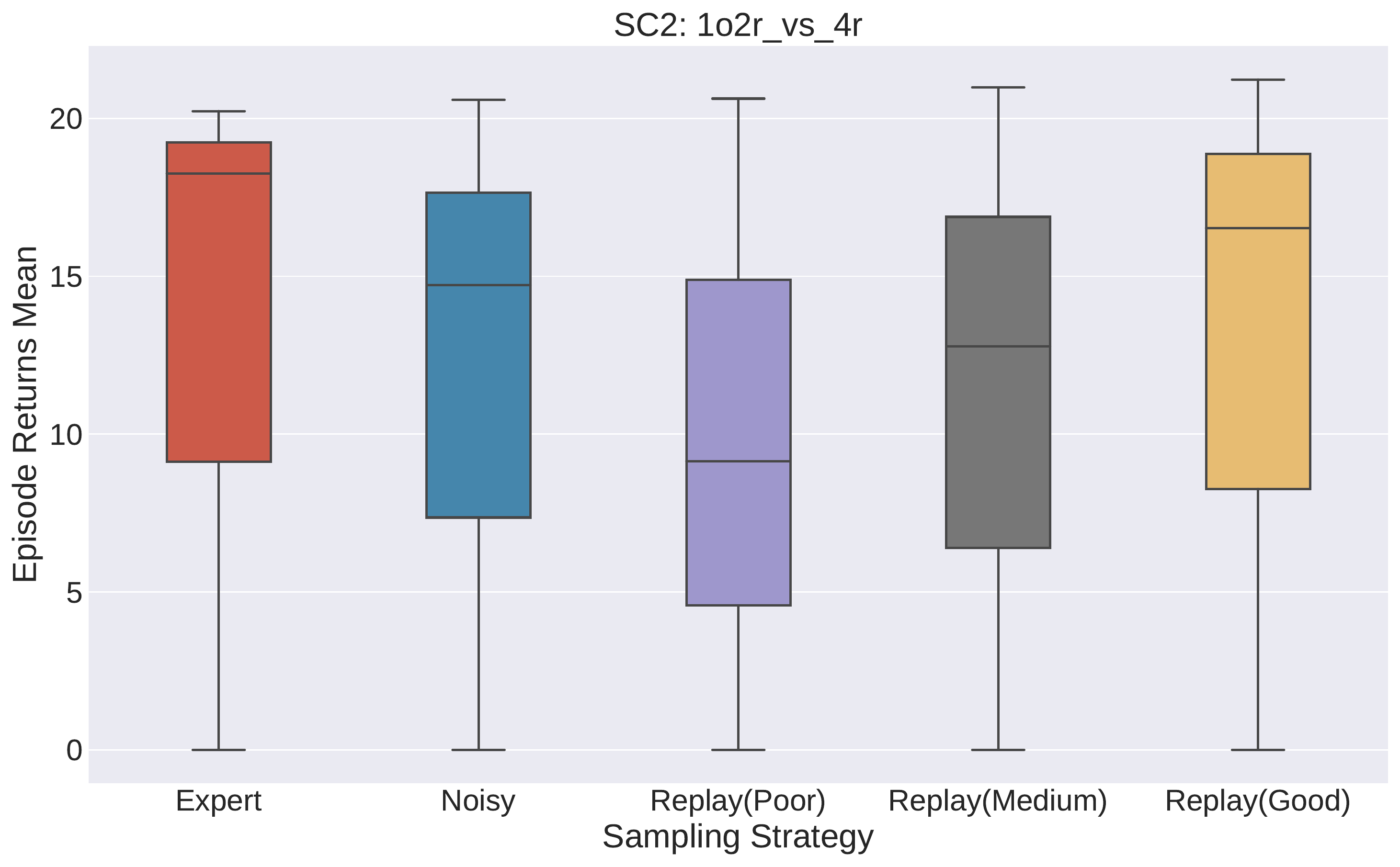}
    	\label{fig:SC2_1o10bVS1r(ALL)}
    	\end{subfigure}
	\end{subfigure}
	\caption{SMAC: 1o10b\_vs\_1r offline dataset distribution. }
	\label{fig:sc22_data_dist}
	\vspace*{-5mm}
\end{figure*}

\begin{figure*}[htbp]
	\centering
	\begin{subfigure}{\linewidth}
		\centering
    	\begin{subfigure}{0.33\linewidth}
    		\centering
    		\includegraphics[width=\linewidth]{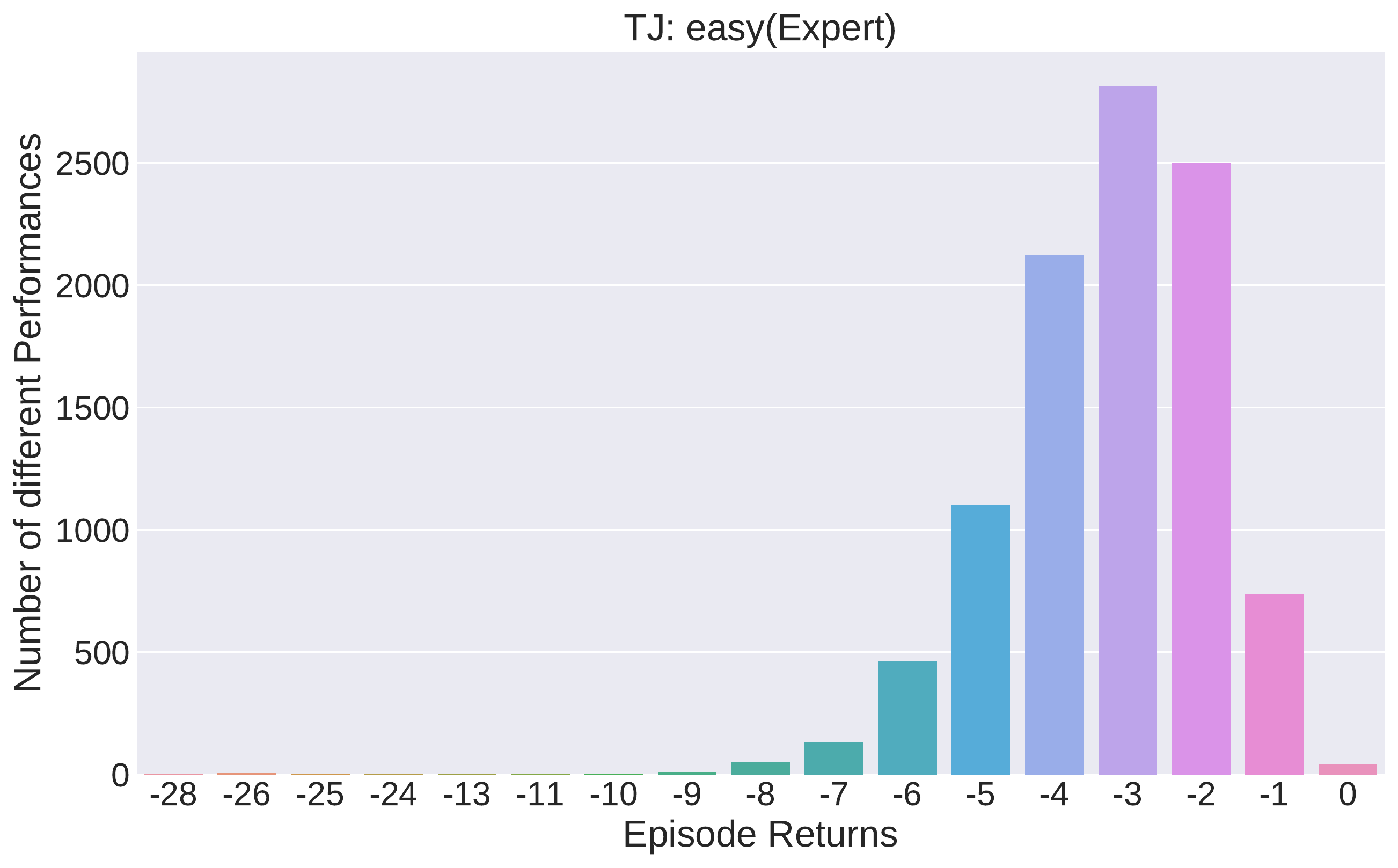}
    	\label{fig:TJ_easy(Expert)}
    	\end{subfigure}
    	\hspace{-0.7em}
    	\begin{subfigure}{0.33\linewidth}
    		\centering
    		\includegraphics[width=\linewidth]{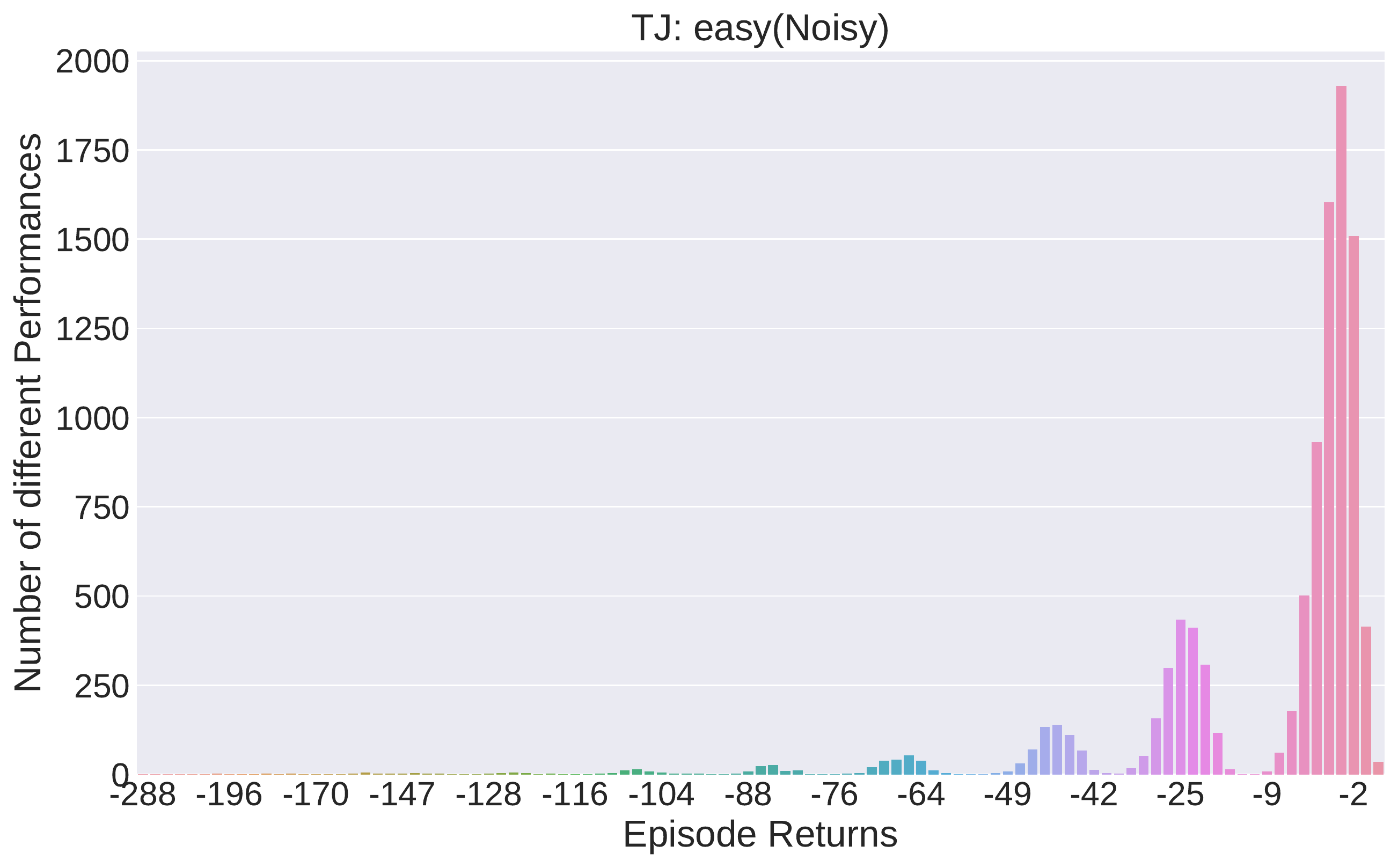}
    	\label{fig:TJ_easy(Noisy)}
    	\end{subfigure}
    	\hspace{-0.7em}
    	\begin{subfigure}{0.33\linewidth}
    		\centering
    		\includegraphics[width=\linewidth]{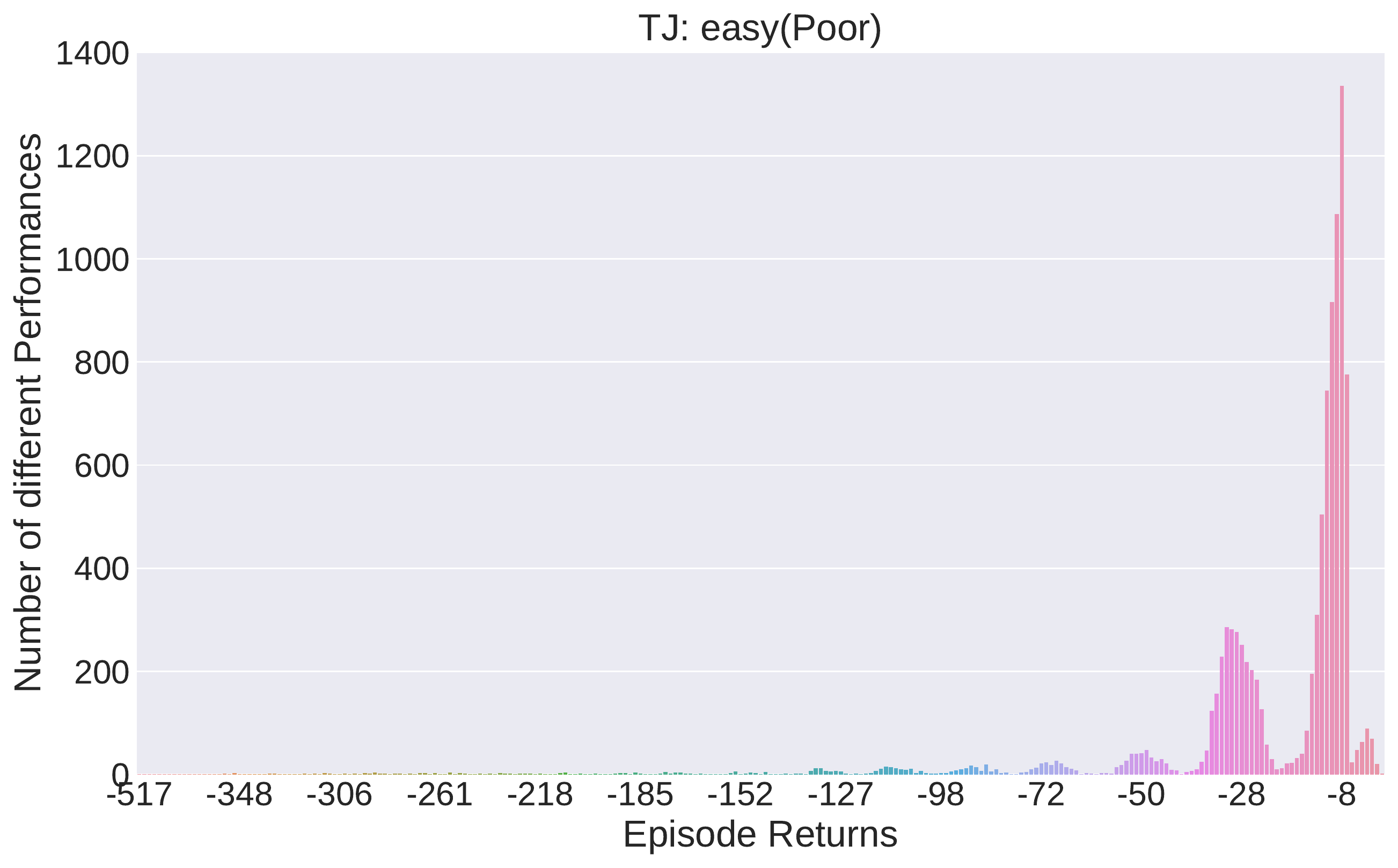}
    	\label{fig:TJ_easy(Poor)}
    	\end{subfigure}
	\end{subfigure}
	\begin{subfigure}{\linewidth}
        \centering
         \hspace{-0.7em}
    	\begin{subfigure}{0.33\linewidth}
    		\centering
    		\includegraphics[width=\linewidth]{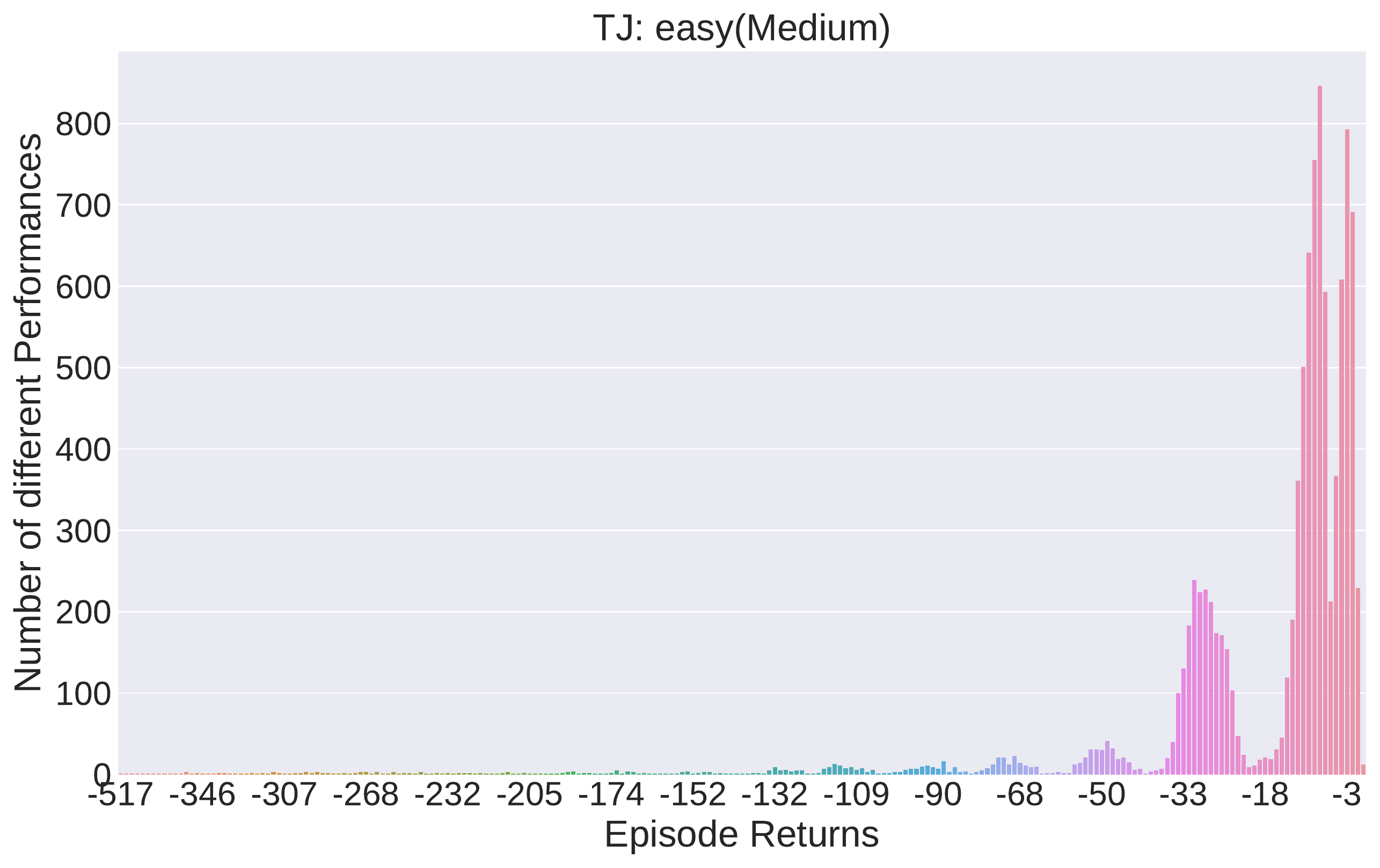}
    	\label{fig:TJ_easy(Medium)}
    	\end{subfigure}
    	\hspace{-0.7em}
    	\begin{subfigure}{0.33\linewidth}
    		\centering
    		\includegraphics[width=\linewidth]{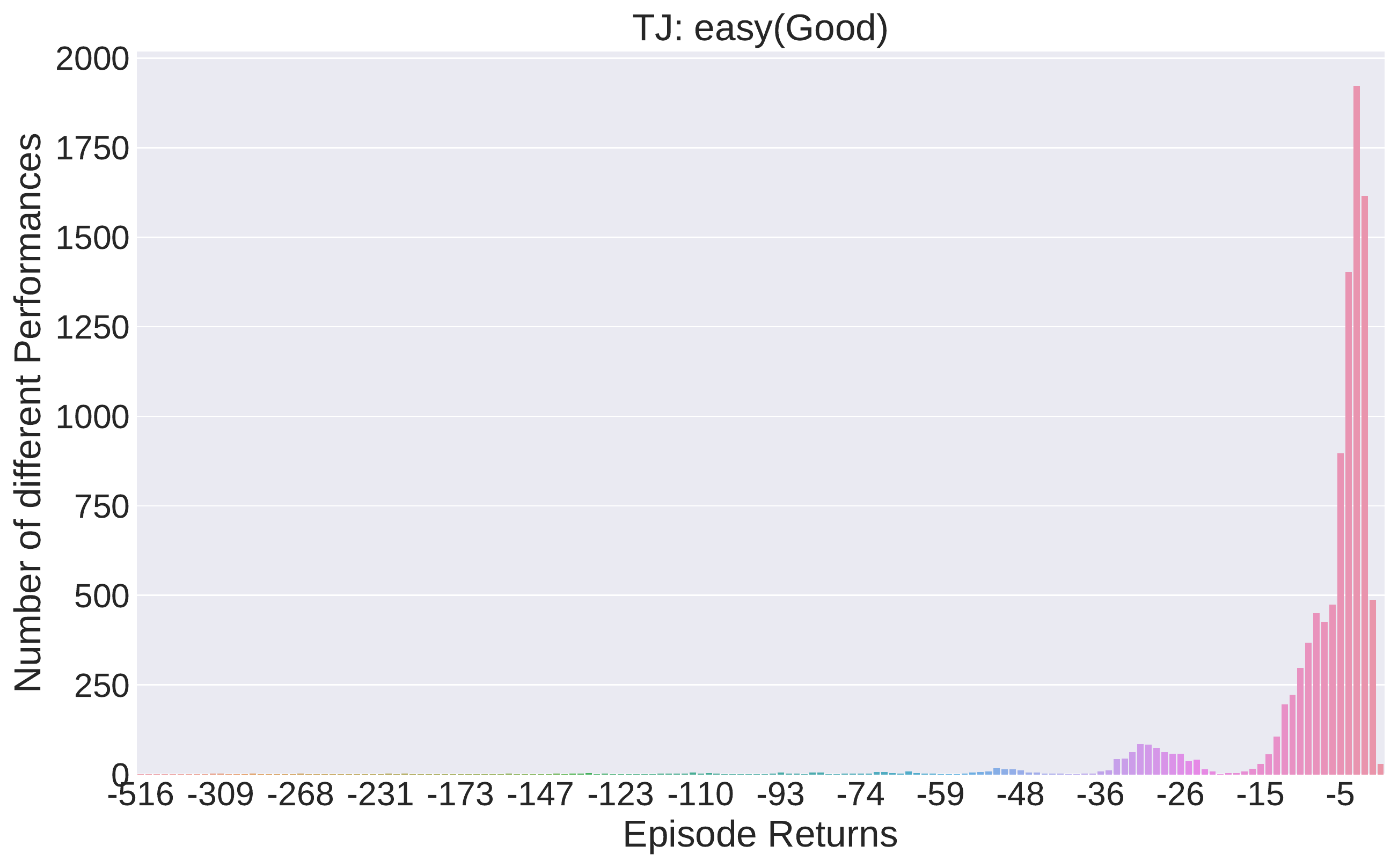}
    	\label{fig:TJ_easy(Good)}
    	\end{subfigure}
    	\hspace{-0.7em}
    	\begin{subfigure}{0.33\linewidth}
    		\centering
    		\includegraphics[width=\linewidth]{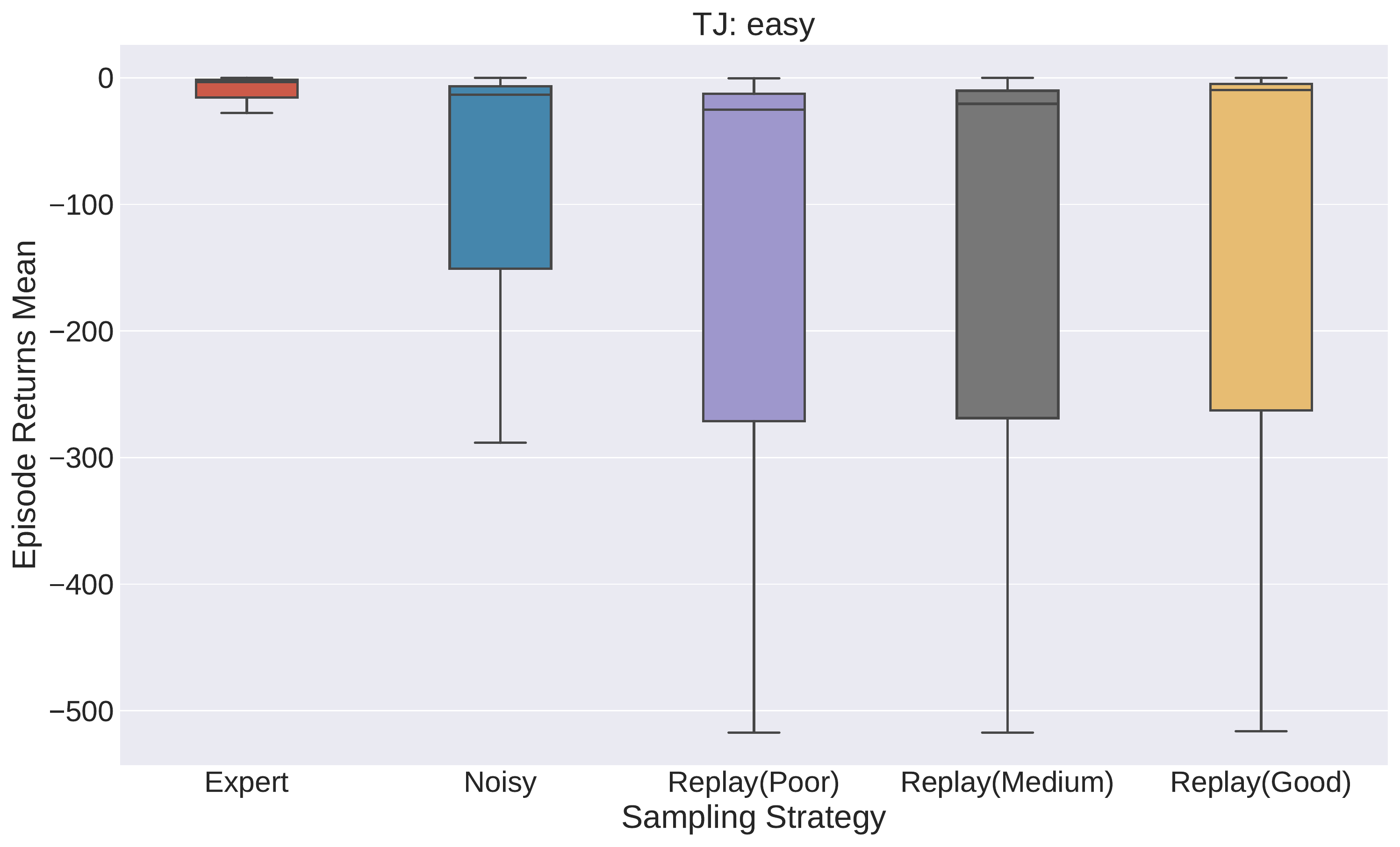}
    	\label{fig:TJ_easy(ALL)}
    	\end{subfigure}
	\end{subfigure}
	\caption{TJ: easy offline dataset distribution. }
	\label{fig:tj_easy_data_dist}
	\vspace*{-5mm}
\end{figure*}

\begin{figure*}[htbp]
	\centering
	\begin{subfigure}{\linewidth}
		\centering
    	\begin{subfigure}{0.33\linewidth}
    		\centering
    		\includegraphics[width=\linewidth]{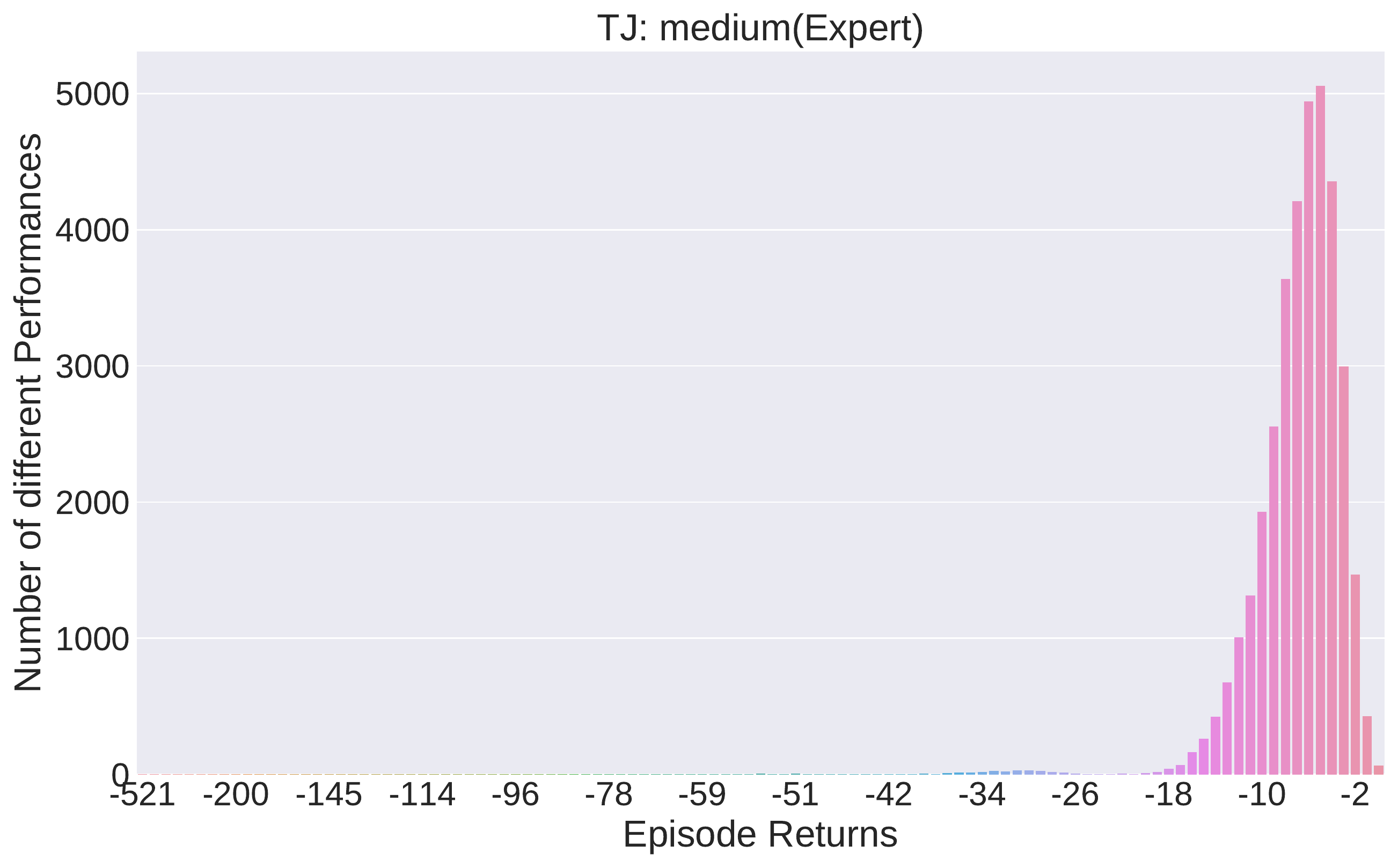}
    	\label{fig:TJ_medium(Expert)}
    	\end{subfigure}
    	\hspace{-0.7em}
    	\begin{subfigure}{0.33\linewidth}
    		\centering
    		\includegraphics[width=\linewidth]{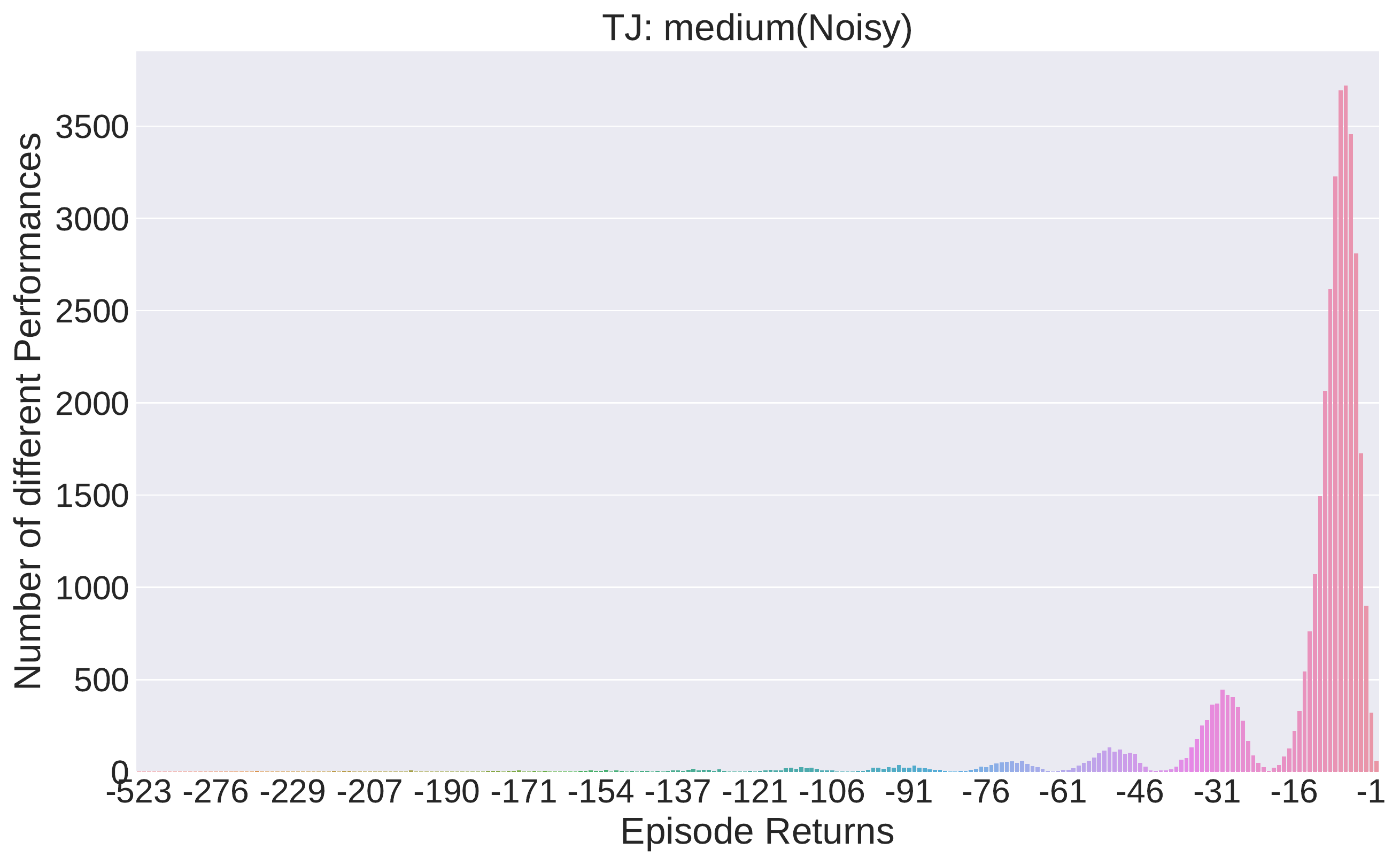}
    	\label{fig:TJ_medium(Noisy)}
    	\end{subfigure}
    	\hspace{-0.7em}
    	\begin{subfigure}{0.33\linewidth}
    		\centering
    		\includegraphics[width=\linewidth]{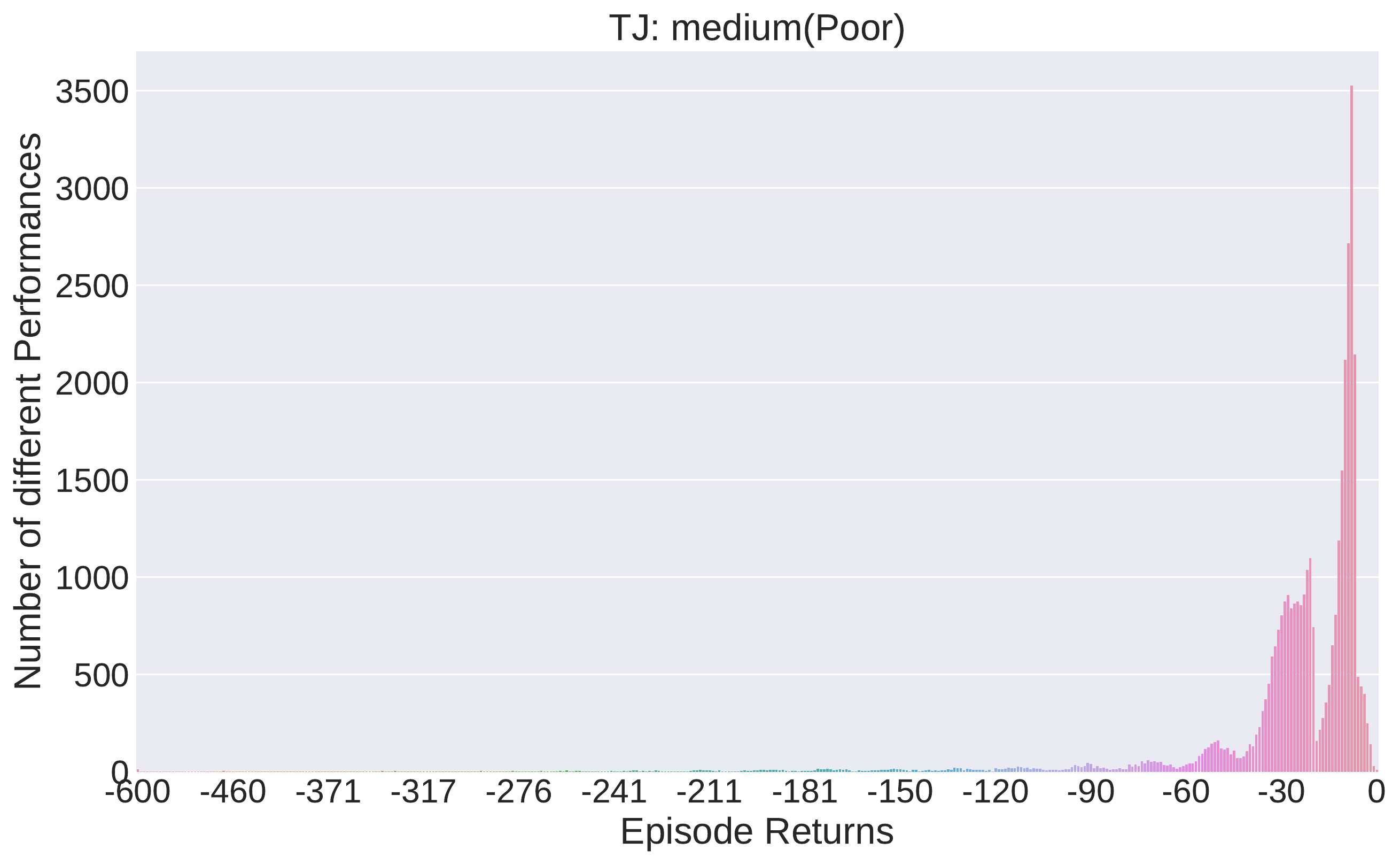}
    	\label{fig:TJ_medium(Poor)}
    	\end{subfigure}
	\end{subfigure}
	\begin{subfigure}{\linewidth}
        \centering
         \hspace{-0.7em}
    	\begin{subfigure}{0.33\linewidth}
    		\centering
    		\includegraphics[width=\linewidth]{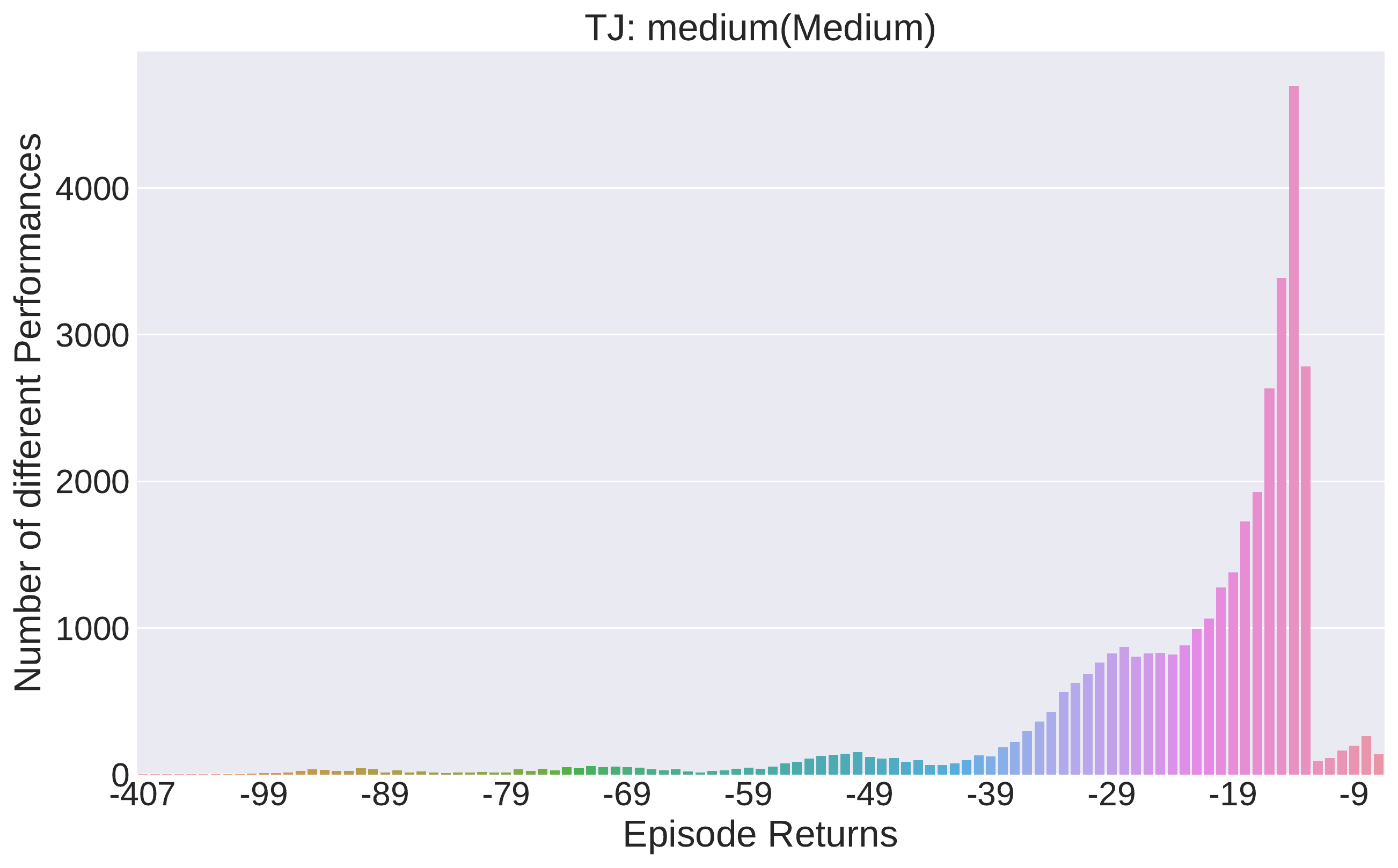}
    	\label{fig:TJ_medium(Medium)}
    	\end{subfigure}
    	\hspace{-0.7em}
    	\begin{subfigure}{0.33\linewidth}
    		\centering
    		\includegraphics[width=\linewidth]{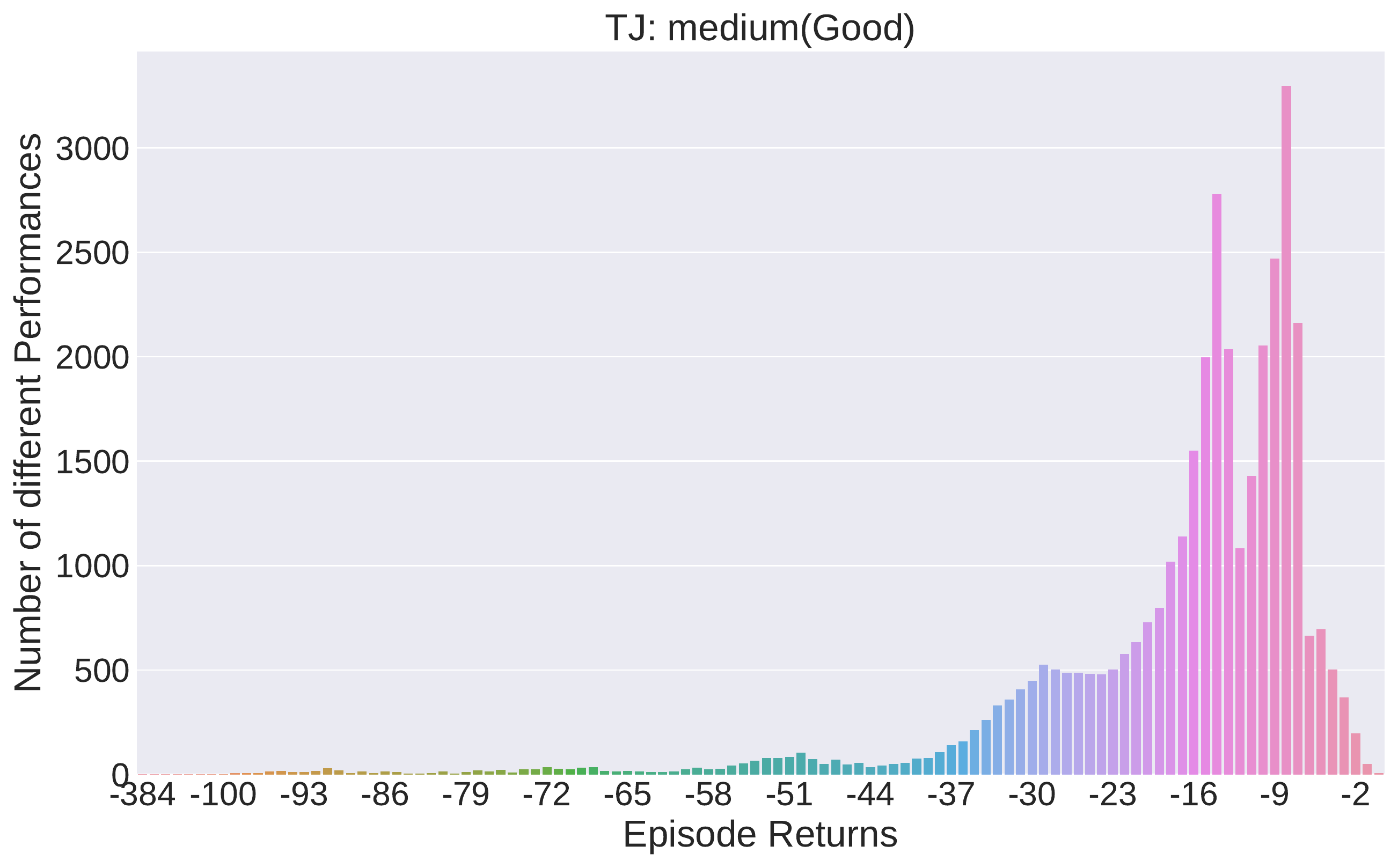}
    	\label{fig:TJ_medium(Good)}
    	\end{subfigure}
    	\hspace{-0.7em}
    	\begin{subfigure}{0.33\linewidth}
    		\centering
    		\includegraphics[width=\linewidth]{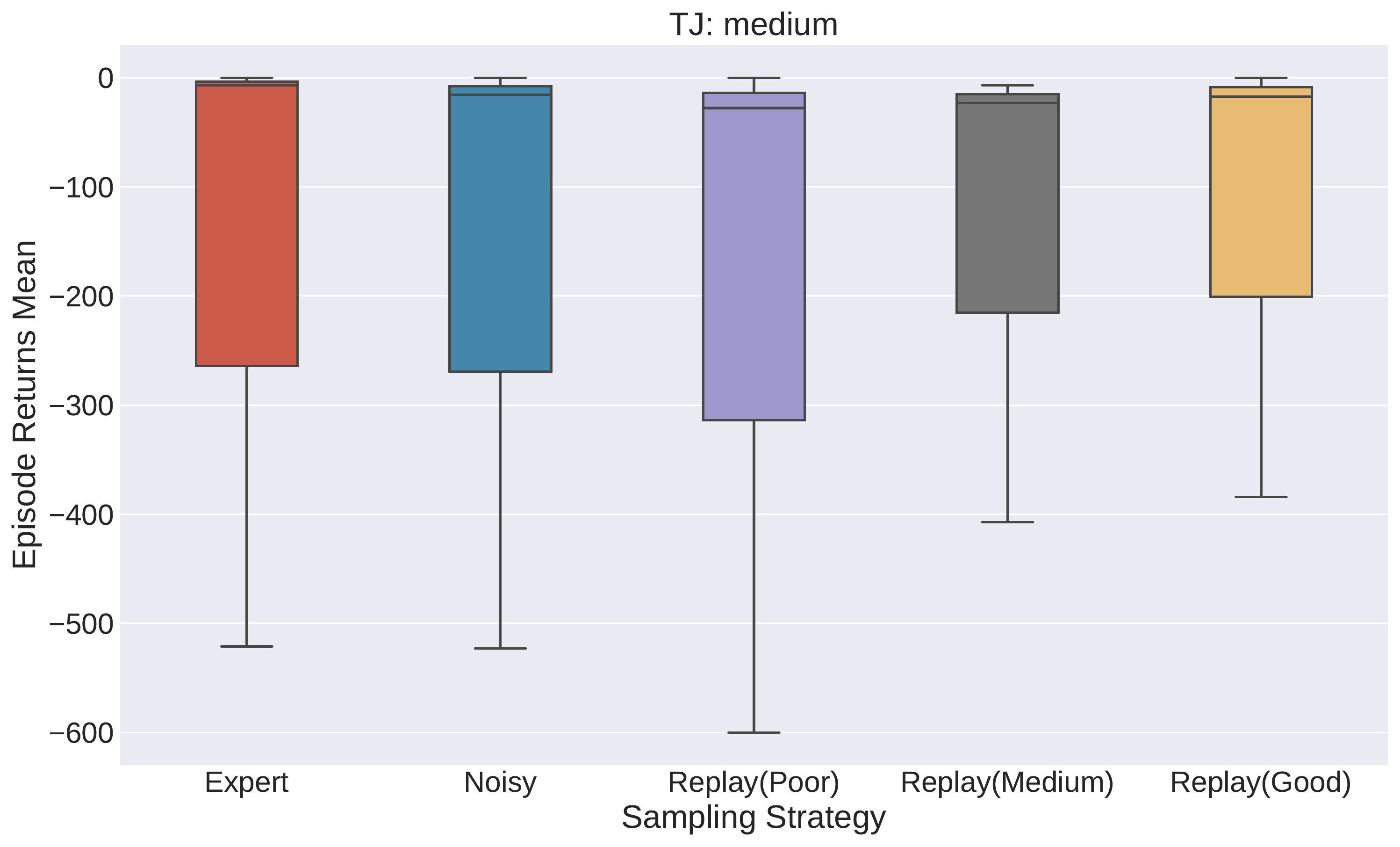}
    	\label{fig:TJ_medium(ALL)}
    	\end{subfigure}
	\end{subfigure}
	\caption{TJ: medium offline dataset distribution. }
	\label{fig:tj_medium_data_dist}
	\vspace*{-5mm}
\end{figure*}

Moreover, considering that the real data may be of various qualities, we include multiple dataset settings to cover different data distributions. In Section~\ref{sec:offline_exp}, we introduce all there different dataset settings and how we construct them. Actually, due to the differences in data collection approaches, these datasets somewhat differ in the quality of data. For example, the data in \textbf{Expert Dataset} are typically trajectories of high returns. More detailed information about these datasets is listed in Table~\ref{tab:data_qualities}, where we offer the return distributions of different datasets. Furthermore, to offer a more intuitive presentation of the data distribution, we plot more information in Figures~\ref{fig:hallway_data_dist}-\ref{fig:tj_medium_data_dist}.\remove{\zzz{Do colors in these figures have meaning?}\explain{Explain: There are not specific meanings for them. The main purpose is to make the picture not too boring, but generally the colors represent the magnitude of return values.}} Take the task scenario of LBF: 11x11-6p-4f-s1 as an example, the return values of the most trajectories in \textbf{Expert}, \textbf{Noisy}, and \textbf{Replay (Good) Datasets} are concentrated on $1.0$ which is the optimal return value in this task. While no trajectory in \textbf{Replay (Poor)} and \textbf{Replay (Medium) Datasets} reaches a return value of $1.0$ and most trajectories are concentrated at much lower return values.

\subsection{More Results of Offline Learning Ablations}
\label{apdx:ablation}
\begin{figure}[htbp]
	\centering
	\begin{subfigure}{0.95\linewidth}
	    \centering
	    \includegraphics[width=\linewidth]{Figure/offline_exp/offline_ablation_legend.pdf}
	\end{subfigure}
	\centering
     	\begin{subfigure}{0.45\linewidth}
     		\centering
     		\includegraphics[width=\linewidth]{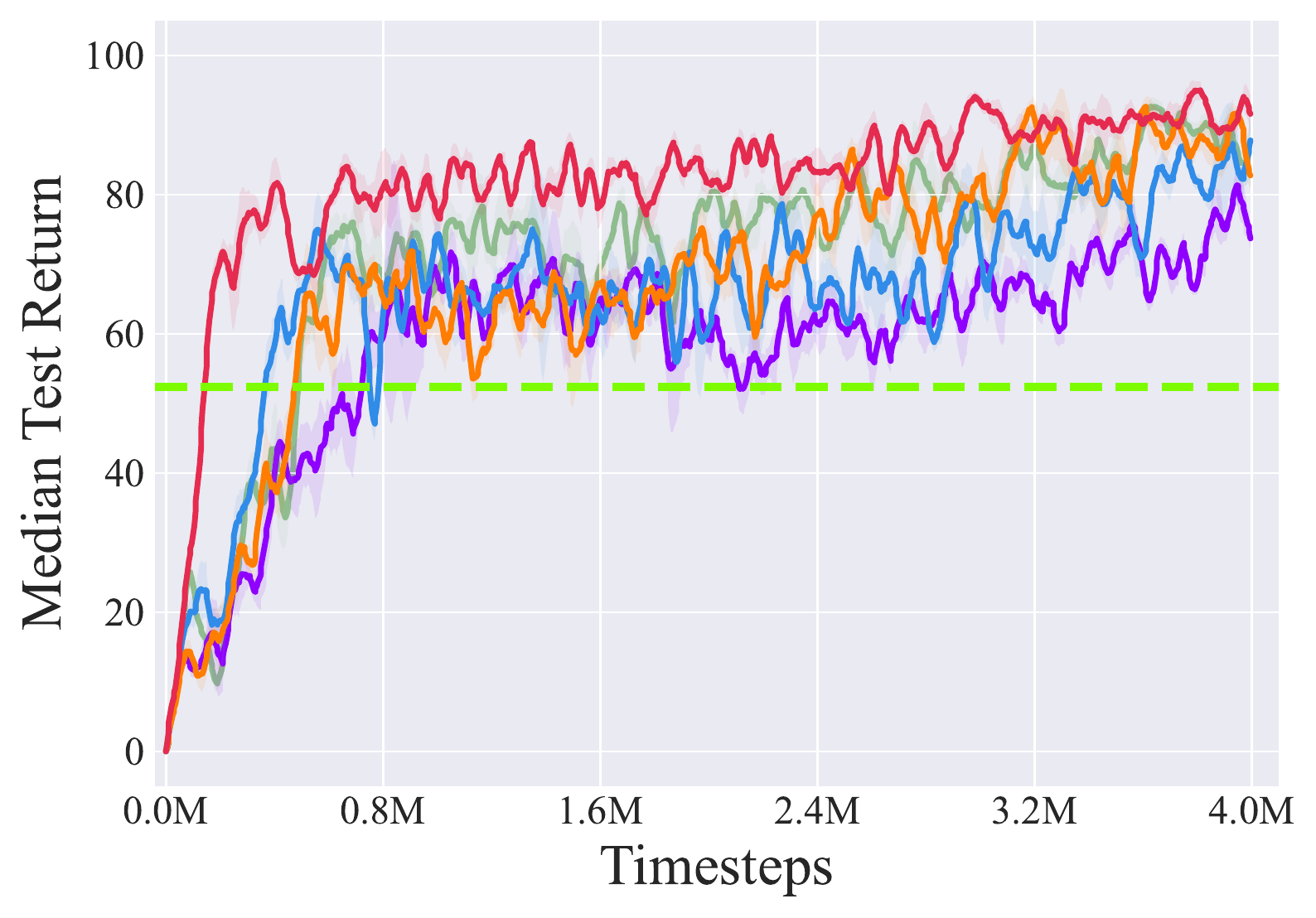}
     	\caption{Replay (Medium)}
     	\label{fig:Hallway(Medium)}
     	\end{subfigure}
	\centering
     	\begin{subfigure}{0.45\linewidth}
     		\centering
     		\includegraphics[width=\linewidth]{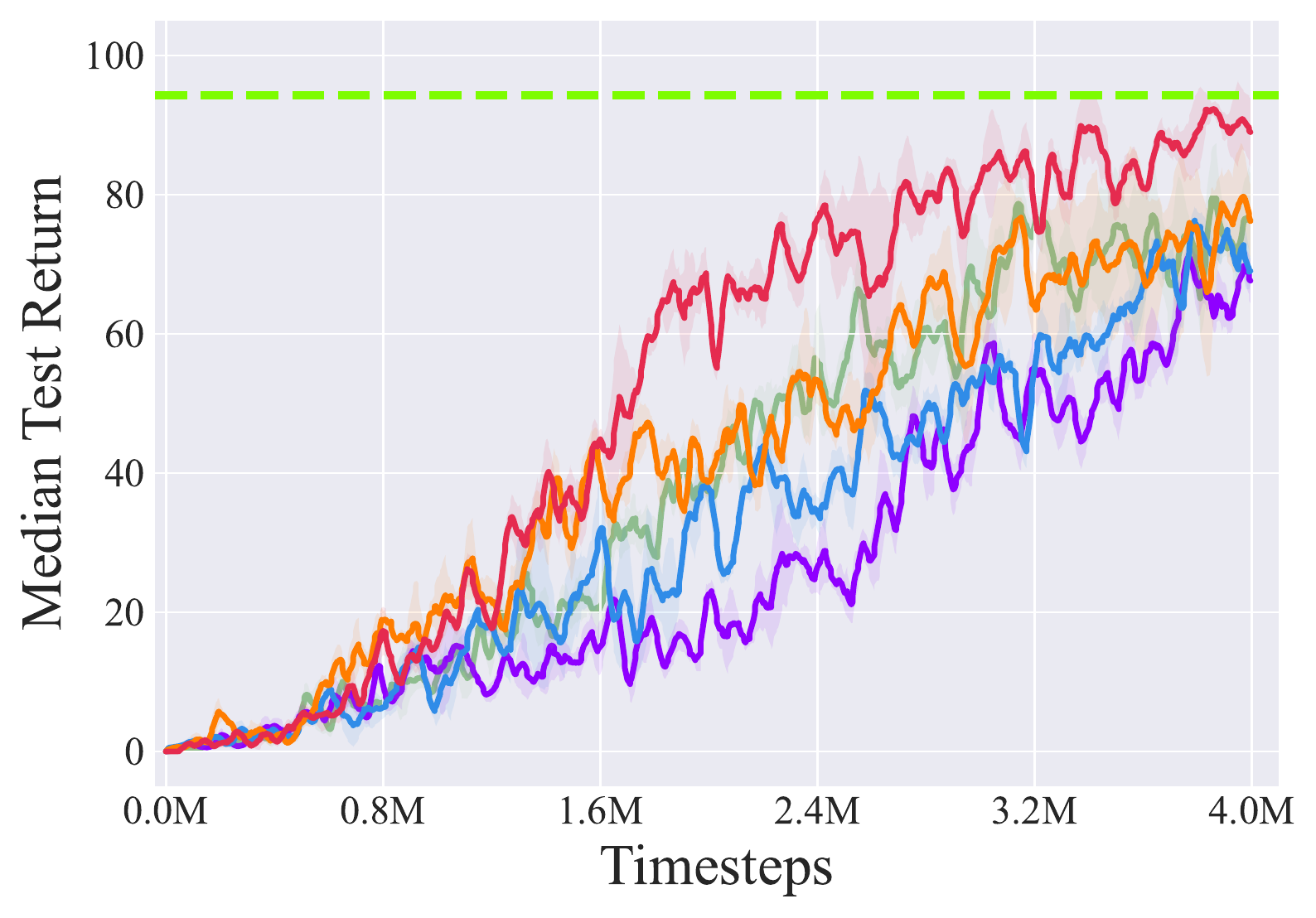}
     	\caption{Expert}
     	\label{fig:Hallway(Poor)}
     	\end{subfigure}
	\caption{Ablation experiments for offline learning on the task of TJ: medium. }
	\label{fig:offline_tj_ablation}
\end{figure}
In Sec.~\ref{sec:offline_ablation}, we conduct offline ablation studies on \textbf{Replay (Medium)} and \textbf{Replay (Poor)} datasets of Hallway: 3x5-4x6x10 to verify the effectiveness of two unsupervised representation losses and the practice of offline pre-training. Here, we further provide the ablation results on the task of TJ: medium. As we can see from Figure~\ref{fig:offline_tj_ablation}, the conclusion on TJ: medium is similar to that we obtained in the maintext, \name~equipped with both unsupervised representation losses and offline pre-training achieves faster learning speed and better convergence performance over other ablations.

\end{document}